\let\mypdfximage\pdfximage
\def\pdfximage{\immediate\mypdfximage}

\documentclass[journal,onecolumn,10pt]{IEEEtran1}

\makeatletter
\def\markboth#1#2{\def\leftmark{\@IEEEcompsoconly{\sffamily}\MakeUppercase{\protect#1}}%
\def\rightmark{\@IEEEcompsoconly{\sffamily}\MakeUppercase{\protect#2}}}
\makeatother

\let\labelindent\relax

\usepackage[T1]{fontenc}
\usepackage{tgtermes}
\usepackage{marvosym}
\usepackage{longtable}
\usepackage{graphics}
\usepackage{graphicx}
\usepackage{caption}
\usepackage[bgreek,english]{babel}
\usepackage{subfig}
\usepackage{epsfig}
\usepackage{color}
\usepackage{multirow}
\usepackage{mathrsfs}
\usepackage{textcomp}
\usepackage{amsfonts}
\usepackage{amsmath}
\usepackage{amssymb}
\usepackage{nccmath}
\usepackage[numbers,square,sort&compress,comma]{natbib}
\usepackage{chapterbib}
\usepackage[ruled,vlined]{algorithm2e}
\usepackage{setspace}
\usepackage{verbatim}
\usepackage{wrapfig}
\usepackage{dblfloatfix}
\usepackage{comment}
\usepackage[strict]{changepage}
\usepackage{ragged2e}
\usepackage{microtype}
\usepackage{enumitem}
\usepackage{nccmath}
\usepackage{hyperref}
\usepackage{doi}
\usepackage{afterpage}
\usepackage{multirow,bigdelim}
\usepackage{booktabs}
\usepackage{color}
\usepackage[outline]{contour}
\usepackage{xcolor}

\setlist{parsep=0pt,listparindent=\parindent}

\DeclareCaptionFont{singlespacing}{\setstretch{1}}
\numberwithin{figure}{section}
\renewcommand{\thefigure}{\arabic{section}.\arabic{figure}}

\renewcommand \thesection{\arabic{section}}
\numberwithin{equation}{section}

\hypersetup{bookmarks=false,bookmarksopen=false,pdfpagemode=empty,pdfstartview=}

\title{\singlespacing\sf\huge An Exact Reformulation of Feature-Vector-based Radial-Basis-Function Networks for Graph-based Observations}
\author{Isaac J. Sledge, \emph{Member, IEEE}, and Jos\'{e} C. Pr\'{i}ncipe, \emph{Life Fellow, IEEE}%
\thanks{\fontdimen2\font=1.55pt Isaac J. Sledge is a Research Engineer with the Advanced Signal Processing and Automated Target Recognition Branch of the US Naval Surface Warfare Center, Panama City, FL 32407, USA (email: isaac.j.sledge@navy.mil).}  
\thanks{\fontdimen2\font=1.55pt Jos\'{e} C. Pr\'{i}ncipe is the Don D. and Ruth S. Eckis Chair and Distinguished Professor with both the Department of Electrical and Computer Engineering and the Department of Biomedical Engineering, University of Florida, Gainesville, FL 32611, USA (email: principe@ufl.edu).  He is the director of the Computational NeuroEngineering Laboratory (CNEL) at the University of Florida.\vspace{0.15cm}}
\thanks{The work of the authors was funded via grants N00014-15-1-2103, N00014-14-1-0542, and N00014-19-WX-00636 from ONR and ILIR grant N00014-19-WX-00687 from ONR.  The first author was also supported by a University of Florida Research Fellowship, a Robert C. Pittman Research Fellowship, and an ASEE Naval Research Enterprise Fellowship.}%
}
\begin{document}
\bstctlcite{IEEEexample:BSTcontrol}

\maketitle
\RaggedRight\parindent=1.5em
\fontdimen2\font=2.1pt
\vspace{-1.55cm}\begin{abstract}\normalsize\singlespacing
\vspace{-0.25cm}{\small{\sf{\textbf{Abstract}}}}---Radial-basis-function networks are traditionally defined for sets of vector-based observations.  In this short paper, we reformulate such networks so that they can be applied to adjacency-matrix representations of weighted, directed graphs that represent the relationships between object pairs.  We re-state the sum-of-squares objective function so that it is purely dependent on entries from the adjacency matrix.  From this objective function, we derive a gradient descent update for the network weights.  We also derive a gradient update that simulates the repositioning of the radial basis prototypes and changes in the radial basis prototype parameters.  An important property of our radial basis function networks is that they are guaranteed to yield the same responses as conventional radial-basis networks trained on a corresponding vector realization of the relationships encoded by the adjacency-matrix.  Such a vector realization only needs to provably exist for this property to hold, which occurs whenever the relationships correspond to distances from some arbitrary metric applied to a latent set of vectors.  We therefore completely avoid needing to actually construct vectorial realizations via multi-dimensional scaling, which ensures that the underlying relationships are totally preserved.

\end{abstract}%
\begin{IEEEkeywords}\normalsize\singlespacing
\vspace{-1.25cm}{{\small{\sf{\textbf{Index Terms}}}}---Graph, dissimilarity data, dissimilarity matrix, graph-based classification, graph signal processing, graph neural network, radial-basis-function networks}
\end{IEEEkeywords}
\IEEEpeerreviewmaketitle
\allowdisplaybreaks
\singlespacing

\vspace{-0.4cm}\subsection*{\small{\sf{\textbf{1$\;\;\;$Introduction}}}}\addtocounter{section}{1}


Conventional radial-basis-function (RBF) networks have a feed-forward architecture that consists of two layers: a non-linear hidden layer followed by a linear output layer.  The hidden-layer processing elements operate on the weighted distance between a vector observation and some other vector, which is referred to as either an RBF prototype or an RBF center.  The RBF prototypes specify the position of a local receptive field. The response of each processing element in this network layer is a non-linear, radially-symmetric function of this observation-prototype distance.  The hidden-layer responses are then weighted and, usually, linearly combined by at the output layer.  

These networks, as conceived by Broomhead and Lowe \cite{BroomheadDS-jour1988a}, rely on a vector-based paradigm: they are applied solely to feature-based vectorial observations \cite{BillingsSA-jour2007a}.  Here, we provide a reformulation of RBF networks so that they can be applied to adjacency-matrix representations of weighted, directed graphs.  The adjacency representations are symmetric, positive, anti-reflexive matrices of relationship-based observations between object pairs.  Such types of observations are prevalent in a number of problem domains, as investigators may be unable to extract meaningful features about various observations yet can easily codify the relationships between them.  Pertinent examples include assessing shape similarity and quantifying the relationship between gene ontology products.

The graph-based RBF networks that we consider have a feed-forward architecture analogous to that of vector-based RBF networks.  That is, the hidden layer non-linearly transforms the weighted relationship-based observations while the output layer weights and linearly combines those transformed results.  There are, however, differences, particularly with the initial hidden layer: the space of RBF prototypes exists outside the space of relational observations.  This, seemingly, implies that we cannot directly describe traditional radial basis prototypes using information from the relational observations.  We show, however, that an alternate representation, which we refer to as relational radial basis prototypes, enables the direct calculation of the RBF observation-prototype distances.  The observation-prototype distances for both networks are guaranteed to be the same under certain general conditions, implying that the vector- and graph-based networks will produce the same results and are thus duals of each other. 



There are two central issues that must be addressed so that graph-based RBF networks can be applied for practical purposes.  The first involves measuring the network error against the desired response according to some objective function.  The second entails modifying an existing training scheme so that the network parameters can be adjusted.

Toward this end, we present an objective function for RBF networks that is suitable for relationship-based observations.  It is a version of the standard sum-of-squared-errors expression where the prototype-observation distance used by the kernel function is replaced with an equivalently computable distance that can be found using properties of the graph edge weights.  We make a relatively mild assumption on the form of the RBF centers used in the prototype-observation distance calculation.  As a byproduct of this assumption, we are guaranteed that our objective function will yield the same responses as the sum-of-squared-errors objective function for vector-based RBF networks applied to an associated set of features.  There are two advantages to this.  First, a vector realization does not need to be constructed, which sidesteps needing to know the underlying metric and performing multi-dimensional scaling \cite{BorgI-book2005a} to obtain a vector realization of the graph.  Second, it permits analysis of various prototype properties, which allows for direct comparisons of the networks.  





With regards to training, there are three major types of approaches for positioning basis prototypes: heuristic \cite{BillingsSA-jour1995a,ChenS-jour1999a,WhiteheadBA-jour2002a,ChenS-jour2010a}, unsupervised \cite{ChenS-jour1991a,MusaviMT-jour1992a,GommJB-jour2000a}, and supervised \cite{HaganMT-jour1994a,SherstinskyA-jour1996a,MaoKZ-jour2002a,AmpazisN-jour2002a,ChenS-jour2008a,HongX-jour2008a,DuD-jour2010a} basis function prototype selection strategies.  Hybrid approaches \cite{MoodyJ-jour1989a,LeeS-jour1991a,PengJX-jour2006a,PengJX-jour2007a}, which combine elements from unsupervised and supervised strategies, could additionally be employed.  There are also many supervised and unsupervised schemes for choosing the network weights \cite{KarayiannisNB-jour1999a,KarayiannisNB-jour2003a}.  Here, we consider a hybrid, gradient-descent-based strategy for training.  This procedure is different from the conventional vector version, as we are unable to directly modify the positions of the latent basis prototypes due to the purely graph-based nature of the observations.  We instead simulate positional shifts in the latent basis prototypes and changes to the kernel parameters by suitably altering the latent RBF prototype-observation distance for each output response error.  This altered distance is then used in the gradient descent updates for the network weights and RBF parameters.  It is additionally used in any future passes through the networks to produce the network response.  If an associated vector-based realization of the graph provably exists, then the supervised training of the graph-based RBF networks is guaranteed to provide the same network parameters as a network trained on the vectorial observations.  There is no need to construct this realization, which mitigates perturbing distance-based relationships in a way that can potentially impact classification performance.

The remainder of this short paper is organized as follows.  We cover the reformulation of RBF networks in section 2.  In section 2.1, we consider an objective function that can be used for measuring the prediction capabilities of vector-based RBF networks.  We assume, as part of this objective function, that the radial basis prototypes are defined by arbitrary weighted averages of a set of vector observations.  We then modify this objective function so that it solely utilizes entries from the adjacency matrices.  In section 2.2, we provide gradient-descent-based updates for tuning the network weights and the prototype-observation distances.  In section 3, we provide simulation results and analyses on graph-only datasets.  We illustrate, for purely graph-based datasets, that computing a vectorial representation can significantly disrupt the class statistics; it can also undermine classification when using vector-based RBF networks.  Our graph-based RBF networks operate directly on the original graph representation, though, and thus achieve superior performance.  In the appendix, we show that our graph-based RBF networks provide the same responses as conventional RBF networks applied to an error-free vectorial realization of these graphs.  These results empirically establish the networks are duals of each other.  We summarize our contributions in section 4.

\subsection*{\small{\sf{\textbf{2$\;\;\;$Graph-based RBF Networks}}}}\addtocounter{section}{1}


In this section, we provide a reformulation of the conventional RBF network architecture for graph-based data.  Such a reformulation is necessary for relationship-based observations, as the notion of an RBF prototype does not directly exist.  Without it, we would be unable to directly assess the observation-prototype distance and hence propagate a graph's adjacency matrix through a network to obtain either a classification or regression response. 

To address this problem, we could have each prototype come from the graph observations.  That is, we would view each of these observations as a type of medoid: an observation which is the least dissimilar to other observations\\ \noindent for a given neighborhood.  This would permit finding the observation-prototype distance by a simple matrix indexing scheme.  It would also be relatively straightforward to iteratively add more medoids to improve the network accuracy, akin to \cite{KaminskiW-jour1997a}.  However, for certain datasets, relying on medoids may yield sub-par graph-based RBF networks.  This is because RBF prototypes may need to exist outside the set of graph observations to obtain accurate networks.  An unnecessarily large number of basis functions may also be needed to achieve adequate performance during training.  While it might be possible to sometimes counteract this issue by choosing suitable weights, we would like to avoid completely relying on this behavior to produce high-performing networks.

We have hence opted to provide an exact reformulation of vectorial RBF networks.  We systematically derive an equivalent version of the vector-data sum-of-squares objective function that relies only on quantities that can be obtained from graph observations.  More specifically, our derivation is based on algebraically manipulating the objective function using properties of general distance metrics.  For a pre-assumed form of the RBF prototypes, we show that the prototype-observation distance used by the radial basis kernels can be written in terms of entries from an adjacency matrix associated with a weighted, directed graph.  For both vector- and graph-based data, the prototype-observation distance expressions are equivalent under mild, often generally satisfiable conditions.  The pre-supposed form of the RBF prototypes is also quite general and permits them to be linear combinations of the observations.

\begin{figure}[t!]
\hspace{-0.2in}\begin{tabular}{c c c}
\includegraphics[width=2.15in]{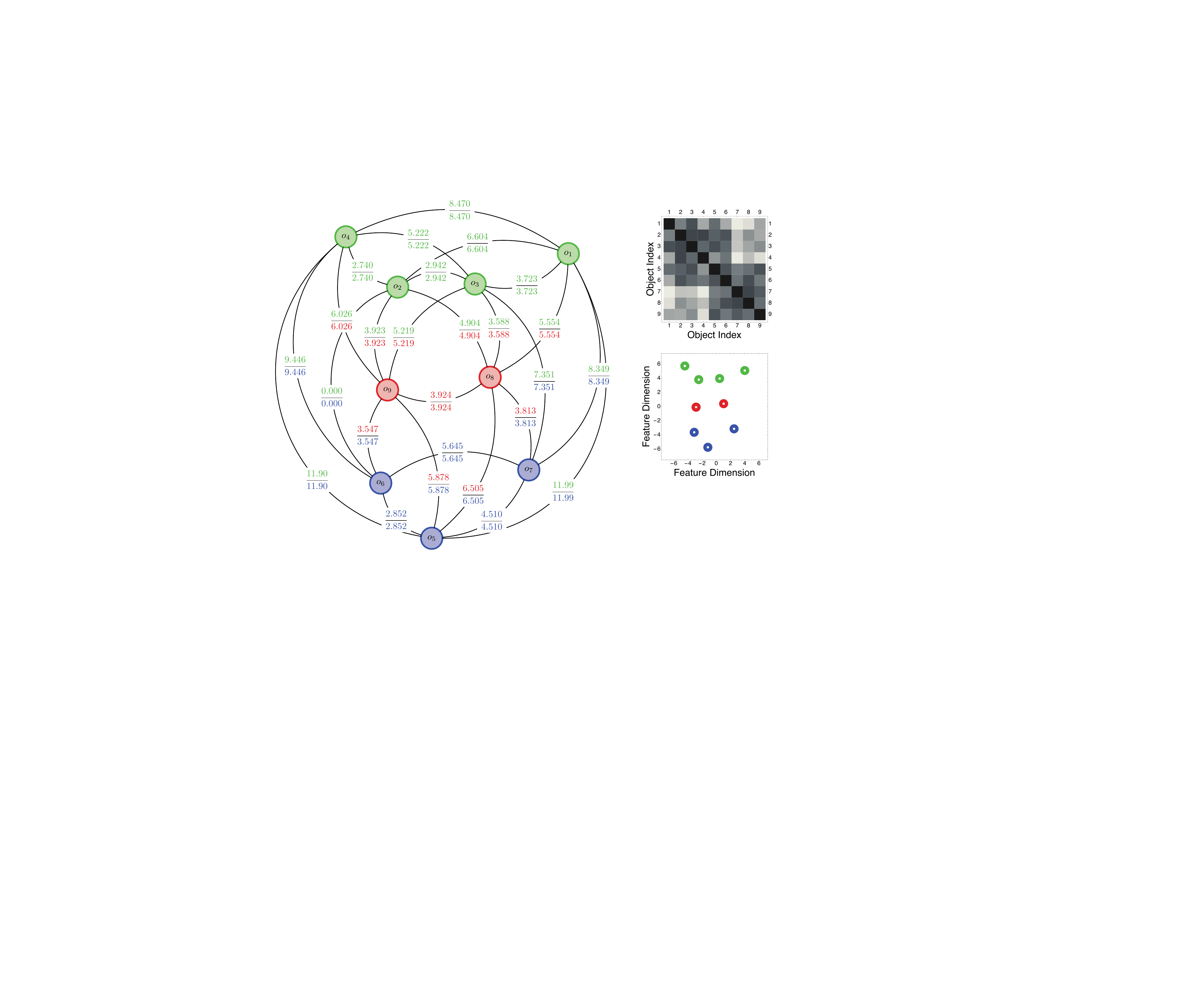} & \hspace{-0.1in} \includegraphics[width=2.15in]{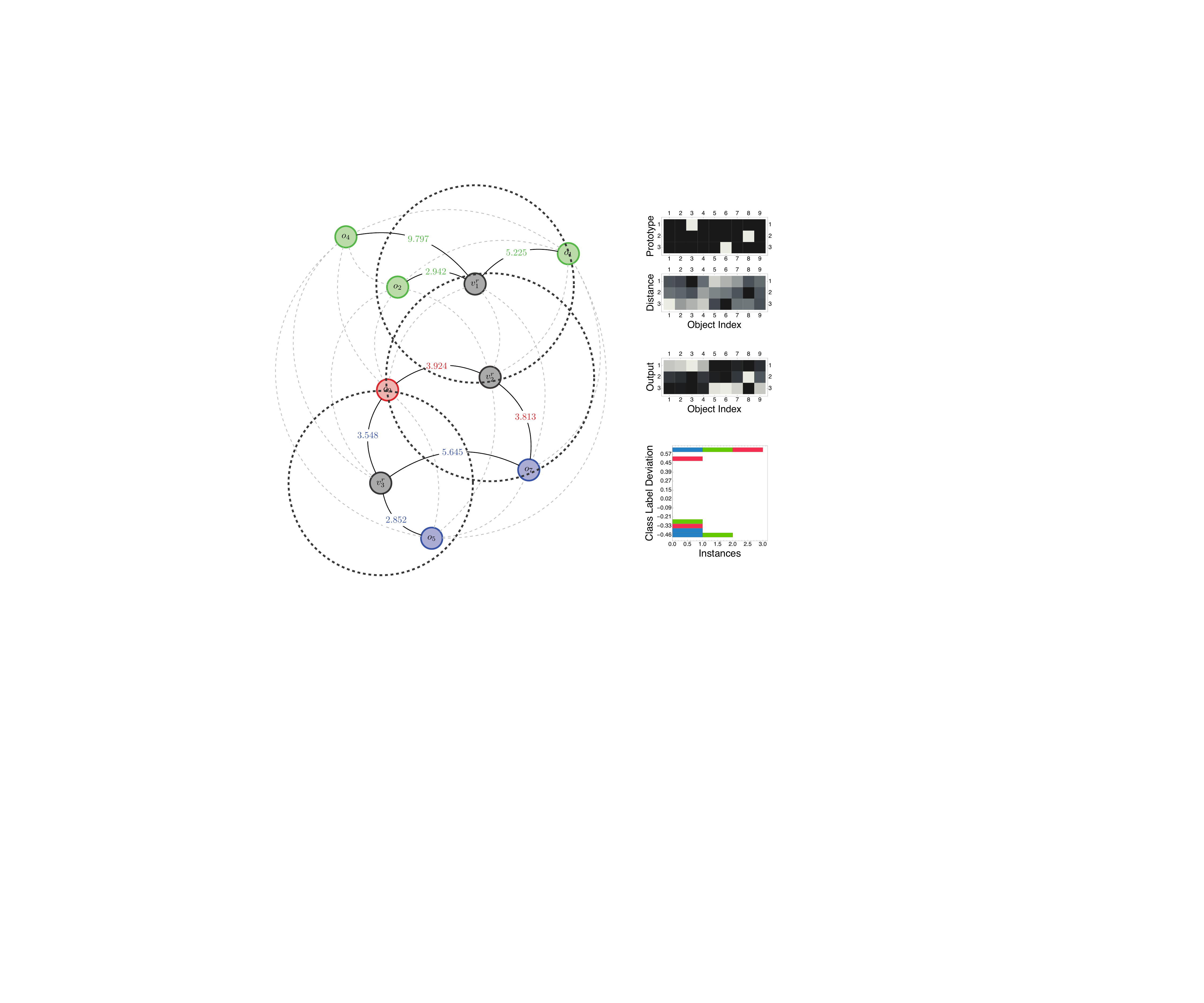} & \hspace{-0.1in} \includegraphics[width=2.15in]{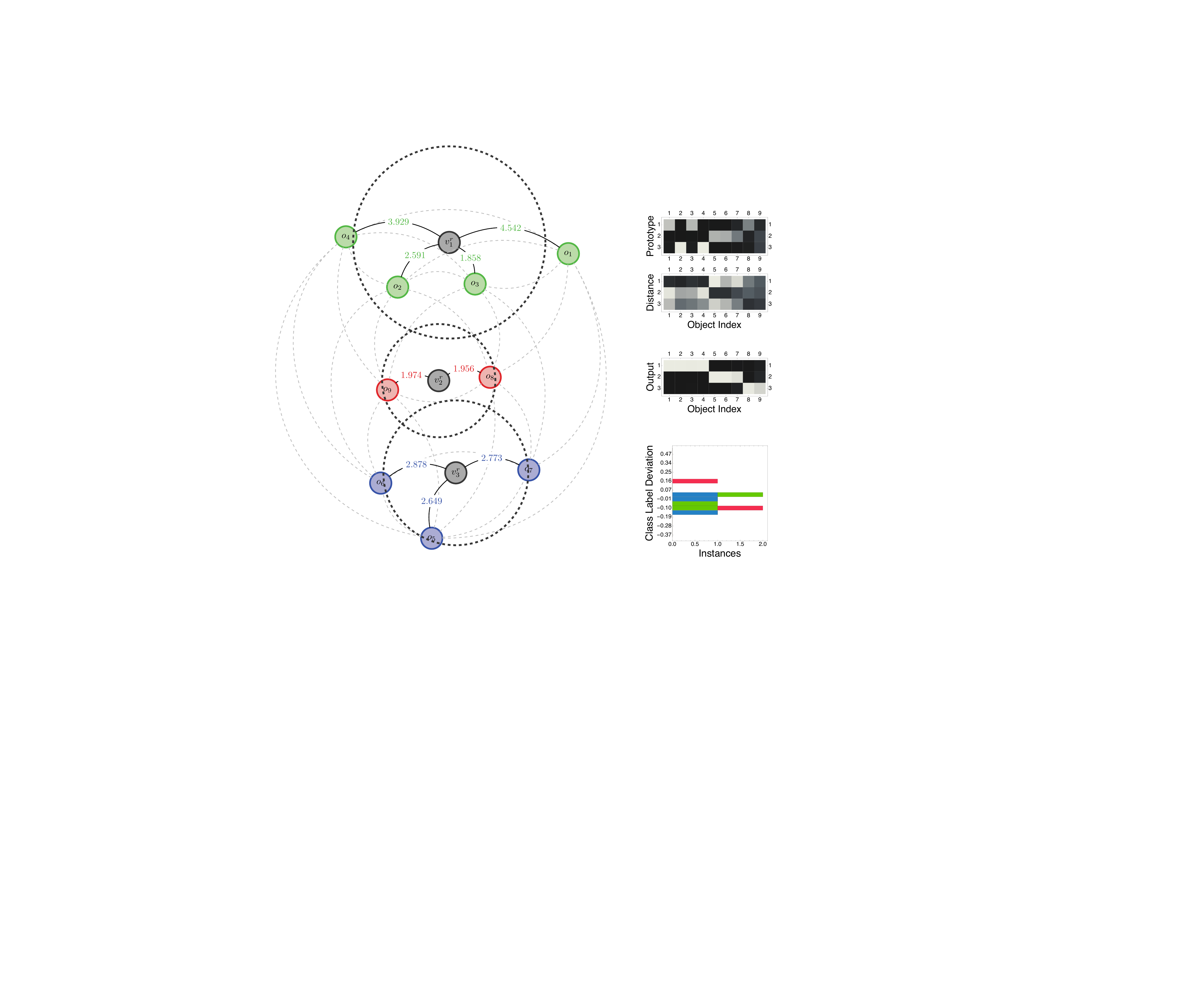}\vspace{-0.1in}\\
\hspace{-0.3in}{\footnotesize (a)} & \hspace{-0.3in}{\footnotesize (b)}  & \hspace{-0.2in}{\footnotesize (c)}
\end{tabular}
\caption[]{\fontdimen2\font=1.8pt\selectfont Illustration of the advantages to using relational RBF prototypes instead of medoids.  In (a), we show a weighted, directed graph constructed from a set of feature vectors that correspond to a three-class learning problem.  The associated adjacency matrix and two-dimensional feature-space realization are shown, respectively, in the top-right corner.  In (b), we show a locally optimal medoid placement for this dataset, along with the ninety-percent support interval for a Gaussian kernel.  When using medoids, the RBF prototypes are constrained to be nodes in the graph.  This often has the disadvantage of misclassifying objects near the class boundaries.  For instance, in this example, both $o_7$ and $o_9$ are misclassified.  This was, in part, because none of the graph nodes captured the class average well.  Large receptive fields were hence needed to cover the classes, which unnecessarily shifted the decision boundaries into other regions of the graph.  Including more prototypes would preempt this occurrence but would lead to non-parsimonius models.  When the RBF prototypes are allowed to exist outside of the graph representation, as shown in (c), then these types of misclassifications are often significantly reduced.  This can be seen from both the model output plots and the error histogram plots in the bottom, right corners of (b) and (c); here, blue, green, and red denote, respectively, the training, testing, and validation set samples.  Both $o_7$ and $o_9$ are properly classified in (c), since the nodes in the graph could be covered well with only a few prototypes with small activation regions.\vspace{-0.4cm}} 
\end{figure}

This reformulation permits the prototypes to exist outside the graph observation space, fixing the aforementioned issues that would be encountered when medoids are used; see figure 2.1.  Moreover, provided that the graph observations correspond to distances from some arbitrary metric, our graph-based RBF network is guaranteed to yield the same result as a traditional RBF network applied to vectorial realization of the adjacency matrix.  Theoretical results obtained for the feature-vector-based networks thus automatically transfer to the graph-based networks.  

Note that we do not need to actually compute a realization of weighted graphs, by way of multi-dimensional scaling, to apply RBF networks this data modality.  This turns out to be a major advantage, as it may be difficult to not only determine the underlying metric for purely graph observations, but also produce an error-free realization.  If multi-dimensional scaling is applied to arbitrary graph-based datasets, then it is possible that the statistics of each class can change, potentially in a way that undermines classification performance \cite{SledgeIJ-conf2010a}.  We show this is the case in our simulations, which causes the classification performance to noticeably suffer.

These are not the only ways in which a graph-based RBF network can be trained.  Another natural option would be to pose the entire training process in a reproducing-kernel Hilbert space.  The disadvantage of doing this, though, is that we would lose the ability to directly analyze the relational RBF prototypes and hence easily compare multiple graph-based RBF networks.  We may also lose the duality property between graph and vector-based RBF networks.  Much of the theoretical properties of vector-based RBF networks would hence need to be re-derived for graph-based networks.



\subsection*{\small{\sf{\textbf{2.1$\;\;\;$Assessing Graph-based Radial Basis Network Performance}}}}


We begin our derivation of the graph-based RBF network objective function by considering the sum-of-squared-error cost used by conventional RBF networks for vector observations.   We view the prototype-observation distance in this objective function as an inner product.  For this inner product, we assume that the RBF prototypes are specified as weighted versions of the feature-space observations.\vspace{0.05cm}
\begin{itemize}
\item[] \-\hspace{0.0cm}{\small{\sf{\textbf{Definition 2.1.}}}} Let $x_i \!\in\! \mathbb{R}^d$, $i \!=\! 1,\ldots,n$, be a set of vectorial feature space observations.  Let $v_j \!\in\! \mathbb{R}^d$,\\ \noindent $j \!=\! 1,\ldots,c$, be a series of vectorial RBF prototypes, represented by $v_j \!=\! \sum_{p=1}^n u_{j,p}x_p/\sum_{r=1}^n u_{j,r}$ where\\ \noindent $u_{j,i} \!\in\! \mathbb{R}$, $j \!=\! 1,\ldots,c$, $i \!=\! 1,\ldots,n$, is a weight term.  The sum-of-squared-error objective function for a\\ \noindent two-layer vectorial RBF network with $g$ output-layer units is given by
\begin{equation}
\sum_{i=1}^n\sum_{k=1}^g \Bigg(y_{i,k} \!-\! f_2\Bigg(w_{0,k} + \sum_{j=1}^c w_{j,k} f_1(d_{j,i})\!\Bigg)\!\Bigg)^{\!2},\;\;\;\;\; d_{j,i} \!=\! \left\langle\! x_i \!-\! \left.\sum_{p=1}^n u_{j,p}x_p\!\right/\!\sum_{r=1}^n u_{j,r}, \cdot \right\rangle
\end{equation}
where $f_1,f_2 : \mathbb{R} \!\to\! \mathbb{R}$ are continuous, differentiable mappings, $w_0,w_j \!\in\! \mathbb{R}$ are scalar weights, and $y_{i,k} \!\in\! \mathbb{R}$ is a desired output response.  The aim is to minimize this expression subject to $w_{0,k},w_{j,k} \!\in\! \mathbb{R}$ and $v_j \!\in\! \mathbb{R}^d$. \vspace{0.05cm}
\end{itemize}
The assumption that we make concerning the form of the RBF prototypes is crucial: it would be difficult to re-state the objective function purely in terms of the adjacency matrix without it.  This assumption is not, however, restrictive: the RBF centers can be linear combinations of the vectors and hence can span them.


In (2.1), we have two everywhere differentiable mappings.  The mapping $f_1 : \mathbb{R} \!\to\! \mathbb{R}$ is the radial basis function, which is weighted by $w_j \!\in\! \mathbb{R}$ and offset from the origin by $w_0 \!\in\! \mathbb{R}$.  A typical choice for the basis function is the\\ \noindent Gaussian kernel, as it defines a reproducing kernel Hilbert space.  Popular alternatives include fractional and non-fractional polynomial kernels.  The mapping $f_2 : \mathbb{R} \!\to\! \mathbb{R}$ is used to produce the output response for the network.  For the purposes of learning, the network response is compared to the desired response $y_{i,k} \!\in\! \mathbb{R}$, where $i \!=\! 1,\ldots,n$\\ \noindent denotes the observation index and $k \!=\! 1,\ldots,g$ the corresponding response index.  This second mapping is typically a linear activation function.



We now demonstrate how the vector-based cost function in (2.1) can be algebraically reformulated, using properties of general distance metrics, to rely solely on entries from the adjacency matrix of a weighted, directed graph.  This leads to an equivalent expression for the prototype-observation distance used by the radial basis kernels.  Toward this end, we first factor out the common normalizing term from the inner product in (2.1), 
\begin{equation}
\sum_{i=1}^n\sum_{k=1}^g \Bigg(y_{i,k} \!-\! f_2\Bigg(w_{0,k} + \sum_{j=1}^c w_{j,k} f_1\Bigg(\!\Bigg(\sum_{i=1}^n u_{j,i}\Bigg)^{\!\!-2}\!\!\left\langle\, \sum_{p=1}^n u_{j,p}(x_i \!-\! x_p),\; \sum_{q=1}^n u_{j,q}(x_i \!-\! x_q) \right\rangle\!\!\Bigg)\!\Bigg)\!\Bigg)^{\!2}. 
\end{equation}
This normalizing term forms part of the expression for a relational version of the RBF prototype, as we will demonstrate shortly.  For now, we will consider it as a scalar weighting factor.


Next, we expand the terms inside of the inner product for (2.2) and perform a bit of manipulation.  This allows us to begin re-writing the objective function in terms of entries $r_{p,q}$, $r_{p,i}$, $r_{q,i} \!\in\! \mathbb{R}$ from the adjacency matrix $R \!\in\! \mathbb{R}^{n \times n}_{0,+}$\\ \noindent and weights $u_{j,p}$, $u_{j,q} \!\in\! \mathbb{R}$.  We assume that the entries of the adjacency matrix correspond to pairwise distances\\ \noindent between latent vector observations for some potentially unknown metric.  The objective function hence becomes
\begin{equation}
\sum_{i=1}^n\sum_{k=1}^g \Bigg(y_{i,k} \!-\! f_2\Bigg(w_{0,k} + \sum_{j=1}^c w_{j,k} f_1\Bigg(\!\Bigg(4\sum_{i=1}^n u_{j,i}\Bigg)^{\!\!-2}\!\Bigg(\sum_{p=1}^n\sum_{q=1}^n u_{j,p} u_{j,q}(r_{p,q} \!-\! r_{p,i} \!-\! r_{q,i})\Bigg)\!\Bigg)\!\Bigg)\!\Bigg)^{\!2}.
\end{equation}
The form of (2.3) is suitable for specifying a graph-based RBF network.  It is equivalent to (2.1), provided that the entries of the adjacency matrix correspond to a distance metric applied to a latent set of vector observations.  We can, however, further simplify the expression in (2.3) by factoring out the relational entries.  This yields the following minimization problem, 
\begin{equation}
\sum_{i=1}^n\sum_{k=1}^g \Bigg(y_{i,k} \!-\! f_2\Bigg(w_{0,k} + \sum_{j=1}^c w_{j,k} f_1\Bigg(\!\Bigg(4\sum_{i=1}^n u_{j,i}\Bigg)^{\!\!-2}\!\Bigg(\sum_{p=1}^n u_{j,p}(e_i \!-\! e_p)^\top \!\Bigg) R \Bigg(\sum_{q=1}^n u_{j,q} (e_i \!-\! e_q)\Bigg)\!\Bigg)\!\Bigg)\!\Bigg)^{\!2}, 
\end{equation}
where $e_k$, $e_p$, $e_q \!\in\! \mathbb{R}_{0,+}^n$ are unit basis vectors.

As a final step, we specify the notion of a relational RBF prototype $v_j^r \!\in\! \mathbb{R}^n$.  These relational prototypes are\\ \noindent vectors of weights that are normalized by the total weight for a given feature-space RBF prototype.  They are not prototypes in the traditional sense, however.  In the feature space, they capture the contribution of each observation toward the weighted average that defines a particular prototype.  In the graph space, the relational prototypes serve a similar purpose, as they determine how much a given relationship influences a latent, weighted average of the vectorial observations in $\mathbb{R}^d$.  


With this notion, we can update the objective function in (2.4) as follows.\vspace{0.05cm}
\begin{itemize}
\item[] \-\hspace{0.0cm}{\small{\sf{\textbf{Definition 2.2.}}}} Let $R \!\in\! \mathbb{R}^{n \times n}_{0,+}$ be a symmetric, non-negative, anti-reflexive adjacency matrix of a weighted,\\ \noindent directed graph.  The sum-of-squared-error objective function for a graph-based RBF network is
\begin{equation}
\sum_{i=1}^n\sum_{k=1}^g \Bigg(y_{i,k} \!-\! f_2\Bigg(w_{0,k} + \sum_{j=1}^c w_{j,k} f_1(d_{j,i})\!\Bigg)\!\Bigg)^{\!2},\;\;\;\;\; d_{j,i} \!=\! (R(v_j^r))_i \!-\! (v_j^r)^\top\! R (v_j^r)/2 
\end{equation}
where $v_{j,i}^r \!=\! u_{j,i}/\sum_{q=1}^n u_{j,q}$ for weights $u_{j,i}$, $u_{j,q} \!\in\! \mathbb{R}$.  Here, $f_1,f_2 : \mathbb{R} \!\to\! \mathbb{R}$ are continuous, differentiable\\ \noindent functions and $w_{0,k}$, $w_{j,k} \!\in\! \mathbb{R}$ are scalar weights.\vspace{0.05cm}
\end{itemize}
We utilize the objective function in (2.5) to define a procedure for uncovering the parameters of a graph-based RBF network.  These parameters include the weights for each layer $w_0$, $w_j \!\in\! \mathbb{R}$ and relational prototypes $v_j^r \!\in\! \mathbb{R}^n_{0,+}$.  They\\ \noindent can also include parameters associated with the radial basis function, such as the kernel bandwidth.

It is important to notice that the relational objective function (2.5) is an exact reformulation of the original feature-space objective function (2.1).  The validity of this claim can be seen from the above derivations; a formal proof is provided in the appendix.\vspace{0.05cm}
\begin{itemize}
\item[] \-\hspace{0.0cm}{\small{\sf{\textbf{Proposition 2.1.}}}} Let $R \!\in\! \mathbb{R}^{n \times n}_{0,+}$ be a symmetric, non-negative, anti-reflexive adjacency matrix of a weighted, di-\\ \noindent rected graph such that entries $r_{p,q} \!\in\! \mathbb{R}_{0,+}$ correspond to distances from a metric applied to pairs of latent vectors\\ \noindent $x_p,x_q \!\in\! \mathbb{R}^d$.  For $v_j \!=\! \sum_{p=1}^n u_{j,p}x_p/\sum_{r=1}^n u_{j,r}$ and $v_{j,i}^r \!=\! u_{j,i}/\sum_{q=1}^n u_{j,q}$, $u_{j,i}$, $u_{j,q} \!\in\! \mathbb{R}$, the objective\\ \noindent function given in (2.1) is equivalent to that in (2.5).

Alternatively, let $R \!\in\! \mathbb{R}^{n \times n}_{0,+}$ be a matrix such that it is non-negative, $r_{i,j} \!\geq\! 0$, symmetric, $r_{i,j} \!=\! r_{j,i}$, anti-\\ \noindent reflexive, $r_{i,i} \!=\! 0$, and where the entries obey the triangle inequality $r_{i,k} \!\leq\! r_{i,j} \!+\! r_{j,k}$, $\forall i,j,k$.  A latent vector\\ \noindent realization exists and the objective function given in (2.1) is equivalent to that in (2.5).\vspace{0.05cm}
\end{itemize}
\noindent It is important to note that no approximations have been made when reformulating (2.1) to (2.5).  Many of the theories developed for feature-space RBF networks automatically apply to graph-based RBF networks as a consequence of this property.  For example,we immediately obtain that graph-based RBF networks are universal approximators when using adjustable-variance Gaussian kernels \cite{HartmanEJ-jour1990a}.  Other results would apply when considering more general types of kernels that satisfy mild constraints \cite{ParkJ-jour1991a,ParkJ-jour1993a}.  A caveat is that these theories are only valid when a feature-space realization provably exists for a given adjacency matrix.  If a feature-space realization of a adjacency matrix does not provably exist, then the adjacency entries would need to be transformed so that it does \cite{SledgeIJ-conf2010a}.






\subsection*{\small{\sf{\textbf{2.2$\;\;\;$Training Graph-Based Radial Basis Networks}}}}

As we highlighted above, a variety of approaches exist for minimizing the sum-of-squared-error objective function.  Here, we consider a gradient-descent-based approach, as it is conducive to learning when using graph-based inputs.  There are also several practical advantages to such a supervised training approach.

It is straightforward to update the weights of the network via gradient-descent procedure; the update is given in (2.7).  More elaborate versions of gradient descent, such as quasi-second-order methods, can also be considered.\vspace{0.05cm}
\begin{itemize}
\item[] \-\hspace{0.0cm}{\small{\sf{\textbf{Proposition 2.2.}}}} Let $R \!\in\! \mathbb{R}^{n \times n}_{0,+}$ be a symmetric, non-negative, anti-reflexive adjacency matrix of a weighted, di-\\ \noindent rected graph such that entries $r_{p,q} \!\in\! \mathbb{R}_{0,+}$ correspond to distances from a metric applied to pairs $x_p,x_q \!\in\! \mathbb{R}^d$.

Suppose that the output-layer activation function of a two-layer, vector RBF network is linear.  The change in the weights $w_{0,k},w_{j,k} \!\in\! \mathbb{R}$ is given by $w_p \!\leftarrow\! w_p \!-\! \eta \sum_{i=1}^n \epsilon_{p,i}^o d_i$ and $w_p \!=\! [w_{0,p},w_{1,p},\ldots,w_{c,p}]^\top$, for some\\ \noindent positive learning rate $\eta \!\in\! \mathbb{R}_+$.  Here, the output error is $\epsilon_{p,i}^o \!=\! y_{p,i} \!-\! \overline{y}_{p,i}$ with $\overline{y}_{p,i} \!=\! w_{0,p} \!+\! \sum_{j=1}^c w_{j,p} f_1(d_{j,i})$.  As\\ \noindent  well, $d_i \!=\! [1,f_1(d_{1,i}),\ldots,f_1(d_{c,i})]^\top$ with $d_{j,i} \!=\! \langle x_i \!-\! v_j,\cdot\rangle$.  The term $\Delta w_p \!=\! \eta \sum_{i=1}^n \epsilon_{p,i}^o d_i$ in $w_p \leftarrow w_p \!-\! \eta\Delta w_p$ is hence equal to
\begin{multline}
\eta \sum_{i=1}^n\sum_{k=1}^g \Bigg(y_{i,k} \!-\! w_{0,k} \!-\! \sum_{j=1}^c w_{j,k} f_1\Bigg(\!\!\left\langle\! x_i \!-\! \left.\sum_{p=1}^n u_{j,p}x_p\!\right/\!\sum_{r=1}^n u_{j,r},\; \cdot \right\rangle\!\!\Bigg)\!\Bigg)\\ f_1\Bigg(\!\!\left\langle\! x_i \!-\! \left.\sum_{p=1}^n u_{j,p}x_p\!\right/\!\sum_{r=1}^n u_{j,r},\; \cdot \right\rangle\!\!\Bigg).
\end{multline}
The dot indicates that the first argument of the inner product is simply repeated as the second argument.  Equation (2.6) is equivalent to saying that $\Delta w_p$ is equal to
\begin{equation}
\eta \sum_{i=1}^n\sum_{k=1}^g \Bigg(y_{i,k} \!-\! w_{0,k} \!-\! \sum_{j=1}^c w_{j,k} f_1(d_{j,i})\Bigg)f_1(d_{j,i}),\;\;\;\;\;\; d_{j,i} \!=\! (R(v_j^r))_i \!-\! (v_j^r)^\top\! R (v_j^r)/2
\end{equation}
for $v_j \!=\! \sum_{p=1}^n u_{j,p}x_p/\sum_{r=1}^n u_{j,r}$ and $v_{j,i}^r \!=\! u_{j,i}/\sum_{q=1}^n u_{j,q}$, for $u_{j,i}$, $u_{j,q} \!\in\! \mathbb{R}$.  This expression is solely in\\ \noindent terms of the adjacency weights, which implies that $d_{j,i}$ in $\overline{y}_{p,i} \!=\! w_{0,p} \!+\! \sum_{j=1}^c w_{j,p} f_1(d_{j,i})$ is the reformulated\\ \noindent distance between the $j$th prototype and the $i$th feature-space observation: $d_{j,i} \!=\! (R(v_j^r))_i \!-\! (v_j^r)^\top R(v_j^r)/2$.
\end{itemize}
In deriving (2.7), we have assumed that the underlying vector relational basis prototypes will not be adjusted across gradient-descent iterations.  In practice, however, both the weights and the prototypes will be changed in an alternating fashion.  For such situations, the reformulated distance in (2.7) will hence need to be offset by some amount for each prototype update; the offset from $(R(v_j^r))_k \!-\! (v_j^r)^\top\! R (v_j^r)/2$ is given by the extra terms shown in (2.9).

\begin{figure}[t!]
\hspace{-0.2in}\begin{tabular}{c c c}
\includegraphics[width=2.15in]{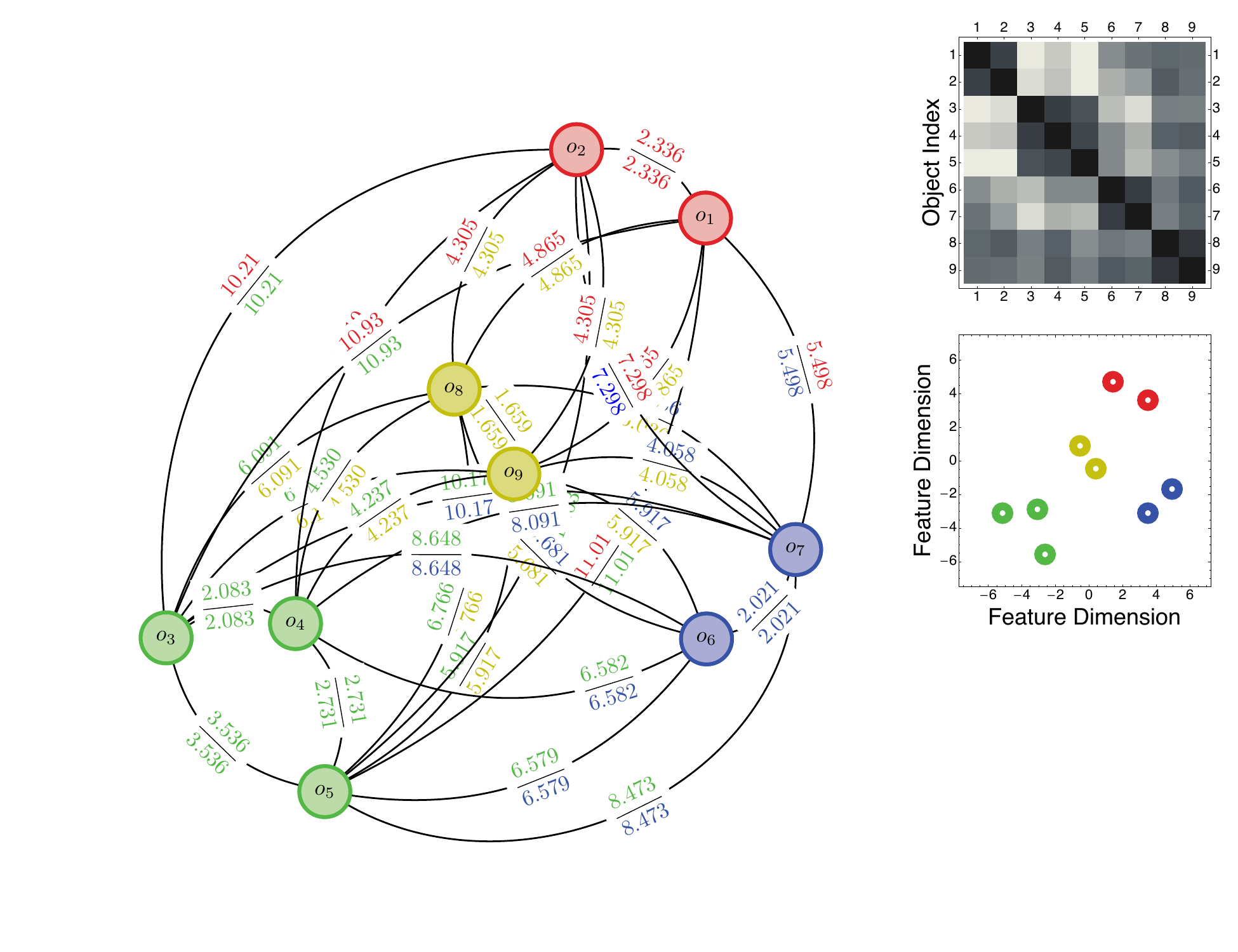} & \hspace{-0.1in} \includegraphics[width=2.15in]{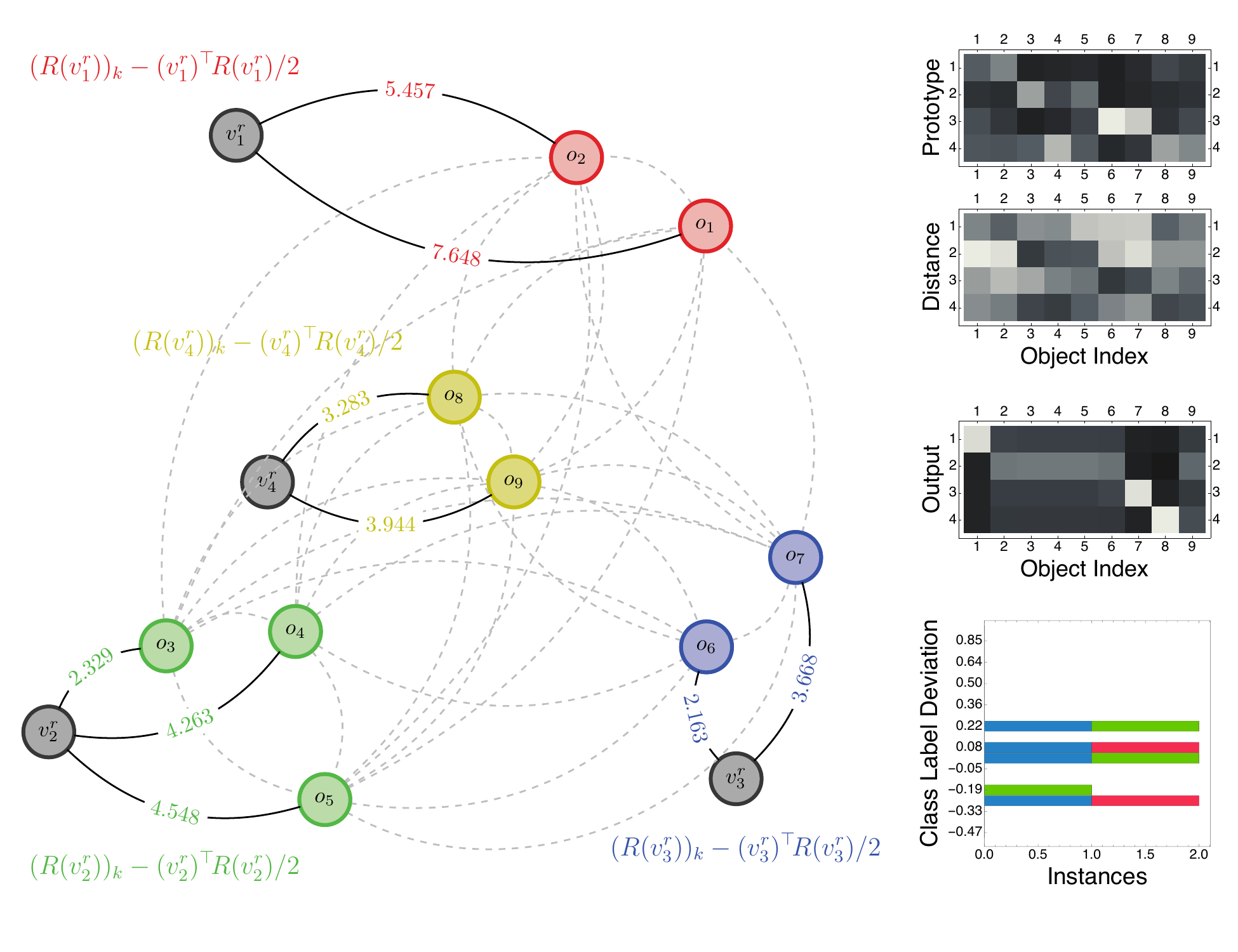} & \hspace{-0.1in} \includegraphics[width=2.15in]{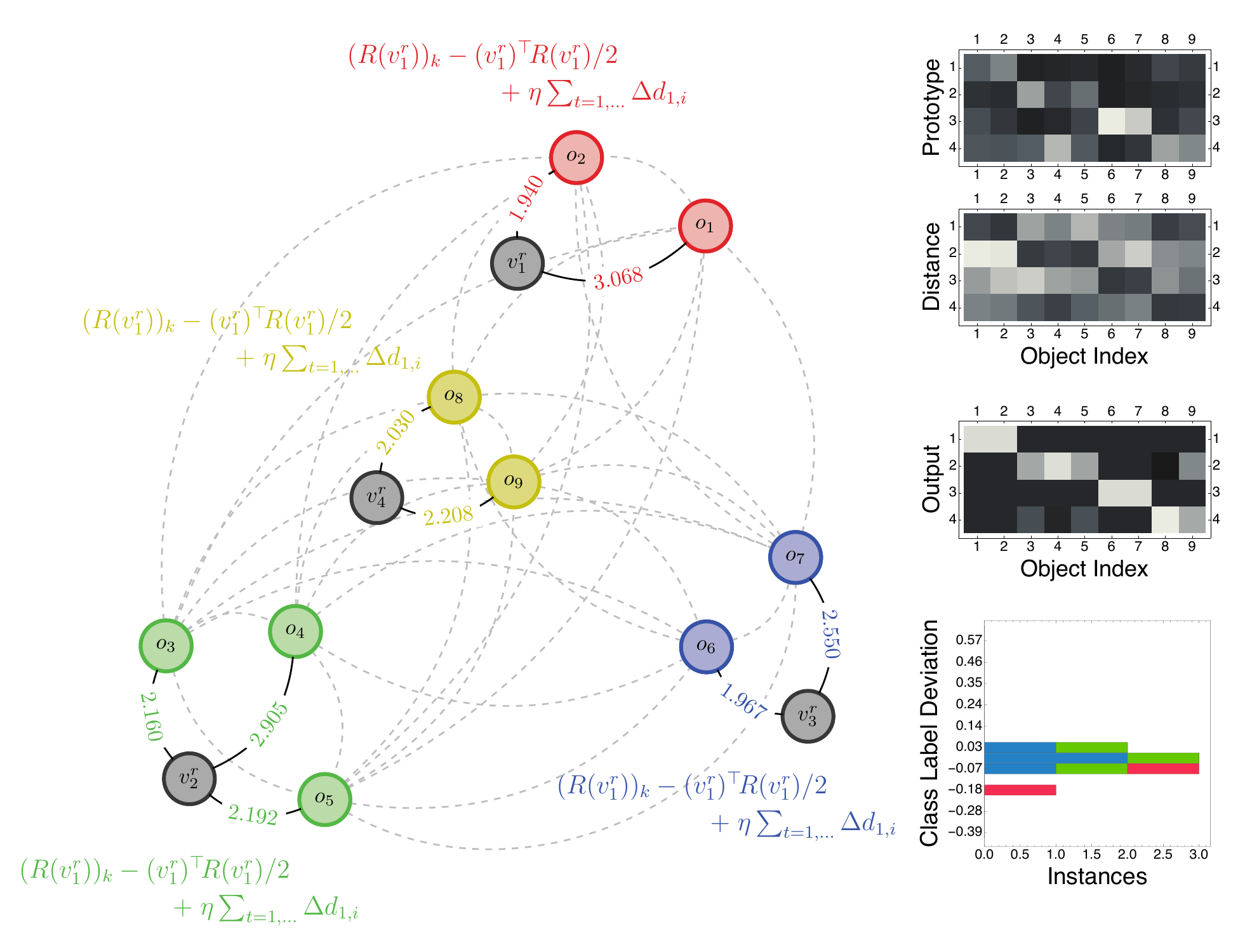}\vspace{-0.1in}\\
\hspace{-0.3in}{\footnotesize (a)} & \hspace{-0.3in}{\footnotesize (b)}  & \hspace{-0.2in}{\footnotesize (c)}
\end{tabular}
\caption[]{\fontdimen2\font=1.8pt\selectfont Illustration of a relational RBF prototype update. In (a), we show a weighted, directed graph constructed from a set of feature vectors that correspond to a four-class learning problem. The associated adjacency matrix and two-dimensional feature-space realization are shown, respectively, in the top-right corner. Each relational RBF prototype exists outside the space of the graph. These prototypes can be thought of as dummy nodes from which the metric distance to other nodes in the graph can be obtained through knowledge of the adjacency matrix and a series of weights. This is shown in (b). The corresponding relational RBF prototype, observation-prototype distance, and network response are given, respectively, in the top-right corner. Changes to the prototypes, in response to learning, can only occur by offsetting the original observation- prototype distance, as shown in (c). This has the effect of translating a corresponding feature-space RBF prototype in a latent, and potentially unknown, vector space.  This, naturally, improves the network's performance; each of the samples in the classes can be distinguished from each other.  Observe that the relational RBF prototypes remain the same across the updates.  Only the observation-prototype distance values and change, which changes the resulting network response.\vspace{-0.4cm}}
\end{figure}

\begin{figure}[t!]
\hspace{-0.2in}\begin{tabular}{c c c}
\includegraphics[width=2.15in]{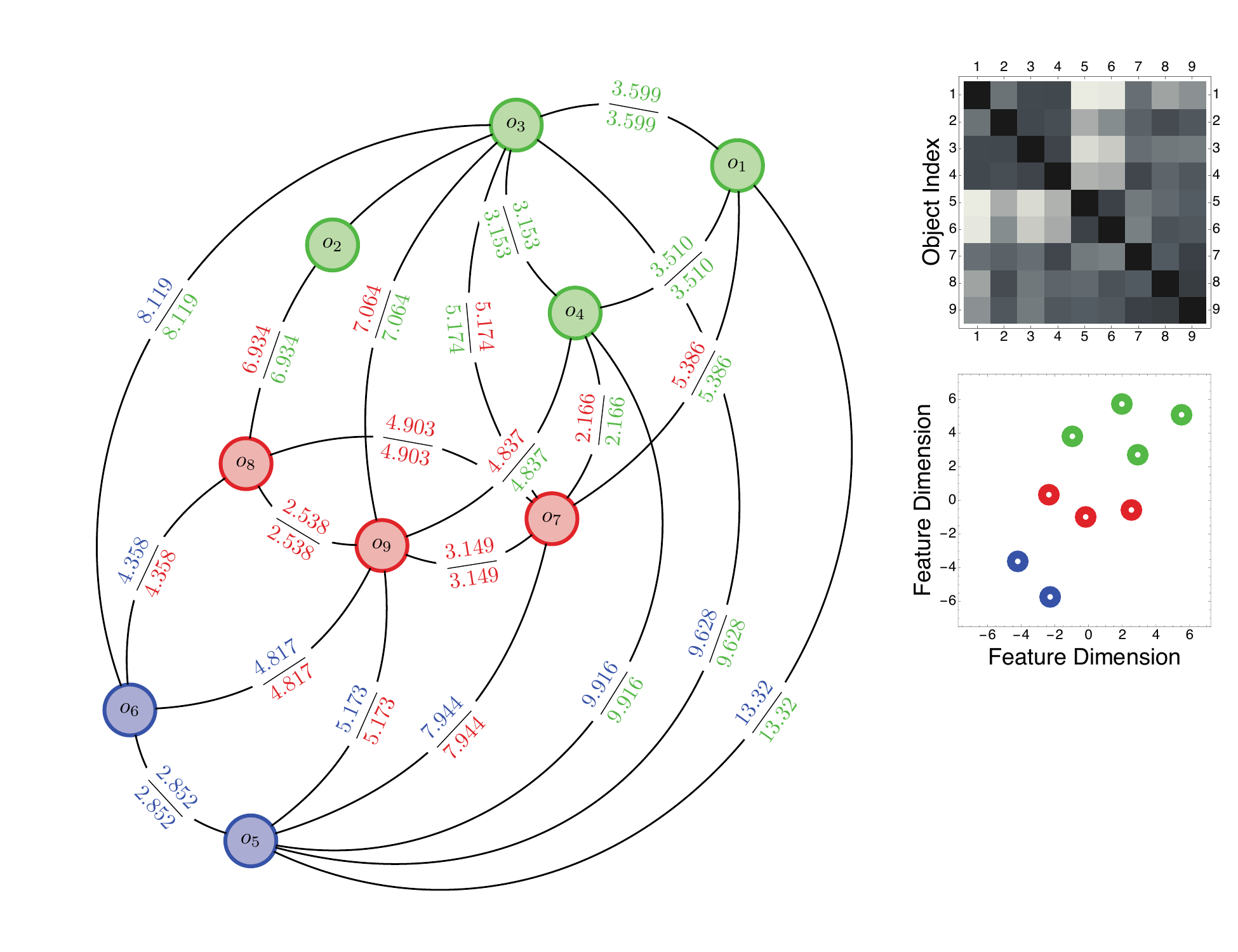} & \hspace{-0.1in} \includegraphics[width=2.15in]{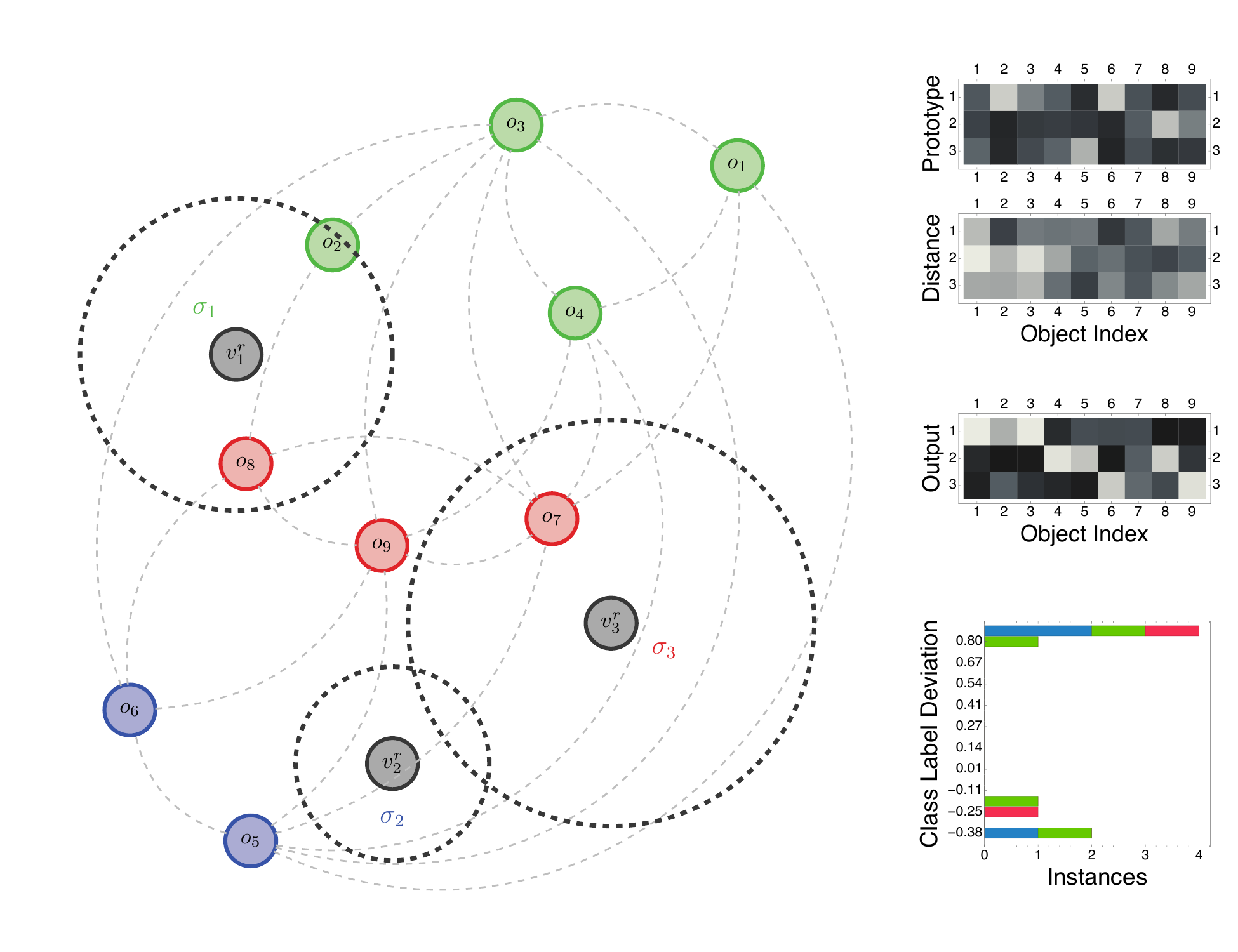} & \hspace{-0.1in} \includegraphics[width=2.15in]{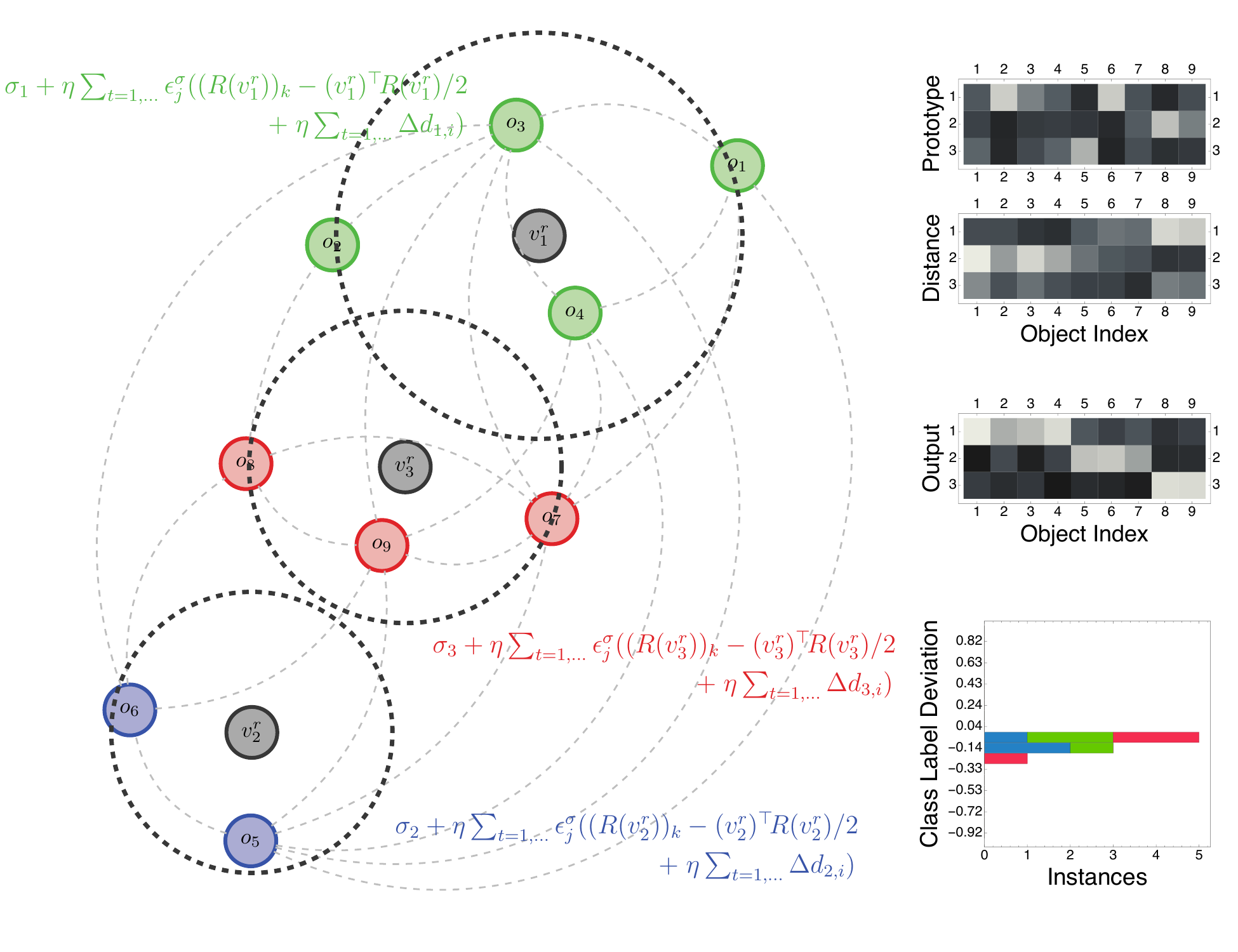}\vspace{-0.1in}\\
\hspace{-0.3in}{\footnotesize (a)} & \hspace{-0.3in}{\footnotesize (b)}  & \hspace{-0.2in}{\footnotesize (c)}
\end{tabular}
\caption[]{\fontdimen2\font=1.8pt\selectfont Illustration of a relational RBF prototype update and a kernel bandwidth update.  In (a), we show a weighted, directed graph constructed from a set of feature vectors that correspond to a three-class learning problem. The associated adjacency matrix and two-dimensional feature-space realization are shown, respectively, in the top-right corner.  As shown in (b), each relational RBF prototype is associated with initial kernel bandwidth values.  We plot ninety-percent support interval for Gaussian basis kernels with these initial bandwidth values.  The corresponding relational RBF prototype, observation-prototype distance, and network response are given, respectively, in the top-right corner. Updates to the kernel bandwidths are a function of the observation-prototype distance, as shown in (c). Each offset made to the observation-prototype distance in response to classification errors must therefore be considered when adapting the kernel bandwidth.  Updating the kernel bandwidth using a gradient-descent-based approach often leads to a decrease in the network error, which is evident when comparing the error histogram plots in the bottom, right corners of (b) and (c); here, the blue, green, and red bars denote, respectively, the training, testing, and validation set samples.  Note that as in figure 2.2, the relational prototype representation remains constant.  Only the distance, the kernel bandwidths, and the network weights, which are not shown, are changed, which influences the network response.\vspace{-0.4cm}}
\end{figure}


Compared to altering the weights, the updates for the RBF prototypes are somewhat more problematic.  This is because we do not have direct access to the vectorial prototypes, so we cannot shift them in response to misclassification errors.  One option, though, is to implicitly update the feature-space prototypes by iteratively altering the observation-prototype distance for each forward propagation through the network.  This notion can also be extended to second-order and quasi-second-order parameter update methods.\vspace{0.05cm}
\begin{itemize}
\item[] \-\hspace{0.0cm}{\small{\sf{\textbf{Proposition 2.3.}}}} Let $R \!\in\! \mathbb{R}^{n \times n}_{0,+}$ be a symmetric, non-negative, anti-reflexive adjacency matrix of a weighted, di-\\ \noindent rected graph such that entries $r_{p,q} \!\in\! \mathbb{R}_{0,+}$ correspond to distances from a metric applied to pairs $x_p,x_q \!\in\! \mathbb{R}^d$.


Suppose that the output-layer activation function of a two-layer, vector RBF network is linear.  For this network, the gradient-descent update for the $q$th radial basis prototype $v_q \!\in\! \mathbb{R}^d$ is $v_q \!\leftarrow\! v_q \!-\! \eta \sum_{k=1}^n \epsilon_{q,k}^h(x_i \!-\! v_q)$,\\ \noindent with the hidden error $\epsilon_{q,k}^h \!=\! w_{i,q} \nabla_{v_q} f_1(d_{q,k})$ for a given $x_i \!\in\! \mathbb{R}^d$.  This is prototype update is equivalent to\\ \noindent replacing the distance $d_{q,k}$ between the $q$th prototype and the $k$th feature-space vector by 
\begin{equation}
\left\langle x_i \!-\! \left.\sum_{p=1}^n u_{j,p}x_p\!\right/\!\sum_{r=1}^n u_{j,r} \!-\! \eta\Bigg(\sum_{r=1}^n \epsilon_{r,j}^h x_r \!-\! \epsilon_{r,j}^h \left.\sum_{p=1}^n u_{j,p}x_p\!\right/\!\sum_{r=1}^n u_{j,r}\Bigg),\,\cdot\, \right\rangle.
\end{equation}
Here, the term $\epsilon_{k,j}^h \!=\! w_{i,j} \nabla_{v_j} f_1(d_{k,j})$ is obtained from the response gradient $\nabla_{v_q}\! \sum_{j=1}^c w_{i,j}f_1(d_{j,k}) =$\\ \noindent $w_{i,q}\nabla_{v_q} f_1(d_{q,k})$, with $\nabla_{v_q} f_1(d_{q,k}) \!=\! 0$ for $j \!\neq\! q$.  Equation (2.8) is the same as requiring that the observation-prototype distance in (2.5), (2.7), and (2.11) be replaced by
\begin{equation}
d_{j,i} \leftarrow d_{j,i} + \left(4\sum_{r=1}^n u_{j,r}\right)^{\!\!-2}\!\!\sum_{p,q=1}^n\! u_{j,p} u_{j,q} \eta\sum_{y=1}^n \epsilon_{y,j}^h\Bigg(\! r_{i,y} \!-\! r_{i,q} \!-\! \eta \epsilon_{y,j}^h r_{p,q}\!\Bigg),
\end{equation}
for $v_j \!=\! \sum_{p=1}^n u_{j,p}x_p/\sum_{r=1}^n u_{j,r}$ and $v_{j,i}^r \!=\! u_{j,i}/\sum_{q=1}^n u_{j,q}$, for $u_{j,i}$, $u_{j,q} \!\in\! \mathbb{R}$.  This expression is solely in\\ \noindent terms of the adjacency weights.  
\vspace{0.05cm}
\end{itemize}
The proof of this claim is given in the appendix.  Notice that this feature-space prototype update is only simulated in the graph space; that is, only the distance is updated.  It is never actually computed in the feature space, as we deal only with the adjacency matrix, not the corresponding feature-space realization.


The expression given in proposition (2.7) provides a means of updating the network weights.  That in (2.9) dictates how to change the prototype-observation distance to account for implicit shifts in the radial basis prototypes.  Together, they provide a means of training graph-based RBF networks.  Additional updates, such as those for the kernel bandwidths, can be considered depending on the chosen basis function.\vspace{0.05cm}
\begin{itemize}
\item[] \-\hspace{0.0cm}{\small{\sf{\textbf{Proposition 2.4.}}}} Let $R \!\in\! \mathbb{R}^{n \times n}_{0,+}$ be a symmetric, non-negative, anti-reflexive adjacency matrix of a weighted, di-\\ \noindent rected graph such that entries $r_{p,q} \!\in\! \mathbb{R}_{0,+}$ correspond to distances from a metric applied to pairs $x_p,x_q \!\in\! \mathbb{R}^d$.

Suppose that the output-layer activation function of a two-layer, vector RBF network is linear  Suppose that the input-layer activation function is a Gaussian kernel $f_1(\cdot) \!=\! \textnormal{exp}(-\langle x_i - \sum_{p=1}^nu_{j,p}x_p/\sum_{r=1}^n u_{j,r},\cdot\rangle/2\sigma_j^2)$.  For a feature-vector-based RBF network, the gradient-descent-based change in the kernel bandwidth parameter $\sigma_j^2 \!\in\! \mathbb{R}_+$ is $\sigma_j \!\leftarrow\! \sigma_j \!-\! \eta \epsilon_j^\sigma$, where $\epsilon_j^\sigma$ is equal to
\begin{multline}
\sum_{i=1}^n\sum_{k=1}^g \Bigg(y_{i,k} \!-\! w_{0,k} \!-\! \sum_{j=1}^c w_{j,k} \textnormal{exp}\Bigg(\!\Bigg\langle\! x_i \!-\! \sum_{p=1}^n u_{j,p}x_p\!\Bigg/\!\sum_{r=1}^n u_{j,r},\cdot\Bigg\rangle\!\!\Bigg/\!2\sigma_j^2\Bigg)\!\Bigg) w_{j,k}\\
 \textnormal{exp}\Bigg(\!\Bigg\langle\! x_i \!-\! \sum_{p=1}^n u_{j,p}x_p\!\Bigg/\!\sum_{r=1}^n u_{j,r},\cdot\Bigg\rangle\!\!\Bigg/\!2\sigma_j^2\Bigg)\!\Bigg)\Bigg(\!\Bigg\langle\! x_i \!-\! \sum_{p=1}^n u_{j,p}x_p\!\Bigg/\!\sum_{r=1}^n u_{j,r},\cdot\Bigg\rangle\!\!\Bigg/\!\sigma_j^3\Bigg)\!\Bigg).
\end{multline}
Equation (2.10) is the same as
\begin{equation}
\sum_{i=1}^n\sum_{k=1}^g \Bigg(y_{i,k} \!-\! w_{0,k} \!-\! \sum_{j=1}^c w_{j,k} \textnormal{exp}(-d_{j,i}^{(t+1)}/2\sigma_j^2)\Bigg)\!\Bigg( w_{j,k} \textnormal{exp}(-d_{j,i}^{(t+1)}/2\sigma_j^2)(d_{j,i}^{(t+1)}/\sigma_j^3)\Bigg)
\end{equation}
for $v_j \!=\! \sum_{p=1}^n u_{j,p}x_p/\sum_{r=1}^n u_{j,r}$ and $v_{j,i}^r \!=\! u_{j,i}/\sum_{q=1}^n u_{j,q}$, for $u_{j,i}$, $u_{j,q} \!\in\! \mathbb{R}$.  This expression is solely in\\ \noindent terms of the adjacency weights.  
\vspace{0.05cm}
\end{itemize} 

All of these updates rely on the selection of weight terms $u_{j,i} \!\in\! \mathbb{R}$, $j \!=\! 1,\ldots,c$, $i \!=\! 1,\ldots,n$, which form the\\ \noindent relational prototypes $v_{j}^r \!\in\! \mathbb{R}^n$ used to exactly recover the prototype-observation distance.  If only one radial-basis prototype is needed to represent the observations for a given class, then $u_{j,p} \!=\! 1$ and $u_{j,q} \!=\! 1$ if $r_{p,q}$ corresponds to the relationship between objects in the same class, while $u_{j,p} \!=\! 0$ and $u_{j,q} \!=\! 0$ otherwise.  This has the effect of ensuring that a minimum prototype-observation distance is achieved for each observation in the class; in the vectorial case, the position of the prototype would correspond to the class average.  If more than one radial-basis prototype is necessary for good performance, then many viable strategies are available for choosing the weights.  The weights $u_{j,p}$ and $u_{j,q}$ can be initialized to random values for same-class observations $r_{p,q}$ and zero otherwise.  The observations for each class can also be partitioned via relational clustering algorithms, many of which would return usable weight terms; pertinent examples are described in \cite{SledgeIJ-jour2010a,SledgeIJ-jour2010b}.  

An overview of the training process is given in algorithm 1.  An illustration of the prototype-observation distance update and the kernel bandwidth update is given in figures 2.1 and 2.2, respectively.  Note that the relational RBF prototypes do not change during these updates, only the magnitude of the prototype-observation distance does.

\subsection*{\small{\sf{\textbf{3$\;\;\;$Simulations and Results}}}}\addtocounter{section}{1}

In this section, we empirically assess our graph-based RBF networks.  We illustrate that performing multi-dimensional scaling on purely graph-based data before training leads to worse classification performance than simply utilizing our graph-based RBF networks.  This finding validates our premise for performing this network reformulation and showcases its merits.  We also consider an extension of our graph-based RBF networks where the relational RBF prototypes are constrained to be medoids and compare the performance against the original version.  This is done to highlight the appropriateness of our exact reformulation.  In the appendix, we verify the reformulation theory by showing that graph-based RBF networks applied to weighted graphs derived from a set of feature vectors provide equivalent responses as vector-based RBF networks trained on those same feature vectors.  

\begin{figure*}[t!]
\vspace{-0.6cm}
{\singlespacing\begin{algorithm}[H]
\DontPrintSemicolon
\SetAlFnt{\small} \SetAlCapFnt{\small}
\caption{Graph-based Radial Basis Network Training}
\AlFnt{\small} {\bf Inputs}: $R \!\in\! \mathbb{R}^{n \times n}$,\! a\! non-negative,\! symmetric,\! anti-reflexive\! adjacency\! matrix; $y \!\in\! \mathbb{R}^{n \times g}$,\! a\! matrix\! of\! desired\! outputs.\;
\AlFnt{\small} {\bf Parameter}: $\eta \!\in\! \mathbb{R}_+$,\! a\! positive\! learning\! rate.\vspace{0.15cm}\;
\AlFnt{\small} Initialize\! the\! network\! weights\! $w_{j,k}$ $\forall j,k$;\! initialize\! the\! weights\! $u_{j,i}$ $\forall j,i$.\;  
\AlFnt{\small} \For{$i \!=\! 1,\ldots,n$ {\bf and} $j \!=\! 1,\ldots,c$}{
\AlFnt{\small}    Initialize\! the\! $j$th\! relational\! RBF\! prototype:\! $v_{j,i}^r \leftarrow u_{j,i}/\sum_{q=1}^n u_{j,q}$.\;
\AlFnt{\small}    Calculate\! the\! distance\! between\! the\! $j$th\! relational\! RBF\! prototype\! and\! $i$th\! observation:\! $d_{j,i}^{(0)} \leftarrow (R(v_j^r))_i - (v_j^r)^\top\! R (v_j^r)/2$.\;
}
\AlFnt{\small} \For{$t \!=\! 0,1,2,\ldots$}{
\AlFnt{\small}    \For{$p \!=\! 1,\ldots,g$}{
\AlFnt{\small}       Update\! the\! network\! connection\! weights\! and\! biases\! $w_p \!=\! [w_{0,p},w_{1,p},\ldots,w_{c,p}]$\! using\! (2.7):\vspace{0.025cm}\\ $\;\;\;\;\;\;\;\;\;\;\;\;\;\;\;\;\;\;\;\;\;\;\;\;\;\;\;\;\;\;\;\;\displaystyle w_p \leftarrow w_p - \eta \sum_{i=1}^n\sum_{k=1}^g \Bigg(y_{i,k} \!-\! w_{0,k} \!-\! \sum_{j=1}^c w_{j,k} f_1(d_{j,i}^{(t)})\Bigg)\!f_1(d_{j,i}^{(t)})$.\; 
\AlFnt{\small}    }
\AlFnt{\small}    \For{$i \!=\! 1,\ldots,n$ {\bf and} $j \!=\! 1,\ldots,c$}{
\AlFnt{\small}       Simulate\! a\! shift\! of\! the\! vector-based\! RBF\! prototype\! by\! updating\! $d_{j,i}^{(t)}$\! using\! (2.9),\! where\! $\epsilon_{y,j}^h \!=\! w_{i,j} \nabla_{v_j} f_1(d_{j,y})$:\vspace{0.025cm}\\ $\;\;\;\;\displaystyle d_{j,i}^{(t+1)} \leftarrow d_{j,i}^{(t)} + \left(4\sum_{r=1}^n u_{j,r}\right)^{\!\!-2}\!\!\Bigg(\sum_{p,q=1}^n\! u_{j,p} u_{j,q} \eta\sum_{y=1}^n \epsilon_{y,j}^h\Bigg(\! r_{i,y} \!-\! r_{i,q} \!-\! \eta \epsilon_{y,j}^h r_{p,q}\!\Bigg)\!\Bigg)$.\vspace{0.125cm}\; 
\AlFnt{\small}       Update\! the\! Gaussian\! kernel\! variances\! using\! (2.11):\vspace{0.025cm}\\ $\;\;\;\;\;\;\;\;\displaystyle \sigma_j \leftarrow \sigma_j - \eta\sum_{i=1}^n\sum_{k=1}^g \Bigg(y_{i,k} \!-\! w_{0,k} \!-\! \sum_{j=1}^c w_{j,k} \textnormal{exp}(-d_{j,i}^{(t+1)}/2\sigma_j^2)\Bigg)\!\Bigg( w_{j,k} \textnormal{exp}(-d_{j,i}^{(t+1)}/2\sigma_j^2)(d_{j,i}^{(t+1)}/\sigma_j^3)\Bigg)$.\; 
       }
       {\bf exit\! if }$\displaystyle\sum_{i=1}^n\sum_{k=1}^g \Bigg(y_{i,k} \!-\! f_2\Bigg(w_{0,k} \!+\! \sum_{j=1}^c w_{j,k} f_1(d_{j,i}^{(t+1)})\Bigg)\!\Bigg)^{\!2}$\! {\bf has\! not\! changed\! enough\! across\! an\! episode\! range}\;
\AlFnt{\small} }
\end{algorithm}}\vspace{-0.3cm}
\end{figure*}

\subsection*{\small{\sf{\textbf{3.1$\;\;\;$Simulation Setup}}}}

For each dataset, the training of the RBF networks is performed in the same manner:\vspace{0.05cm}
\begin{itemize}
\item[] \-\hspace{0.0cm}{\small{\sf{\textbf{RBF Network Topology.}}}} We considered a conventional, two-layer architecture for both the graph- and vector-based RBF networks.  The hidden layer was composed of RBF basis functions, which we chose to be Gaussian kernels with variances that are learned during training.  The output layer contained linear activation functions.\vspace{0.025cm}

\item[] \-\hspace{0.0cm}{\small{\sf{\textbf{RBF Network Initialization.}}}} Our graph-based RBF networks require the specification of unit-interval weights to initialize the relational RBF prototypes.  We used the relational $k$-means algorithm for this purpose \cite{HathawayRJ-jour1989a}.  The number of clusters was dictated by the number of RBF prototypes.  We seeded this algorithm with a random partition and ran the clustering process until the partition did not change across consecutive iterations.  It has been proved that relational $k$-means algorithm applied to an adjacency-matrix representation of a graph is equivalent to $k$-means applied to a vector realization when the same distance metric specifies the edge weights between pairwise observations.  The final partition thus also defines the prototypes for vector-based RBF networks. 

Both the vector- and graph-based RBF networks have weights, decision-surface biases, and other parameters that are learned via gradient descent.  We randomly initialized these scalar weight values in the range $[\textnormal{-1.75},\textnormal{1.75}]$.  We randomly selected the bandwidth of the Gaussian basis functions from the range $[\textnormal{0.25},\textnormal{3.75}]$.\vspace{0.025cm} 

\item[] \-\hspace{0.0cm}{\small{\sf{\textbf{RBF Network Training and Adaptation.}}}} We randomly binned 70\% of each dataset into a training set and equally split the remainder into testing and validation sets.  The samples that compose both the training and validation sets were also randomly chosen.  

For our simulations, we assume that back-propagation gradient descent would be performed until the error on the validation set increased and then continued to monotonically increase for 30 iterations. If the error of the validation set does not decrease past the initial value that triggered the test, then we return the network weights and prototype parameters that achieved the best performance on the validation set.  

We assigned an initial number of RBF prototypes for each dataset.  Additional prototypes could be added, up to a maximal amount, as follows.  If the validation error monotonically increased for 5 iterations, then a new RBF prototype was introduced for the observation with the highest absolute-magnitude deviation from the expected response.  This new RBF prototype was centered at that observation, which reduces the sum of squared errors.  Random weight values and a bandwidth were generated in the same manner listed above.

We assigned random learning rates in the range $[\textnormal{0.05},\textnormal{2.0}]$ to each weight.  We used an adaptive learning rate scheme to tuning the learning rates rates, which entails scaling them by a constant amount every time a performance threshold is met.  That is, if the mean-squared error for the next current iteration exceeds the mean-squared error from a previous iteration by more than 5\%, then the new weights were discarded.  In addition, the learning rate magnitude for all weights was decreased by 30\%.  Otherwise, the new weights were kept.  If the new error was less than the old error, then the learning rate was increased by 5\%.  This adaptive learning rate scheme has a simple interpretation: it increases the learning rate only to the extent that the network can learn without large error increases.   When the learning rate is too high to guarantee reducing the error, the learning rate is decreased until stable learning resumes.  More advanced schemes could also be considered \cite{PanchapakesanC-jour2002a}.
\end{itemize}

\subsection*{\small{\sf{\textbf{3.2$\;\;\;$Graph-based RBF Network Performance}}}}

\subsection*{\small{\sf{\textbf{3.2.1$\;\;\;$Dataset Descriptions}}}}

We utilize two datasets to highlight the performance of graph-based RBF networks on pure-graph data relative to vector-based RBF networks applied to feature-space realizations of those graphs found via multi-dimensional scaling.  We demonstrate that converting graphs to features, which is the traditional way of dealing with purely graph-based data, can reduce classification performance.  This conversion is also completely unnecessary in view of the duality theory that we presented.\vspace{0.05cm}
\begin{itemize}
\item[] \-\hspace{0.0cm}{$R_\textnormal{HG-194}$\small{\sf{\textbf{: Human Gene Products Dataset.}}}} The first dataset is composed of 194 human gene products retrieved from the ENSEMBL database.  These gene products were from the myotubularin (ENSF 339), receptor precursor (ENSF 73), and collagen alpha chain (ENSF 42) protein families.   These families, respectively, regulate various processes in muscle tissue formation and maintenance, cell division and cell differentiation in cellular membranes, and connective tissue found in cartilage, tendons, and bones.  A total of 21, 87, and 86 sequences were available for the three families, leading to a three-class problem.  For each of these sequences, we performed gene-product summarization via gene-ontology terms \cite{GOC-jour2004}.  A fuzzy-set measure \cite{PopescuM-jour2006a} was employed aggregate the ontological information shared by gene pairs, which furnished the edge weights.\vspace{0.025cm}

\item[] \-\hspace{0.0cm}{$R_\textnormal{PC-102}$\small{\sf{\textbf{: Prostate Cancer Dataset.}}}} The second dataset was derived from expression array results of 50 non-tumor prostate samples and 52 prostate tumor samples obtained from patients undergoing surgery \cite{SinghD-jour2002a}.  Oligonucleotide microarrays containing probes for approximately 12,600 genes and expressed sequence tags were used to generate the expressions.  Since prostate tumors are, histologically, among the most heterogeneous of cancers, the goal is to simply separate between tumor and non-tumor samples.  We do this by employing a Bayesian similarity metric to the gene expressions \cite{HunterL-jour2001a}.  Given an estimate of the joint density function over all the genes, this metric attempts to integrate over all possible sequences that could have been common ancestors of the sequence pairs to determine their relatedness.  Such an approach takes into account not only the correlation among the genes, but also the complex distributional properties of the gene density functions.\vspace{0.05cm}
\end{itemize}

\begin{figure*}
\hspace{-0.025cm}\begin{tabular}{c}
   \includegraphics[height=1.0in]{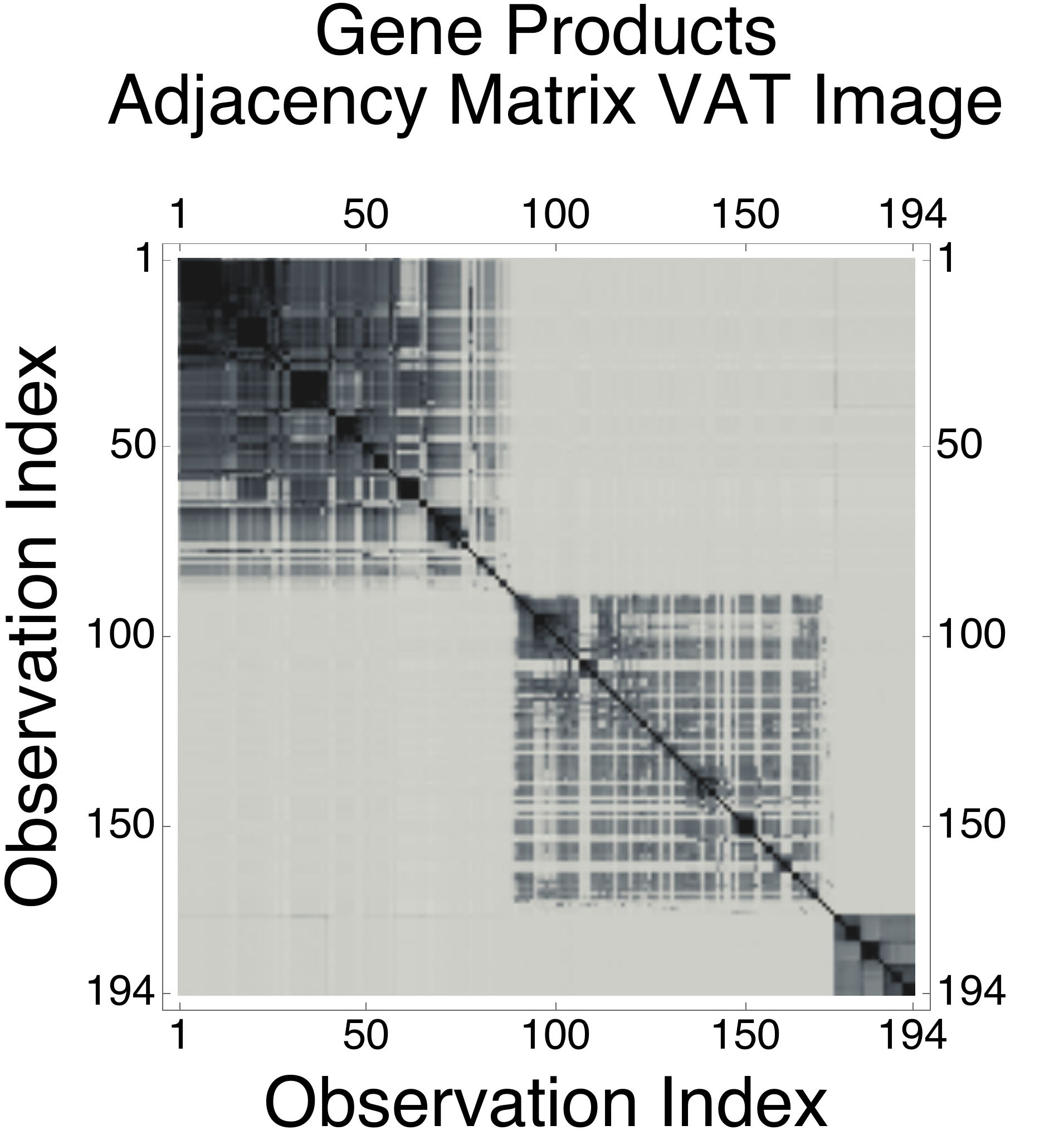} \includegraphics[height=1.0in]{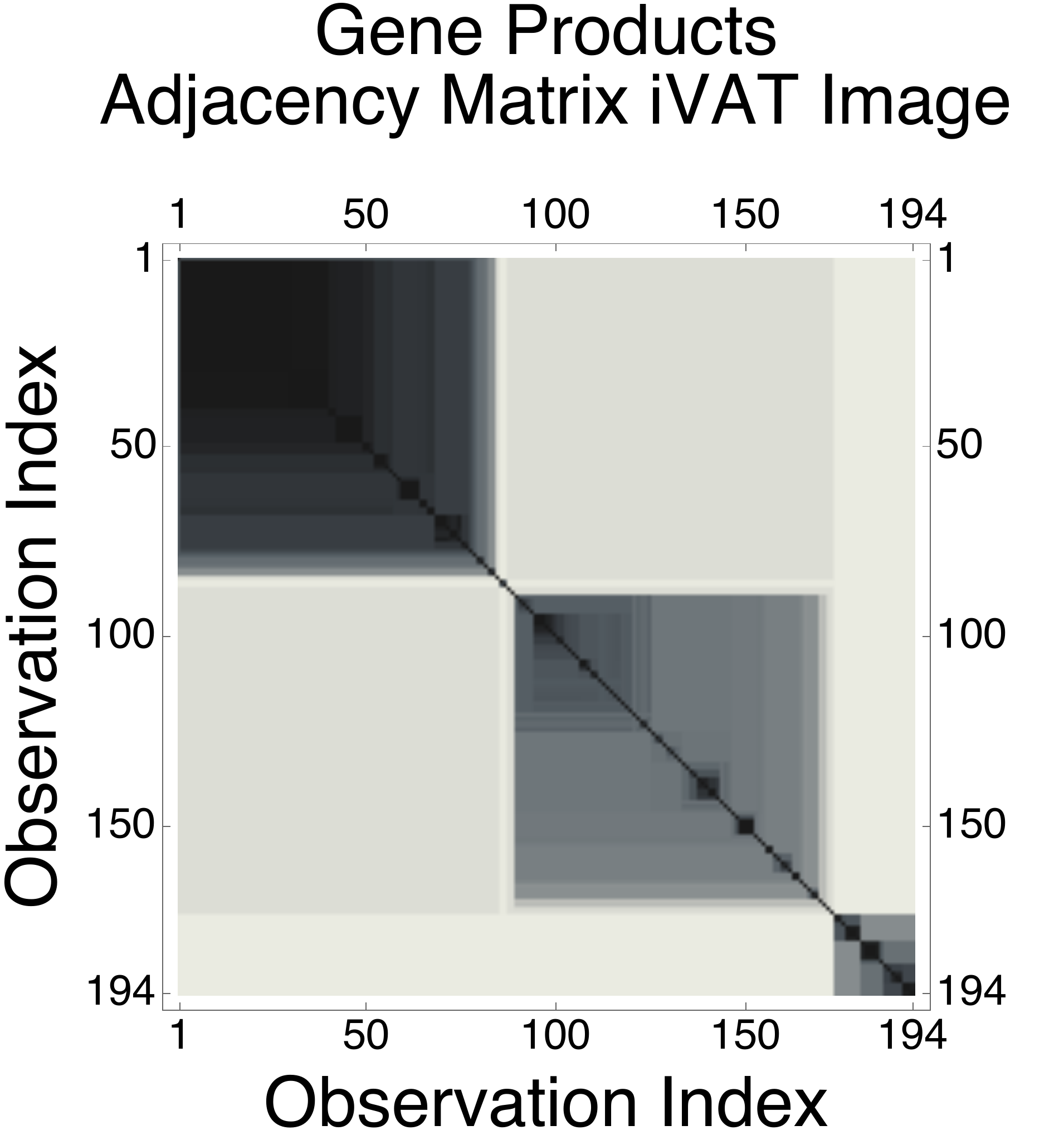} $\;\;$ \includegraphics[height=1.0in]{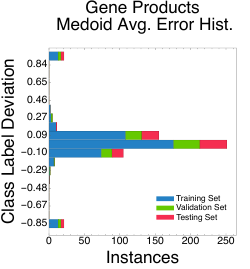} \hspace{-0.15cm} \includegraphics[height=1.0in]{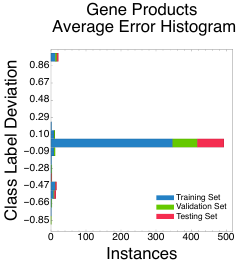} \hspace{-0.0675cm} \includegraphics[height=1.0in]{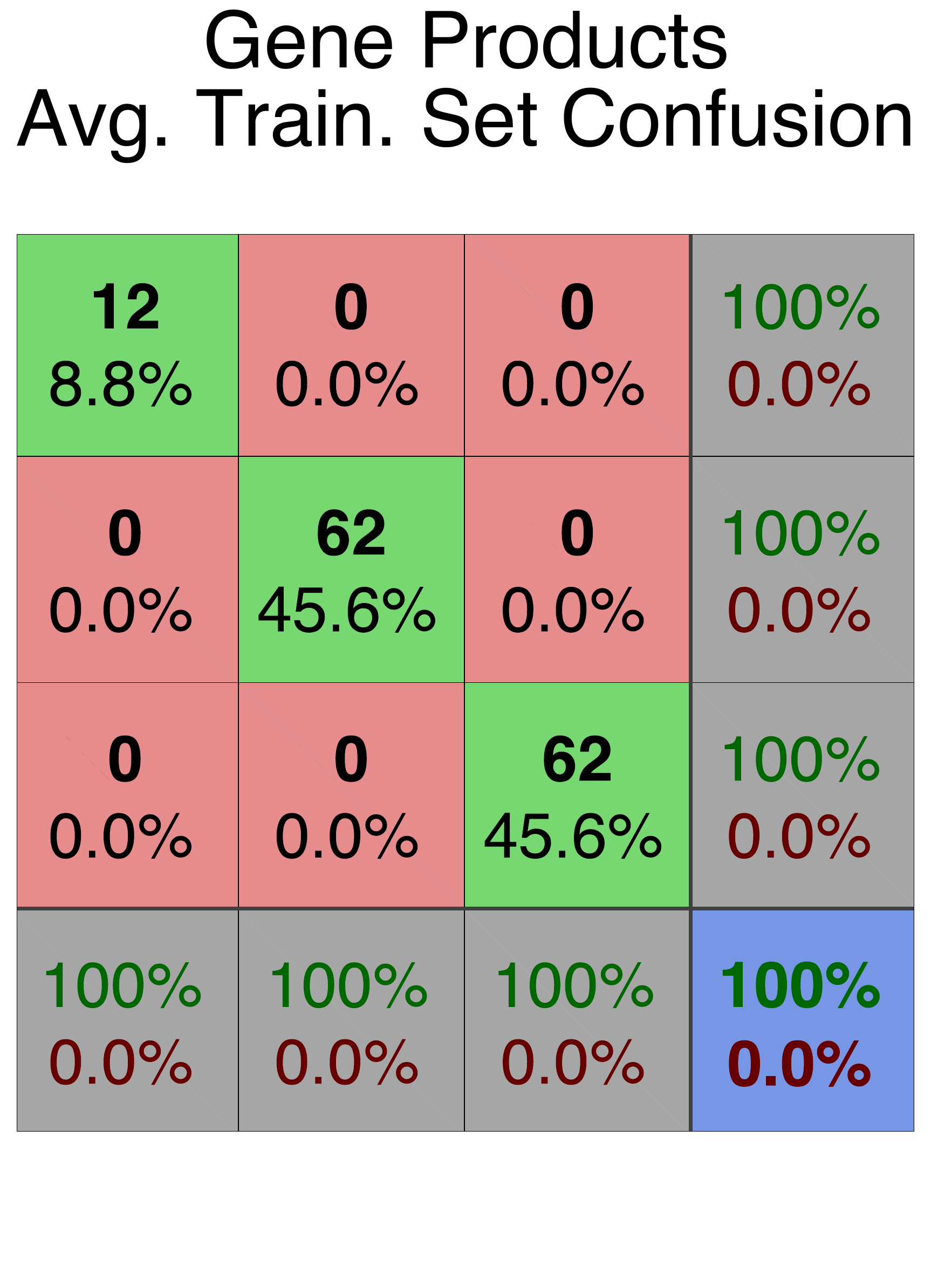} \hspace{0.075cm} \includegraphics[height=1.0in]{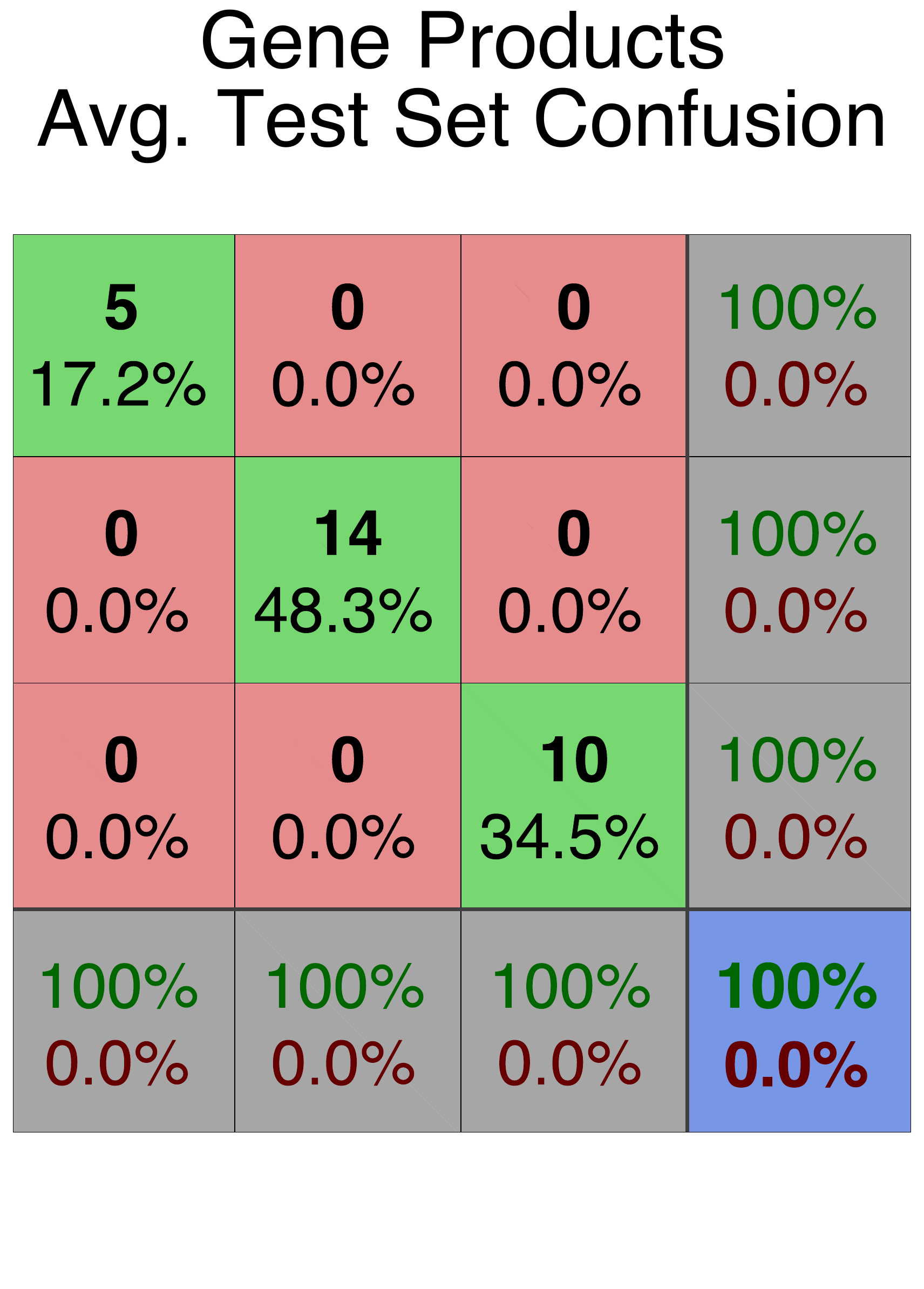} \hspace{0.075cm} \includegraphics[height=1.0in]{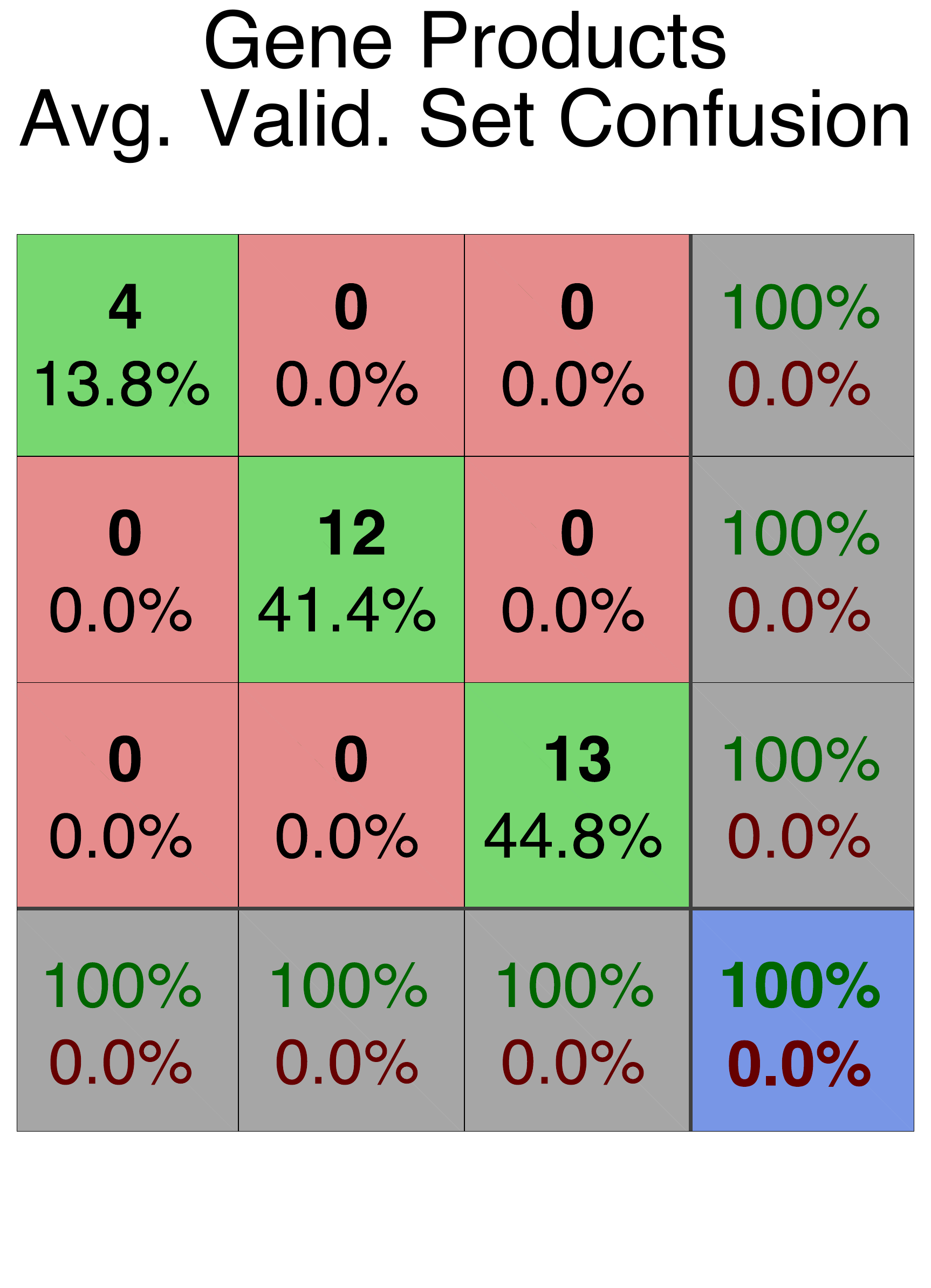}\vspace{-0.15cm}\\
   {\footnotesize (a)}\vspace{0.15cm}\\
   \includegraphics[height=1.0in]{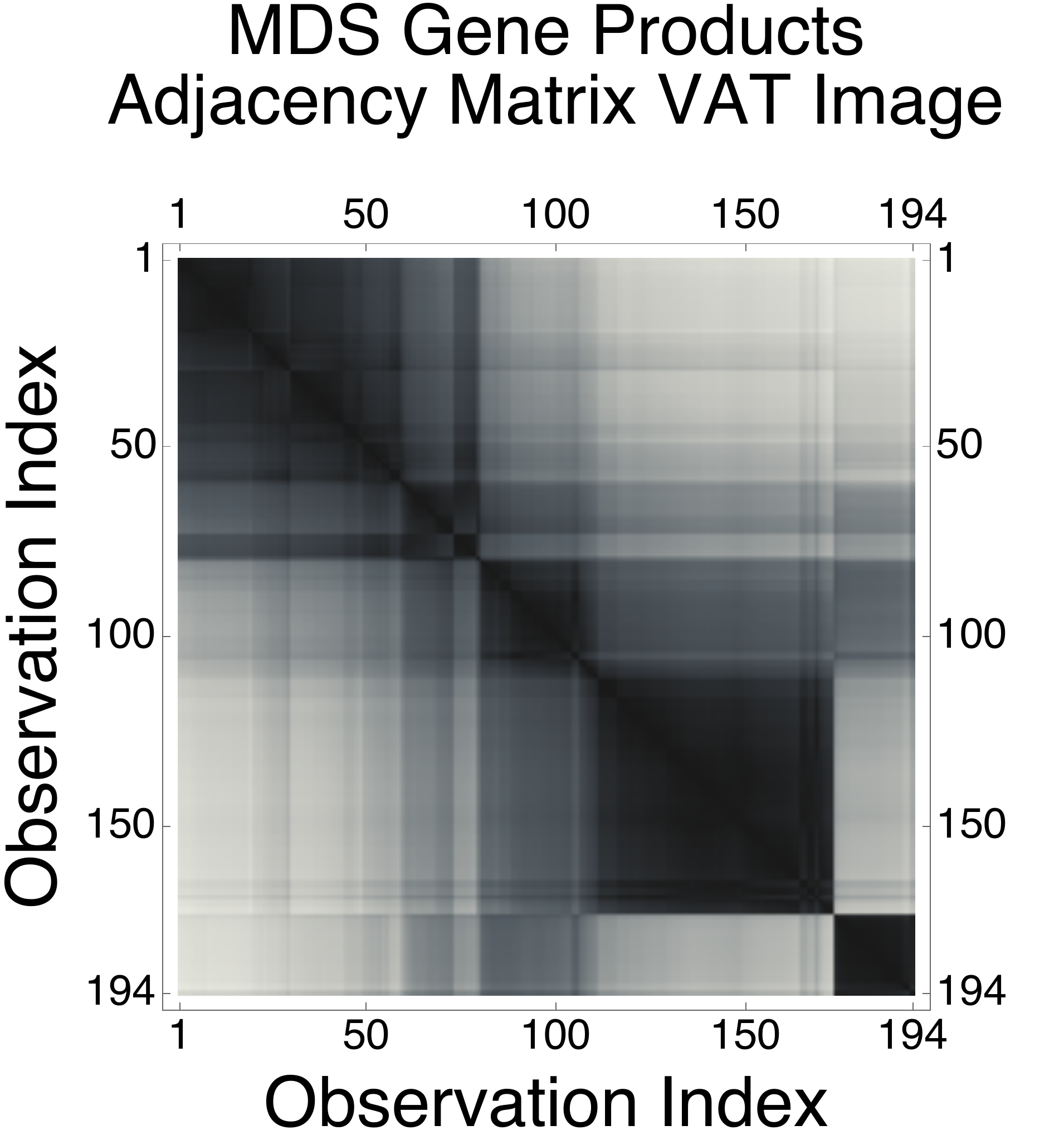} \includegraphics[height=1.0in]{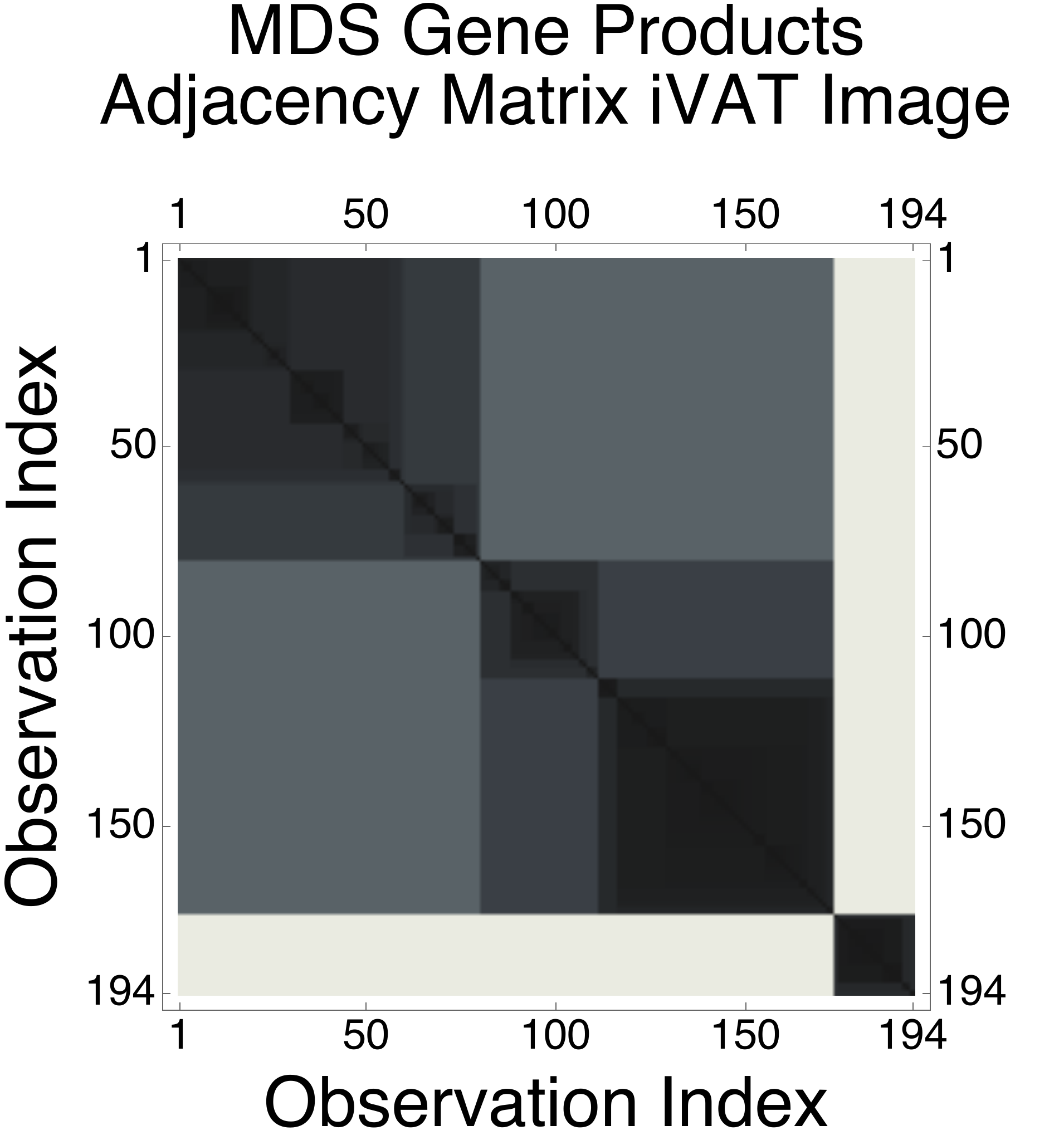} $\;\;$ \includegraphics[height=1.0in]{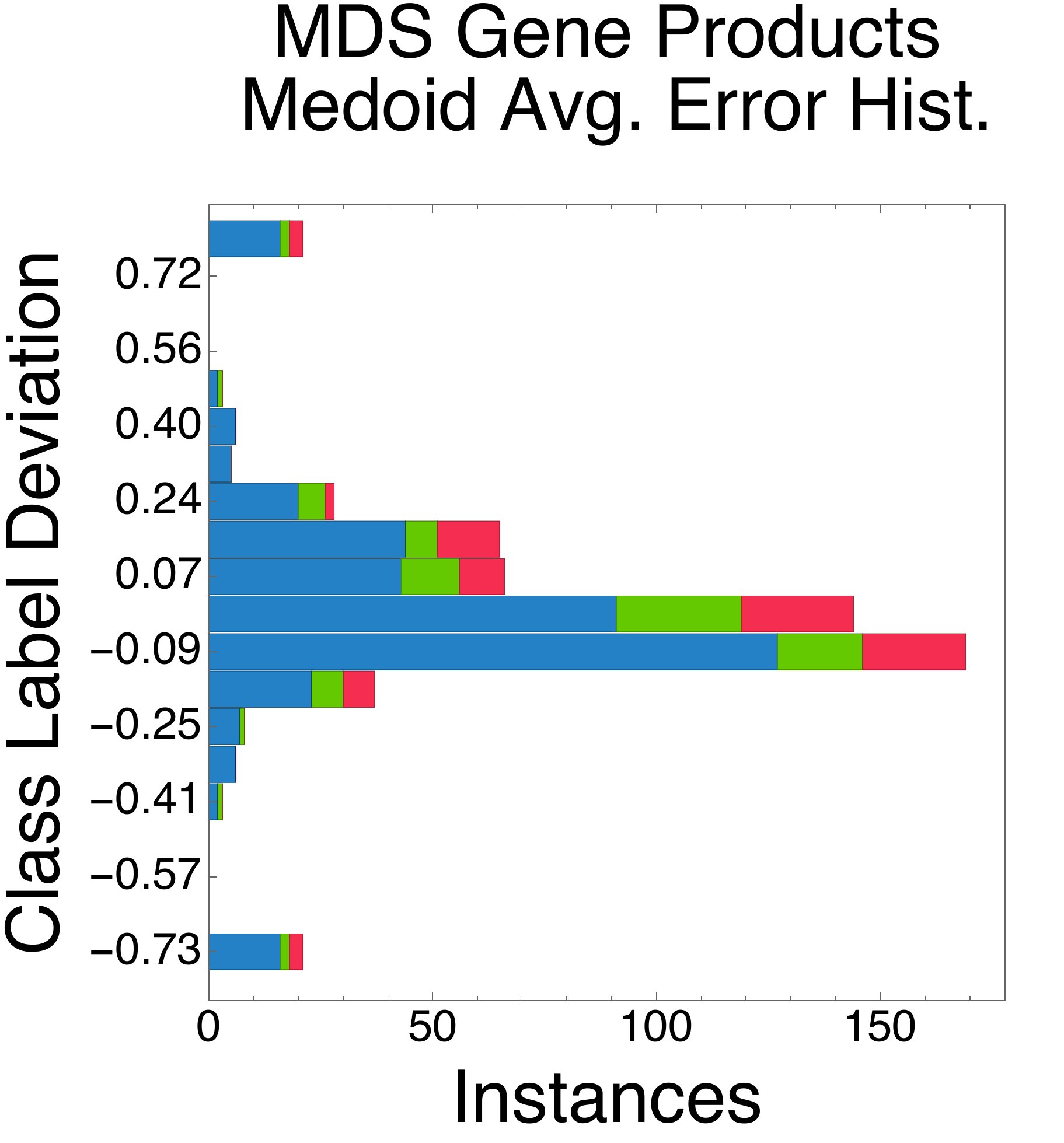} \hspace{-0.15cm} \includegraphics[height=1.0in]{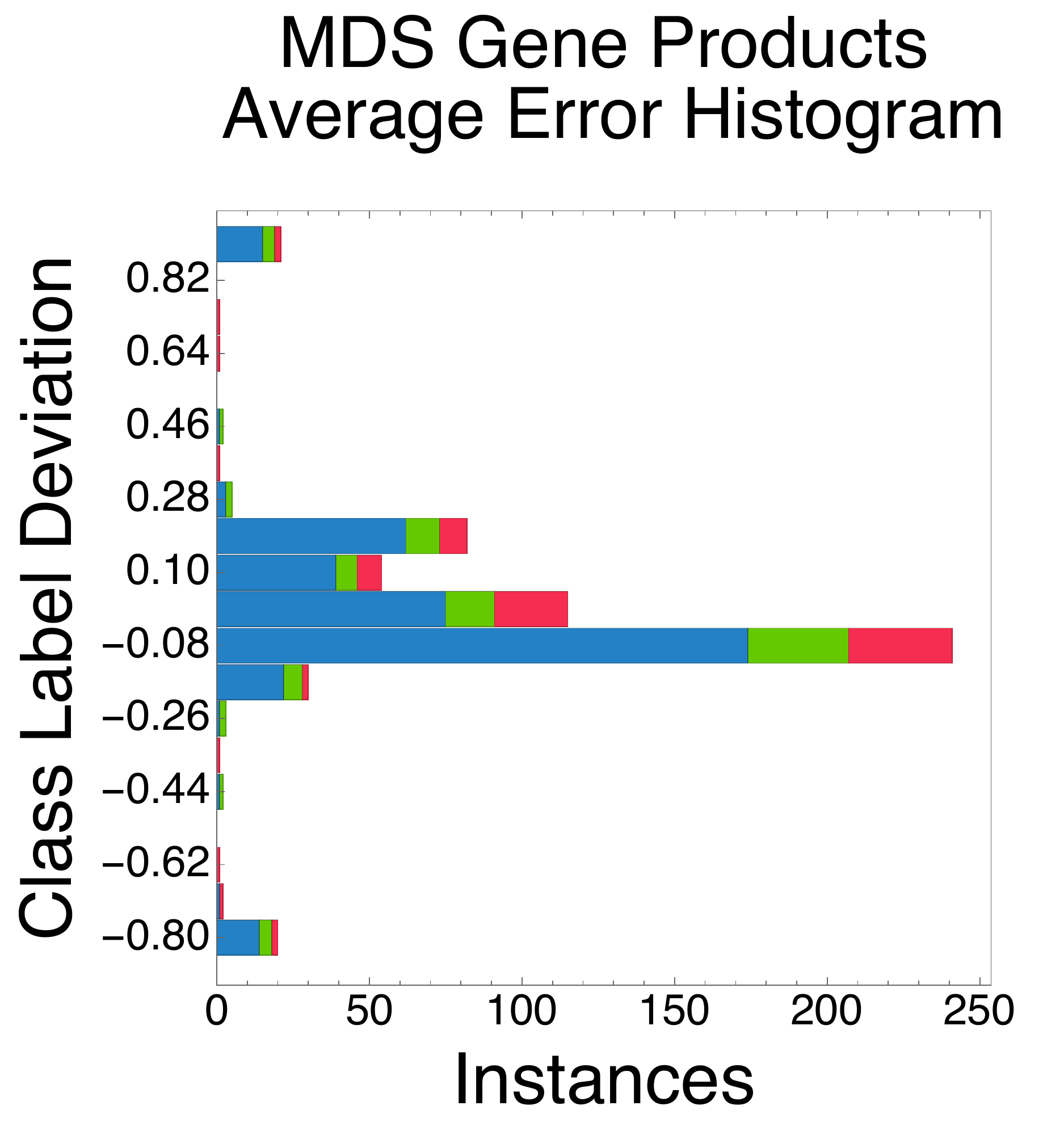} \hspace{-0.0675cm} \includegraphics[height=1.0in]{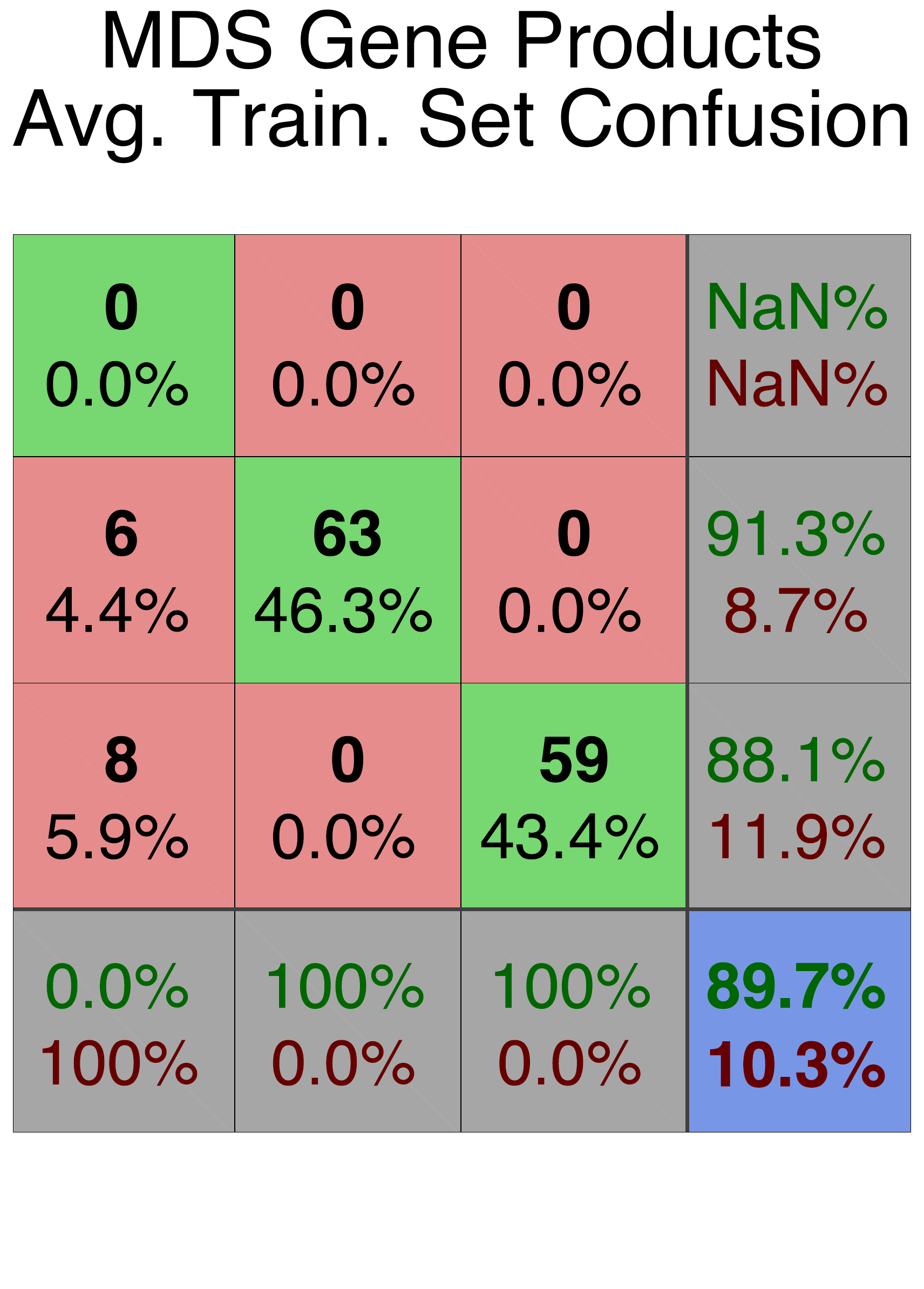} \hspace{0.075cm} \includegraphics[height=1.0in]{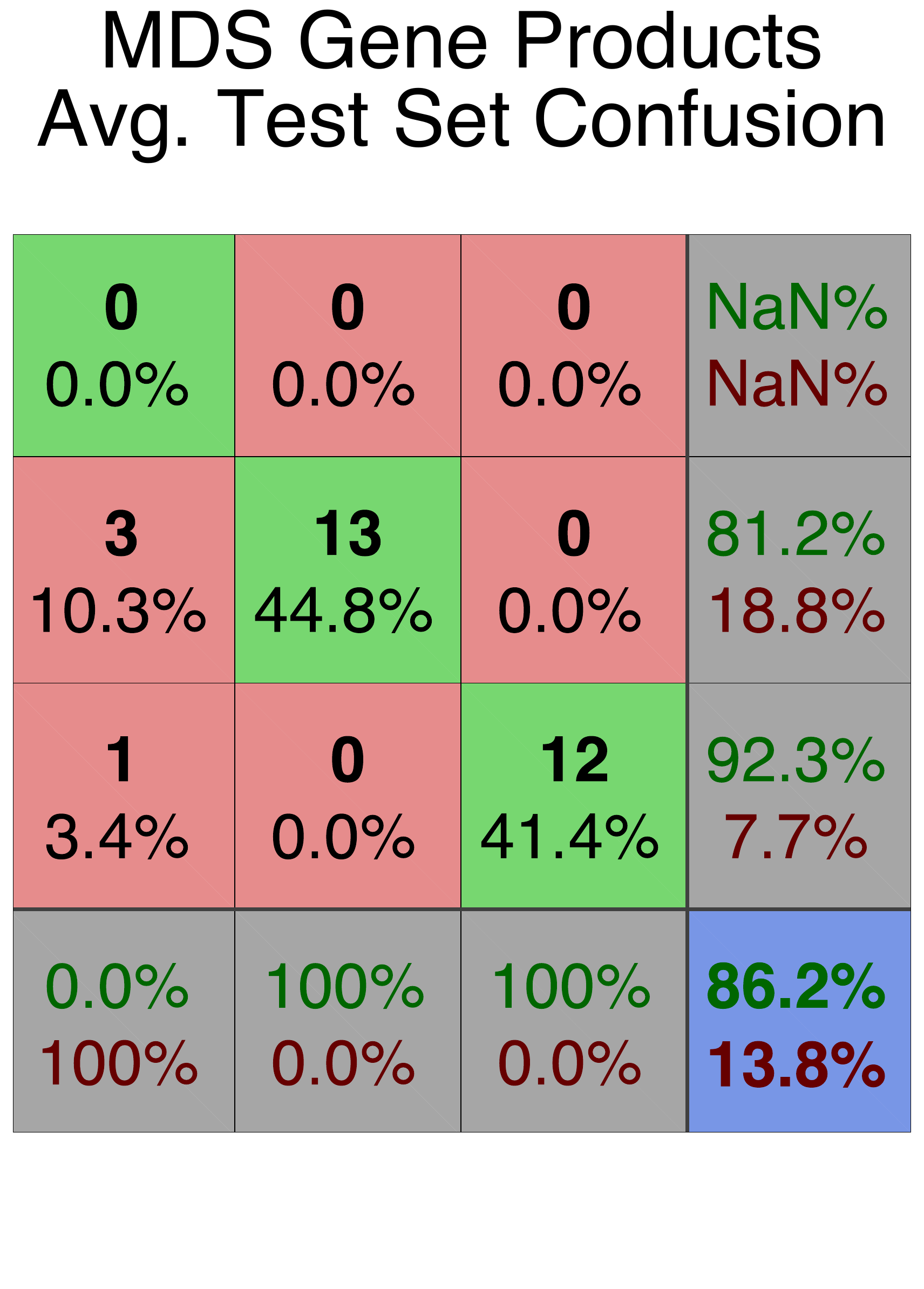} \hspace{0.075cm} \includegraphics[height=1.0in]{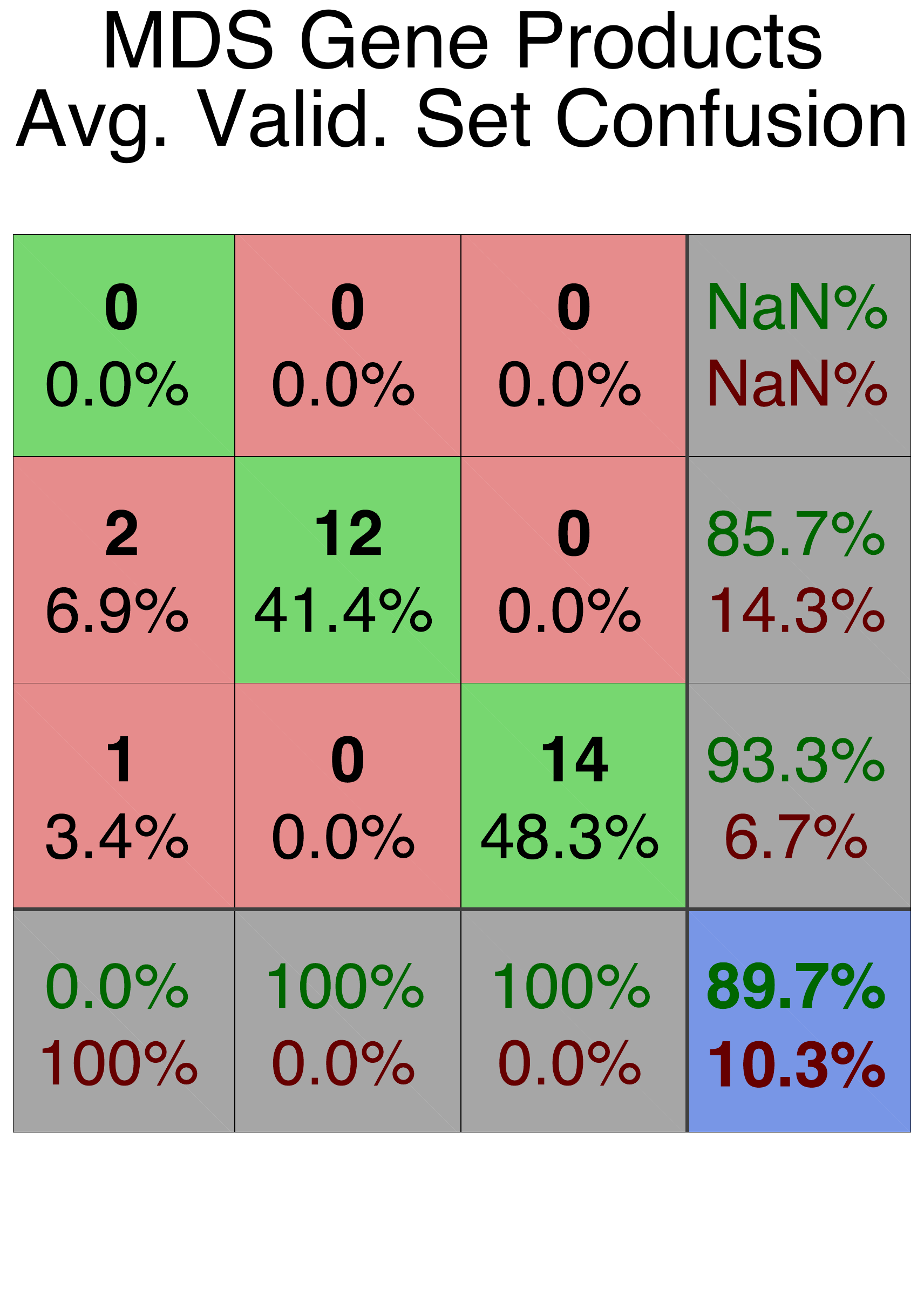}\vspace{-0.15cm}\\
   {\footnotesize (b)}\vspace{0.15cm}\\
   \includegraphics[height=1.0in]{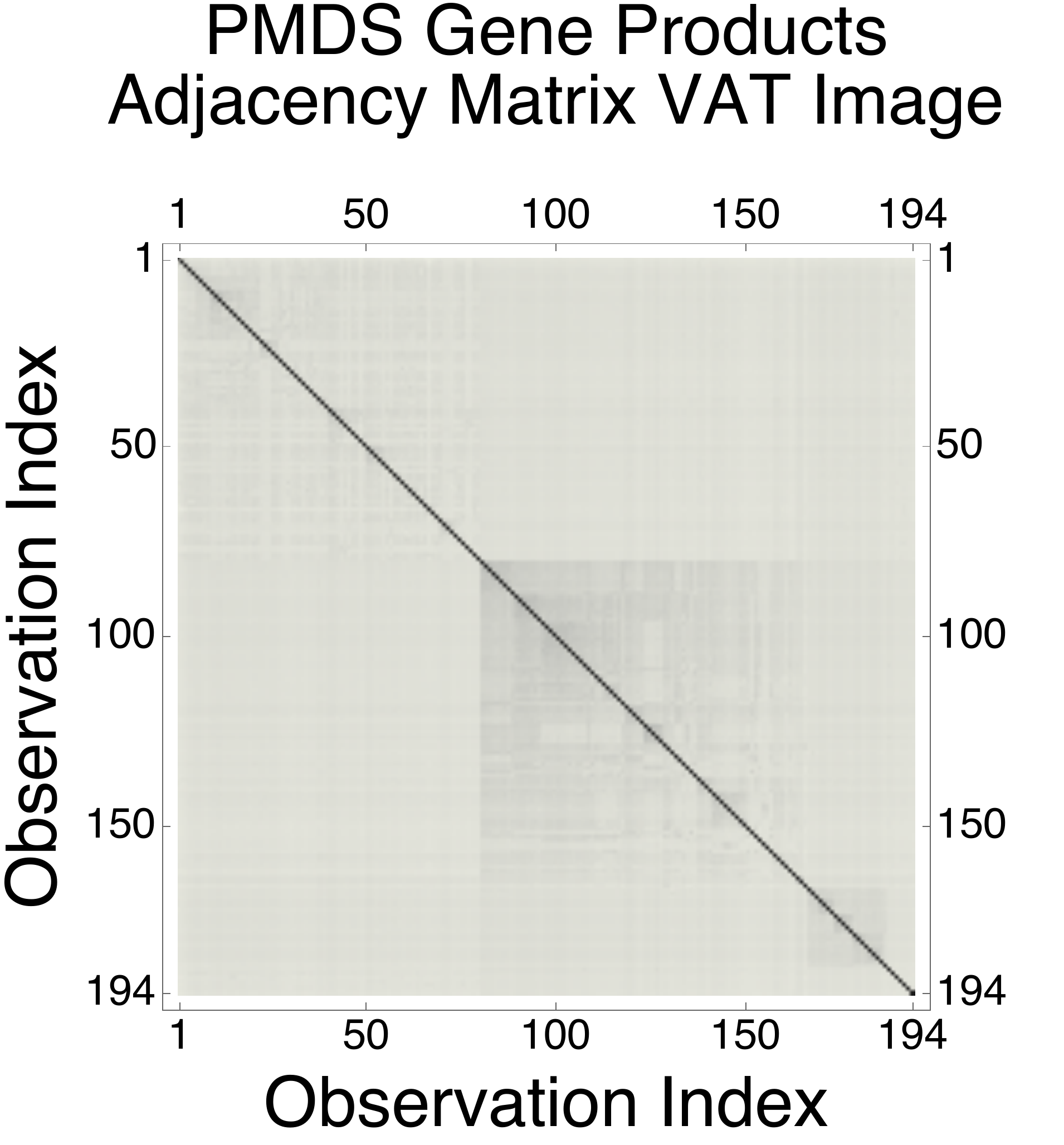} \includegraphics[height=1.0in]{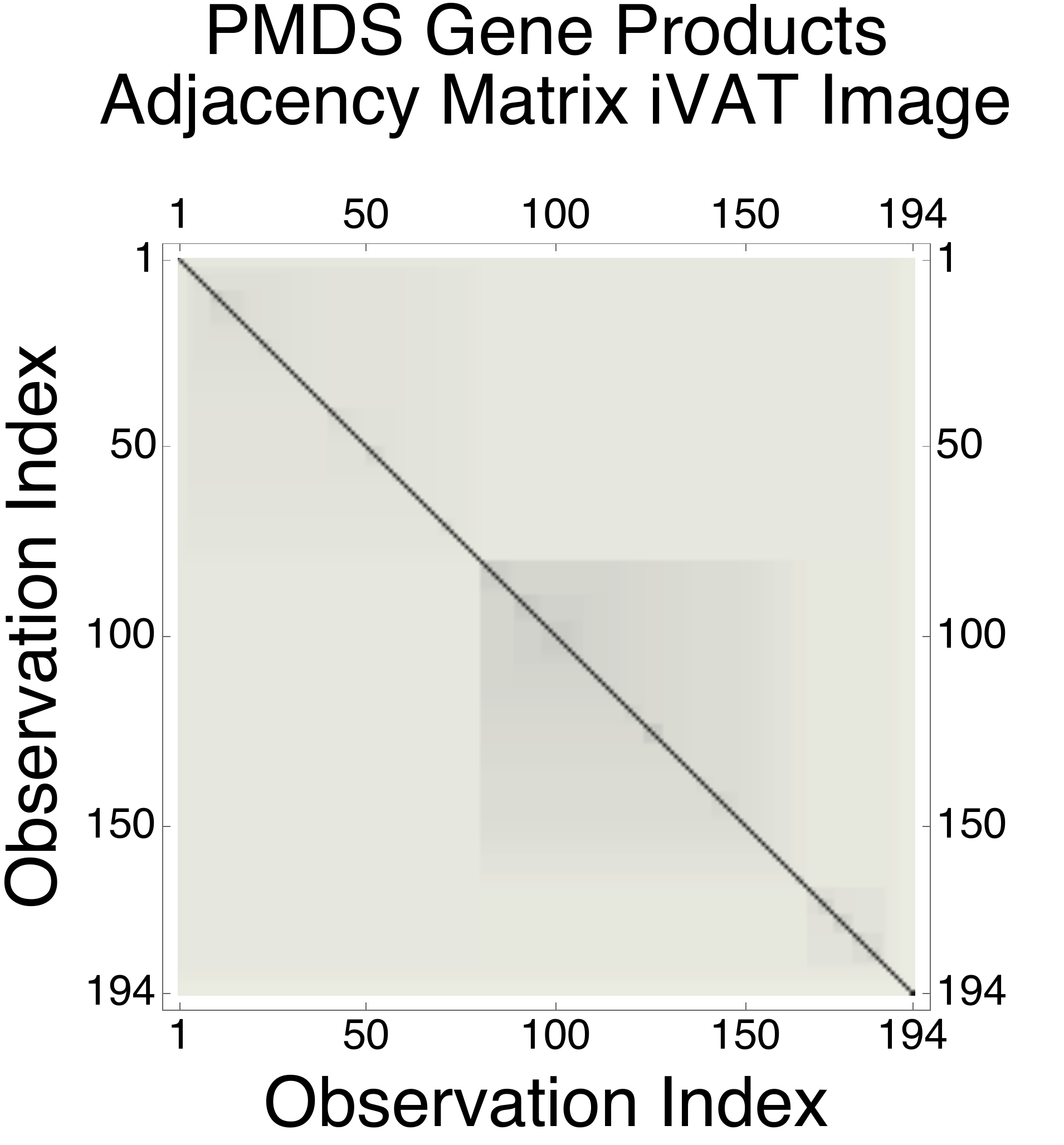} $\;\;$ \includegraphics[height=1.0in]{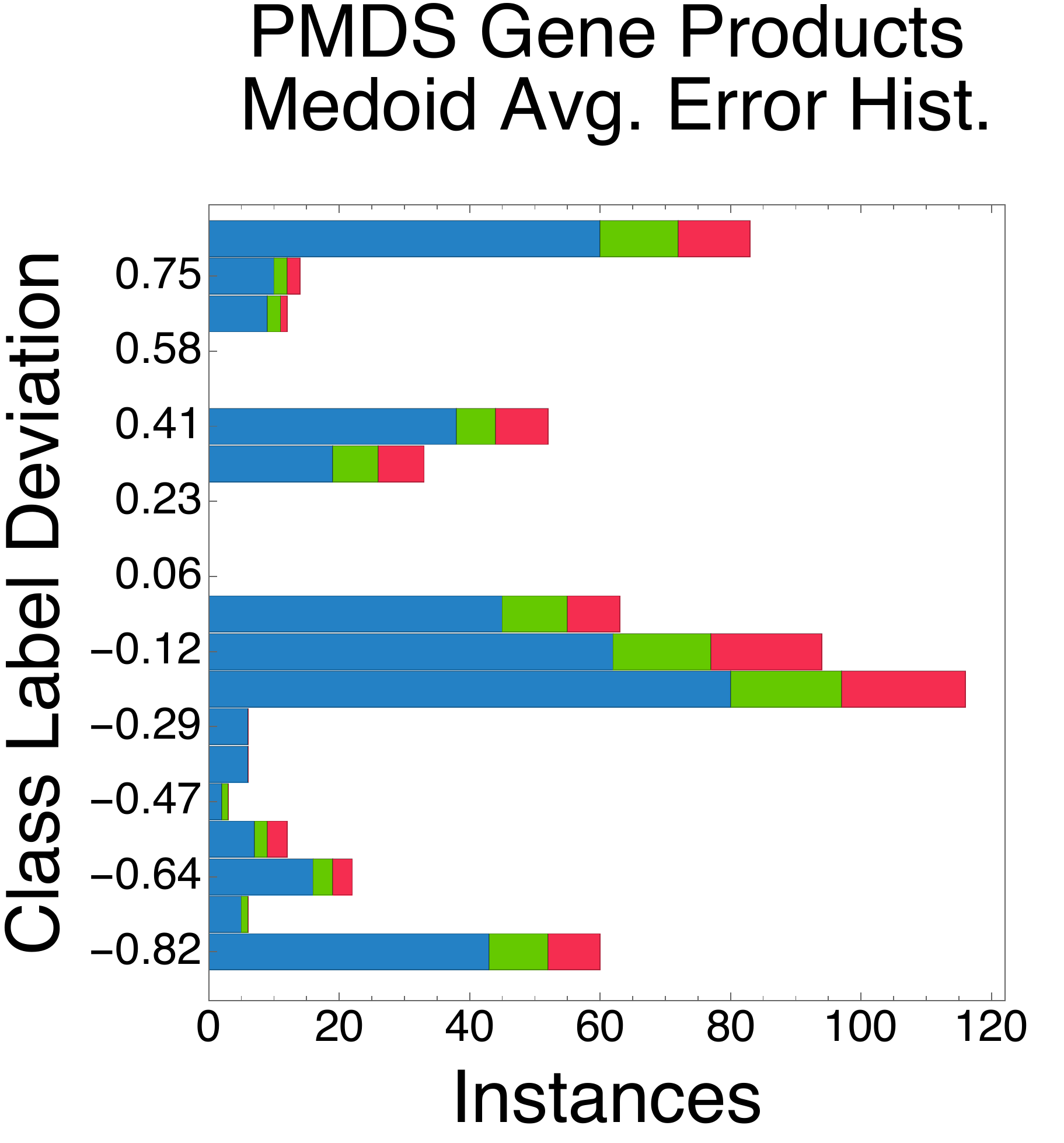} \hspace{-0.15cm} \includegraphics[height=1.0in]{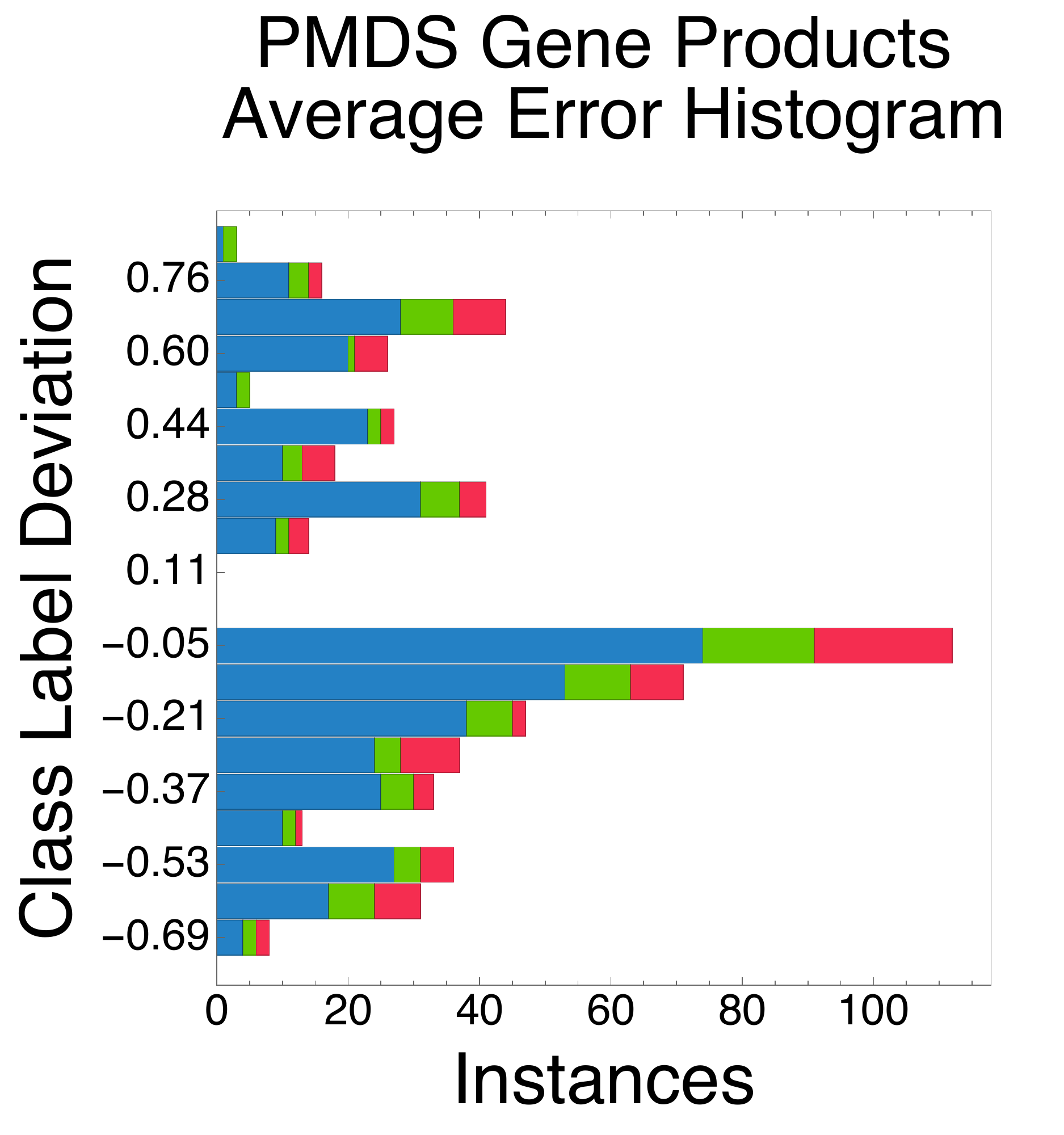} \hspace{-0.0675cm} \includegraphics[height=1.0in]{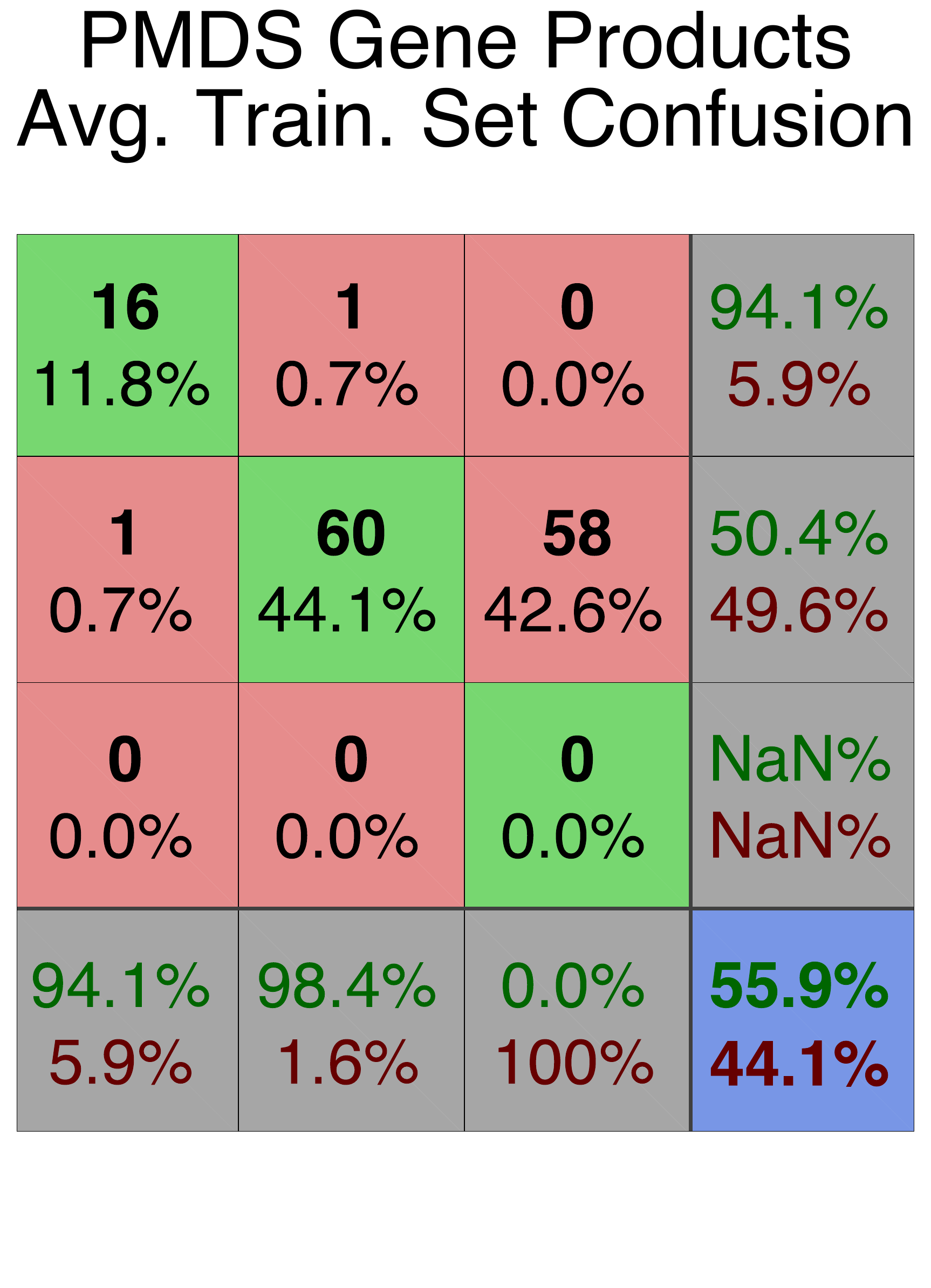} \hspace{0.075cm} \includegraphics[height=1.0in]{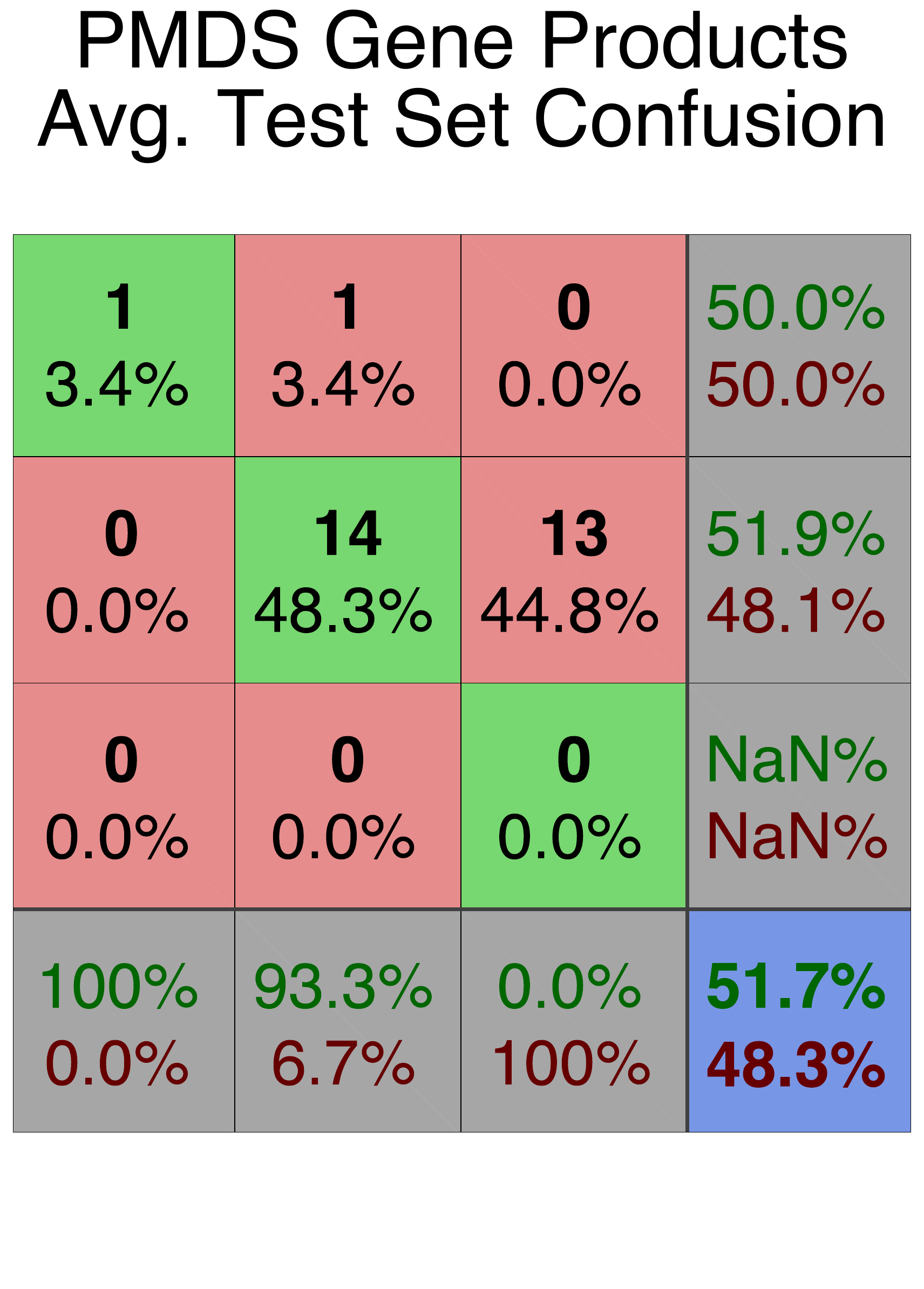} \hspace{0.075cm} \includegraphics[height=1.0in]{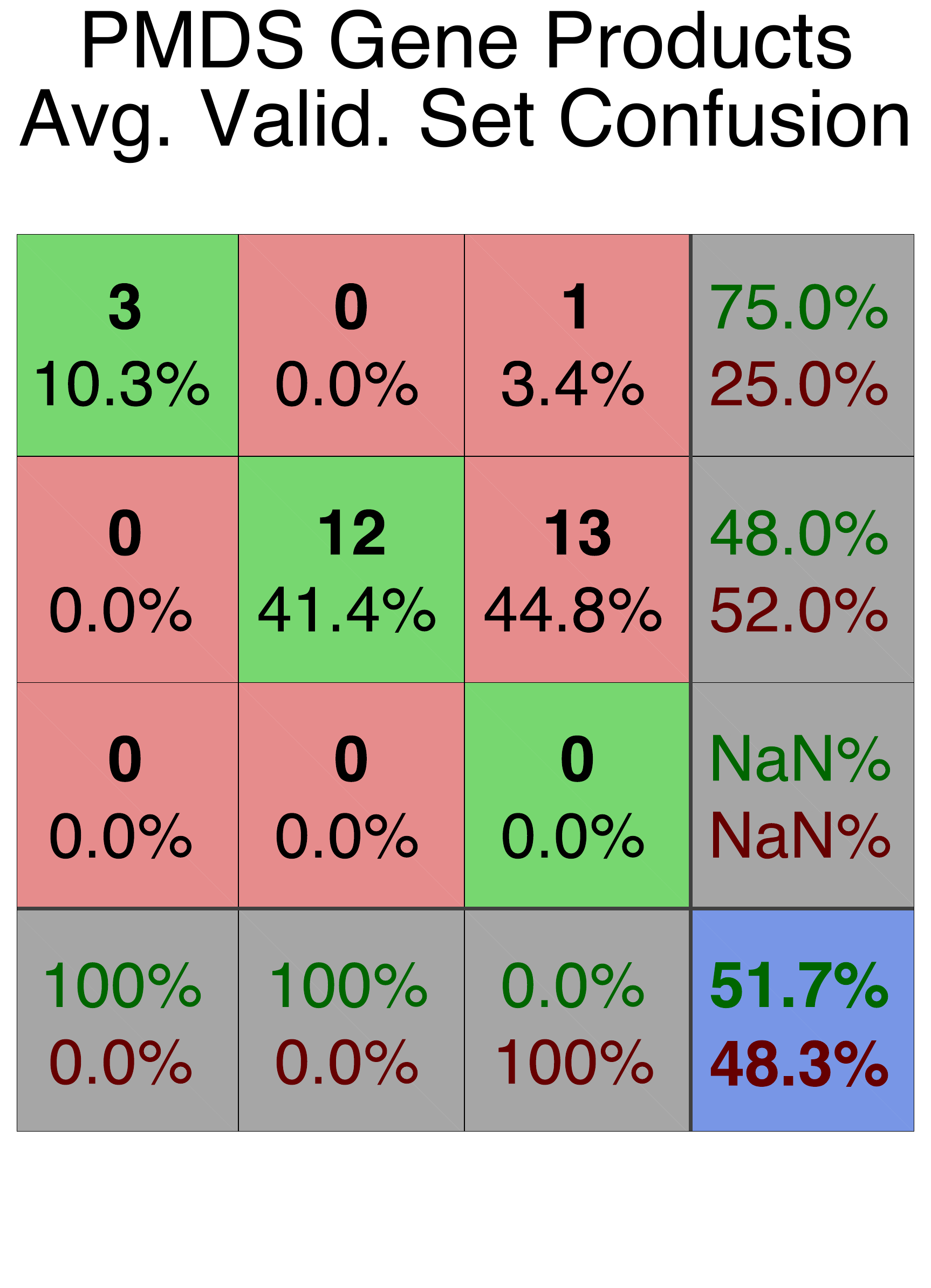}\vspace{-0.15cm}\\
   {\footnotesize (c)}\\
   \end{tabular}
\caption*{Figure 3.1: Classification results for a purely graph-based dataset compared to transformed versions of it.  We consider a dataset, $R_\textnormal{HG-194}$, based on applying a gene-ontology similarity measure to a series of human gene products.  Row (a) corresponds to results for the original dataset.  Rows (b) and (c) correspond to a multi-dimensional-scaling (MDS) and proximal-multi-dimensional-scaling (PMDS) version of the original dataset, respectively.  For each row, the first two columns show the VAT and iVAT images of the graph adjacency matrices.  The third through seventh columns highlight the classification results.  The fourth column shows the deviation in the predicted response from the desired response.  The third column highlights the predicted response deviation when using medoids instead of relational RBF prototypes; the resulting performance is worse when medoids are employed.  The remaining columns contain confusion matrices for the training, test, and validation sets.  The results show that transforming graph-based data into a series of feature vectors, so that conventional, vector-based RBF networks can be used, leads to worse performance.\vspace{-0.4cm}}
\end{figure*}

\noindent Both of these datasets are purely graph-based: they do not have underlying vector realizations from which the edge weights are directly obtained.  The edge weights do, however, satisfy the duality properties, so they do have a vector realization for some a priori unknown metric.

The VAT \cite{BezdekJC-conf2002a,BezdekJC-jour2005a,BezdekJC-jour2006a} and iVAT \cite{HavensTC-jour2012a} algorithms were applied to the graph adjacency matrices to reveal the relationship group structure.  The resulting plots are presented in the first two columns of figures 3.1(a) and 3.2(a).  Dark blocks along the main diagonal of the VAT and iVAT images indicate the presence of compact groups of objects; light off-diagonal blocks indicate the separateness of the objects, with lighter values indicating greater separation.


We created feature-space representations of the two graphs via multi-dimensional scaling \cite{BorgI-book2005a}.  Multi-dimensional scaling translates the pairwise distances in the adjacency matrix to a configuration of feature vectors such that the distances are preserved as well as possible.  We applied classical (CMDS) and proximal multi-dimensional scaling (PMDS).  We briefly consider other embedding techniques in our experiments.  In either case, we retained the maximal dimensionality of the features that were possible and converted them back into graphs using the Euclidean distance metric.  The first two columns in figures 3.1(b)--3.1(c) and 3.2(b)--3.2(c) provide VAT and iVAT plots, respectively, of the transformed datasets when using CMDS and PMDS.  These transformed datasets provide a means of comparing the graph- and vector-based network performance when feature vectors are not naturally available, as is the case for these two datasets.


\subsection*{\small{\sf{\textbf{3.2.2$\;\;\;$Simulation Findings}}}}

Classification results for $R_\textnormal{HG-194}$ and $R_\textnormal{PC-102}$, along with their transformed versions, are provided in the third through seventh columns of figures 3.1(a)--3.1(c) and 3.2(a)--3.2(c).  We considered 100 Monte Carlo simulations with different parameter initializations along with different training, testing, and validation datasets.  

For $R_\textnormal{HG-194}$ and its various converted variants, each simulation began with a total of 15 RBF prototypes and could increase up to a maximum of 30 centers.  For $R_\textnormal{PC-102}$, 12 RBF prototypes could initially be used, and the total could rise to 32 prototypes. 


\vspace{0.05cm}{\small{\sf{\textbf{Results.}}}} As illustrated by the VAT and iVAT plots in figure 3.1(a), the relationships for the $R_\textnormal{HG-194}$ dataset have a mostly compact, well-separated structure.  Each of the dark blocks along the diagonal solely contains sequence-sequence dissimilarities from one of the three protein families.  The top, left dark block corresponds to sequences from the receptor-precursor family, followed by the collagen-alpha-chain family in the middle-right and the myotubularin family in the bottom right of the plot.  Since the within-class dissimilarities a low and the between-class dissimilarities are high for the chosen metric, we expected that the graph-based RBF networks would yield good classification performance.  The error histogram and confusion matrices in figure 3.1(a) show this occurred: the correct classification label was returned an overwhelming majority of the time.  Likewise, there were no classification errors, on average, for any of the randomly-chosen training, testing, and validation sets.  Only in five simulations did we witness non-zero classification errors.


Figures 3.1(b) and 3.1(c) present results when applying graph-based RBF networks to CMDS and PMDS versions of the original $R_\textnormal{HG-194}$ dataset.  We denote these datasets by $R_\textnormal{HG-194}^\textnormal{CMDS}$ and $R_\textnormal{HG-194}^\textnormal{PMDS}$, respectively.  Both algorithms resulted in a significant alteration of the relationship structure.  In the case of $R_\textnormal{HG-194}^\textnormal{CMDS}$, there was significant overlap between two of the protein sequence classes, receptor-precursor family and the collagen-alpha-chain family, that did not exist before.  Many of the sequences within a particular class were, additionally, adjusted to have high similarity to each other, which did not reflect the original class statistics.  PMDS had the effect of erasing most of the sequence relationship structure.  Classification performance consequently suffered in either case.  It dropped by nearly 10\% to 14\% from $R_\textnormal{HG-194}$ when $R_\textnormal{HG-194}^\textnormal{CMDS}$ was used.  The histogram error plot for $R_\textnormal{HG-194}^\textnormal{CMDS}$ shows a higher prevalence for deviation near the true response; a greater number of predicted labels are also entirely incorrect.  Most of the errors were made distinguishing between the receptor-precursor and collagen-alpha-chain families.  For $R_\textnormal{HG-194}^\textnormal{PMDS}$, the error histogram was mostly uniform.  Due to the almost non-existent class structure, much of the data was assigned to a single class, reducing classifier effectiveness by 44\% to 48\%.  

\begin{figure*}
\hspace{-0.025cm}\begin{tabular}{c}
   \includegraphics[height=1.0in]{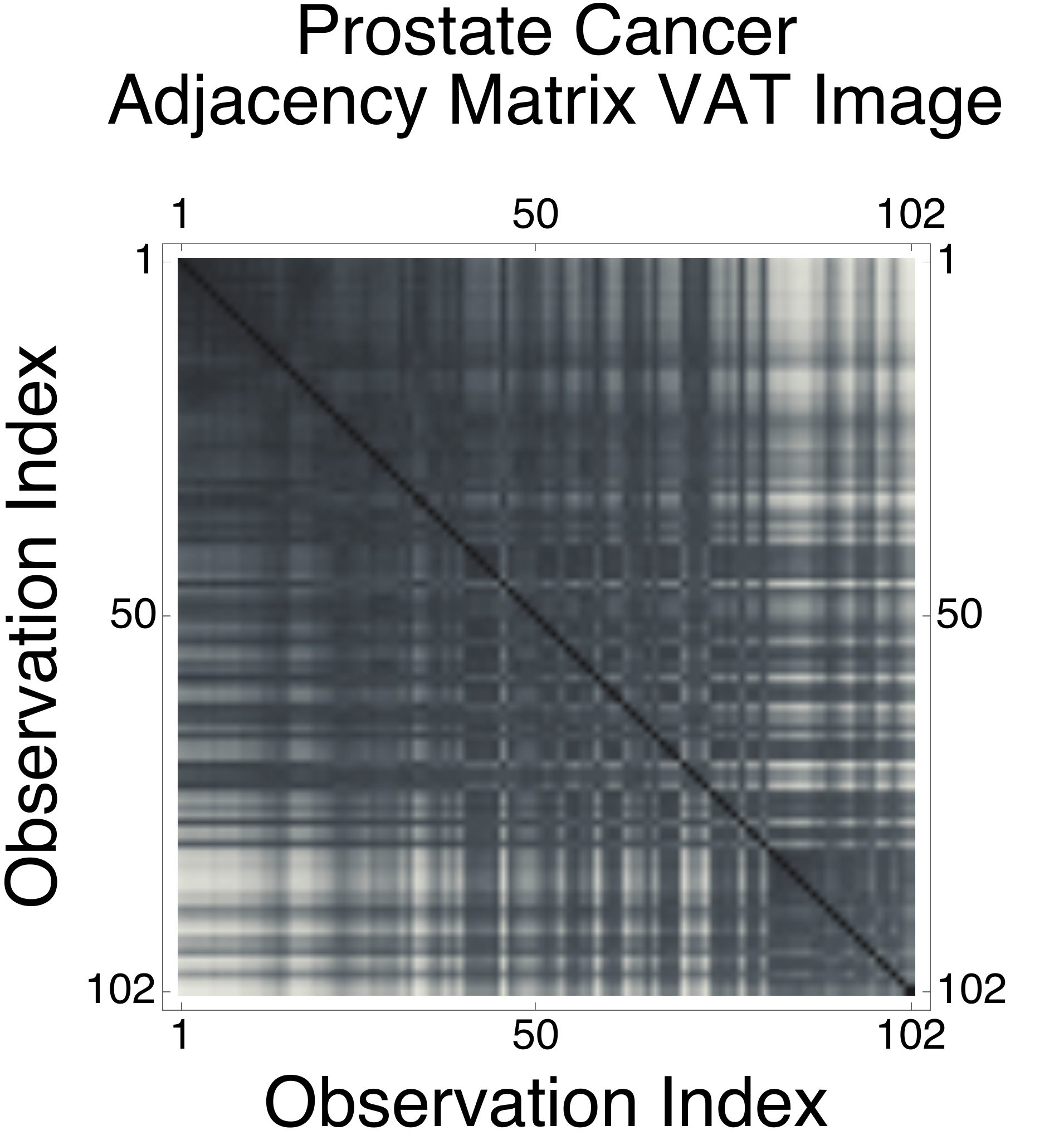} \includegraphics[height=1.0in]{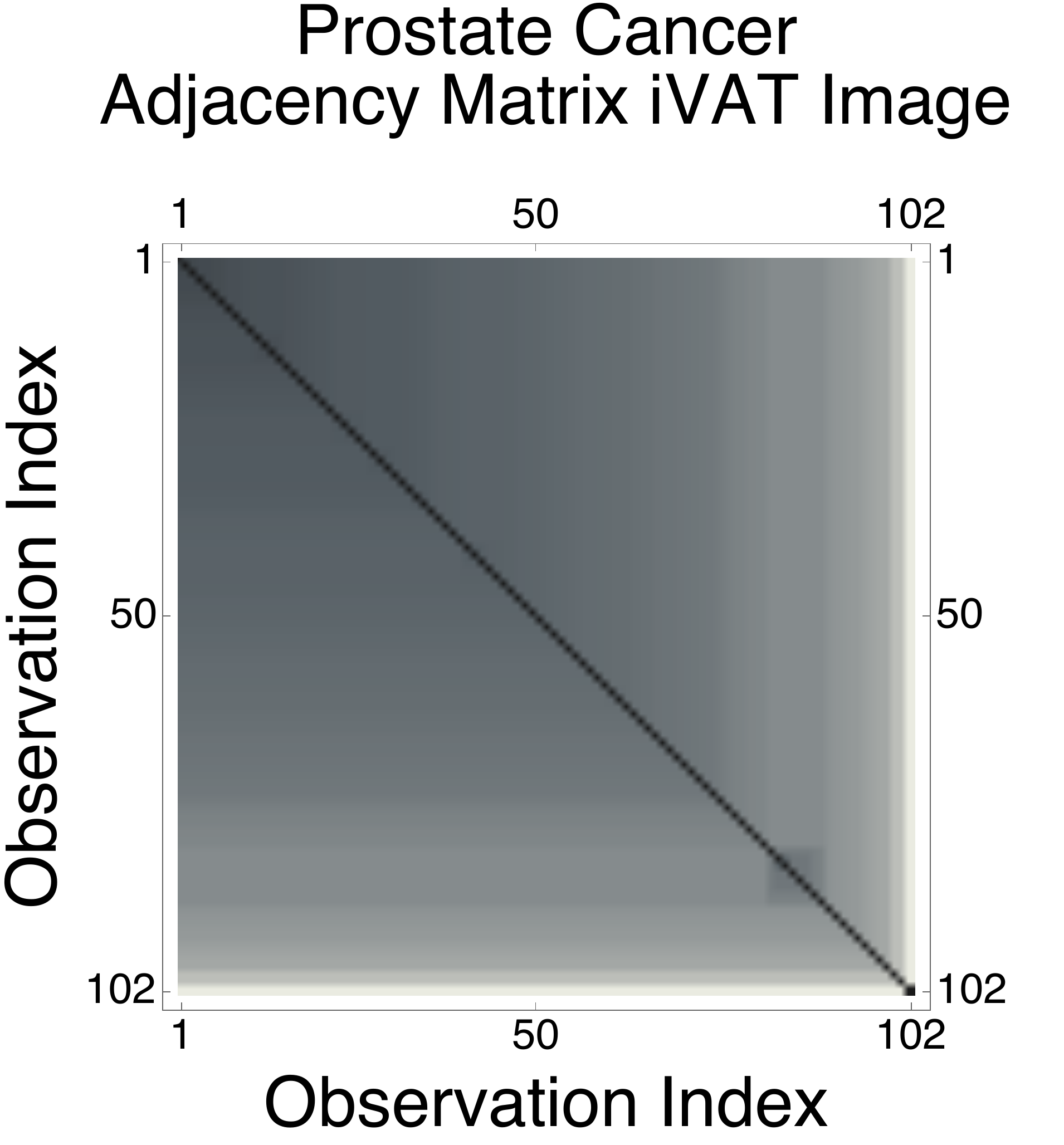} $\;\;$ \includegraphics[height=1.0in]{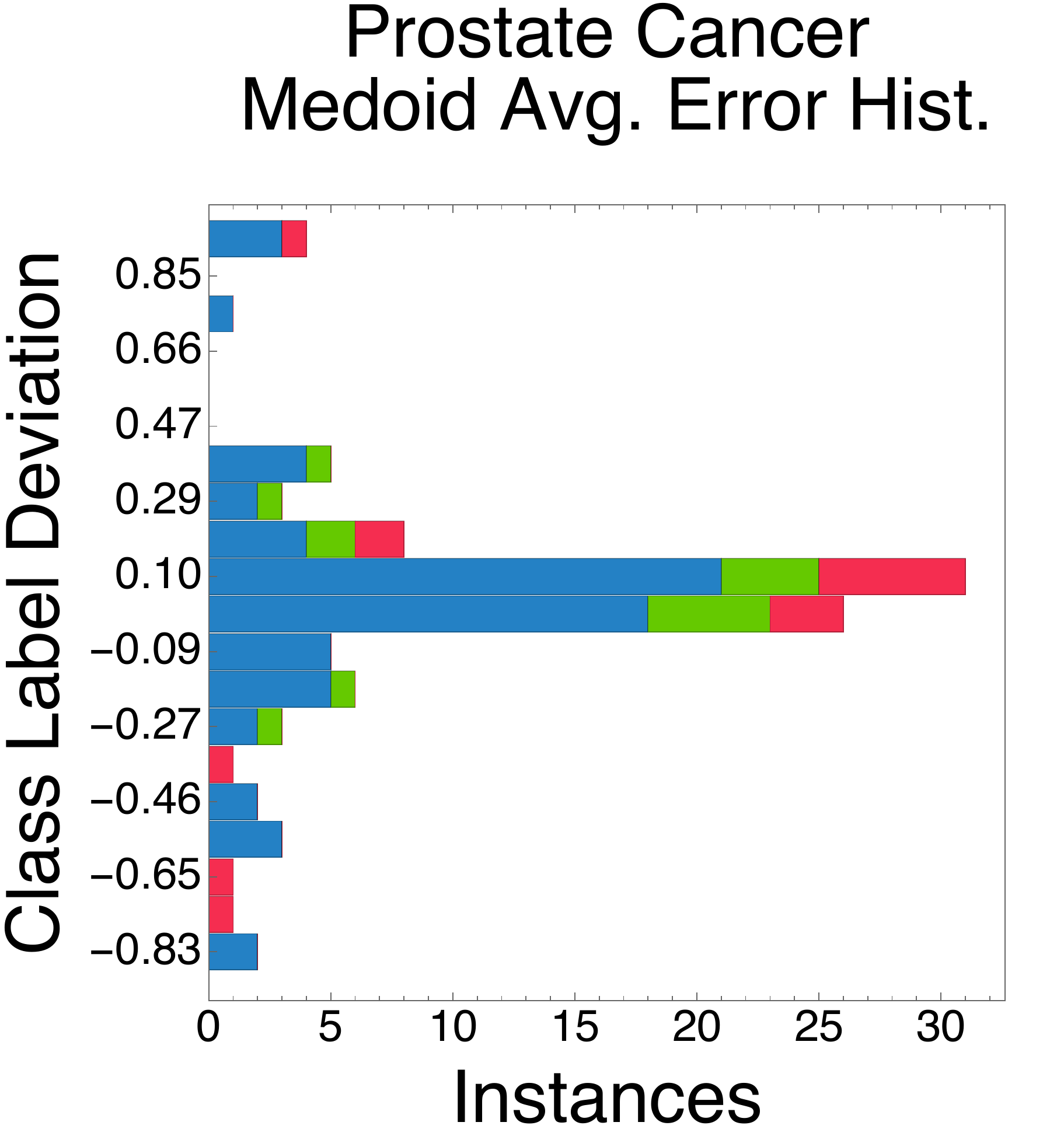} \hspace{-0.15cm} \includegraphics[height=1.0in]{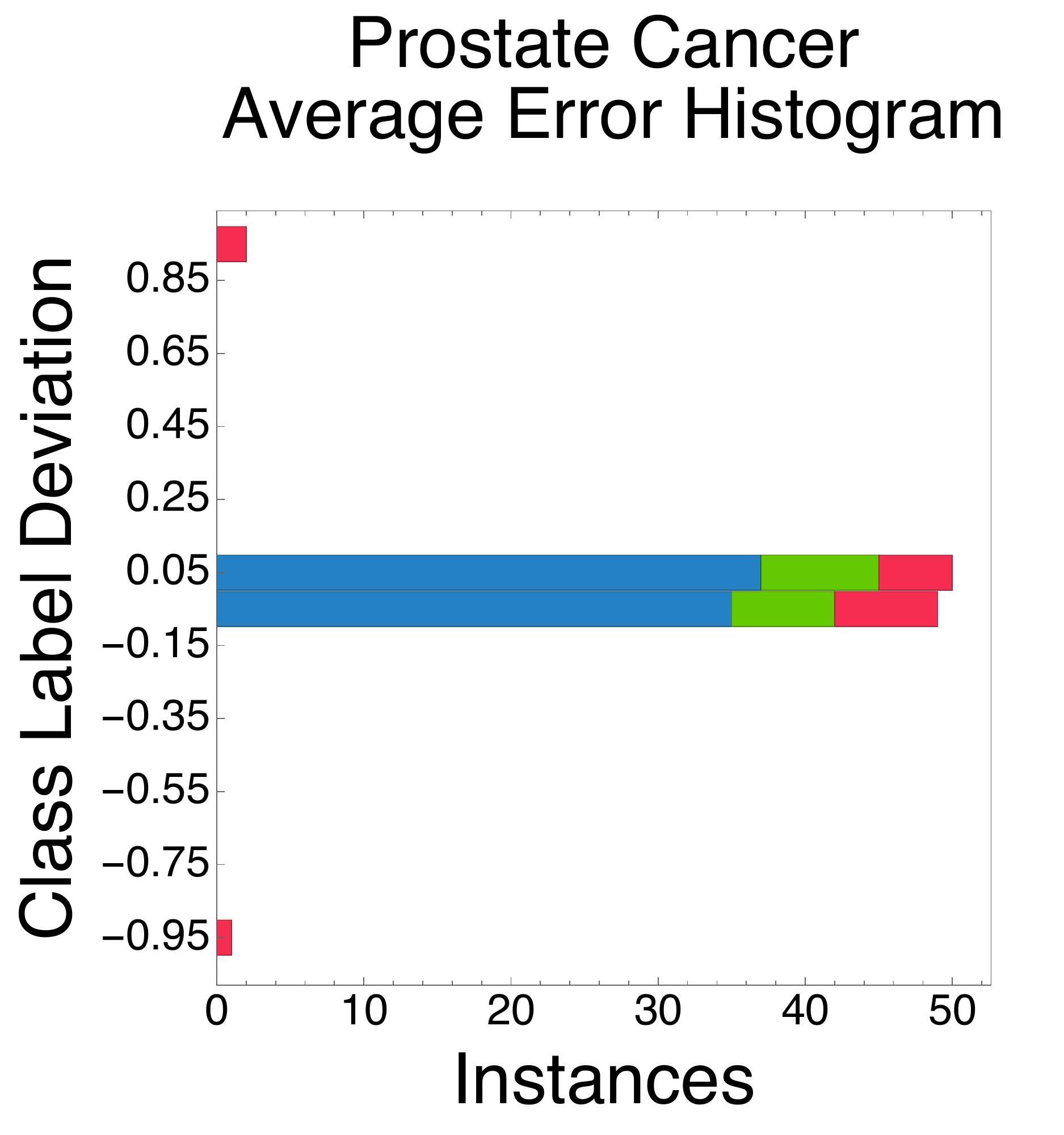} \hspace{-0.0675cm} \includegraphics[height=1.0in]{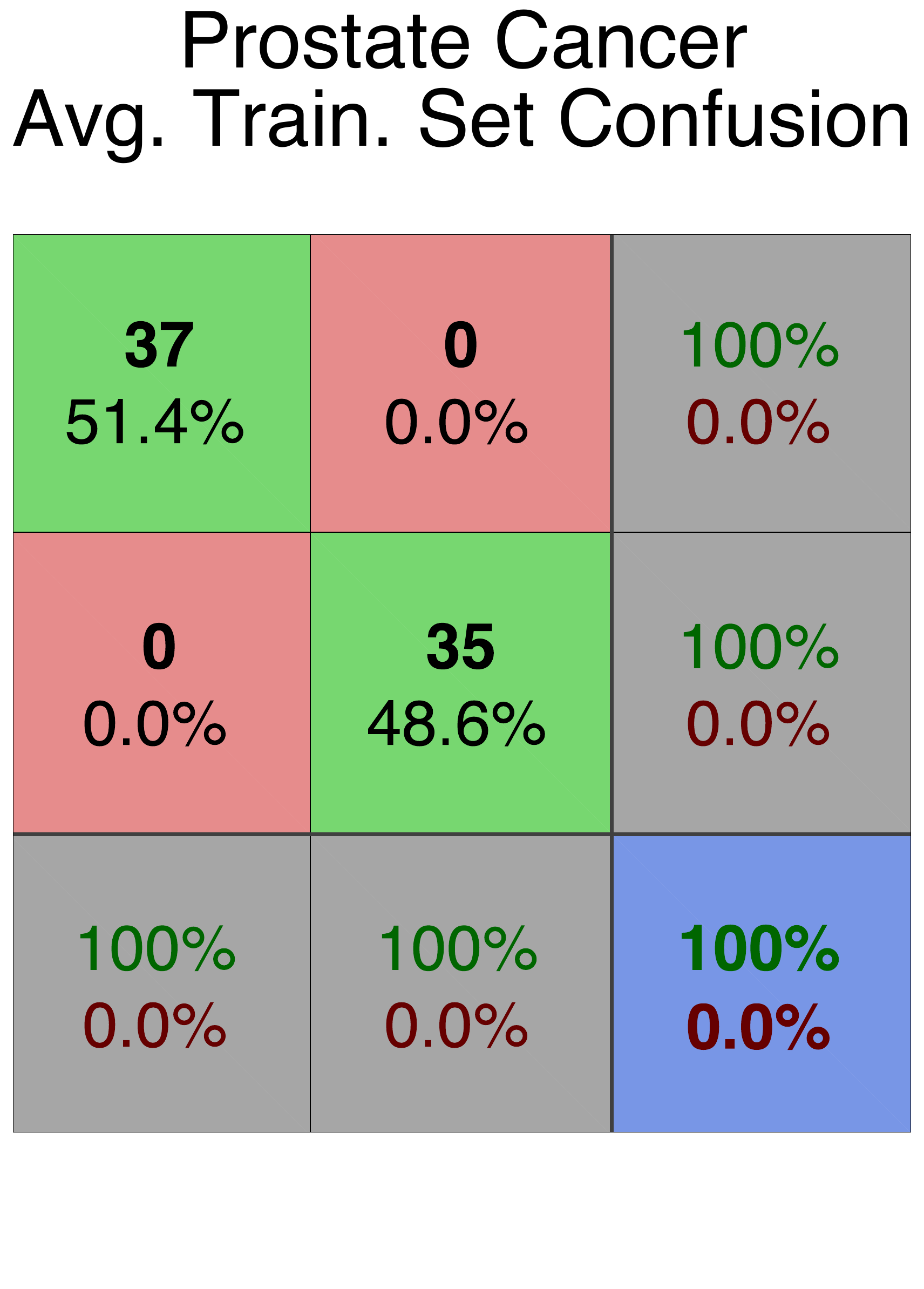} \hspace{0.075cm} \includegraphics[height=1.0in]{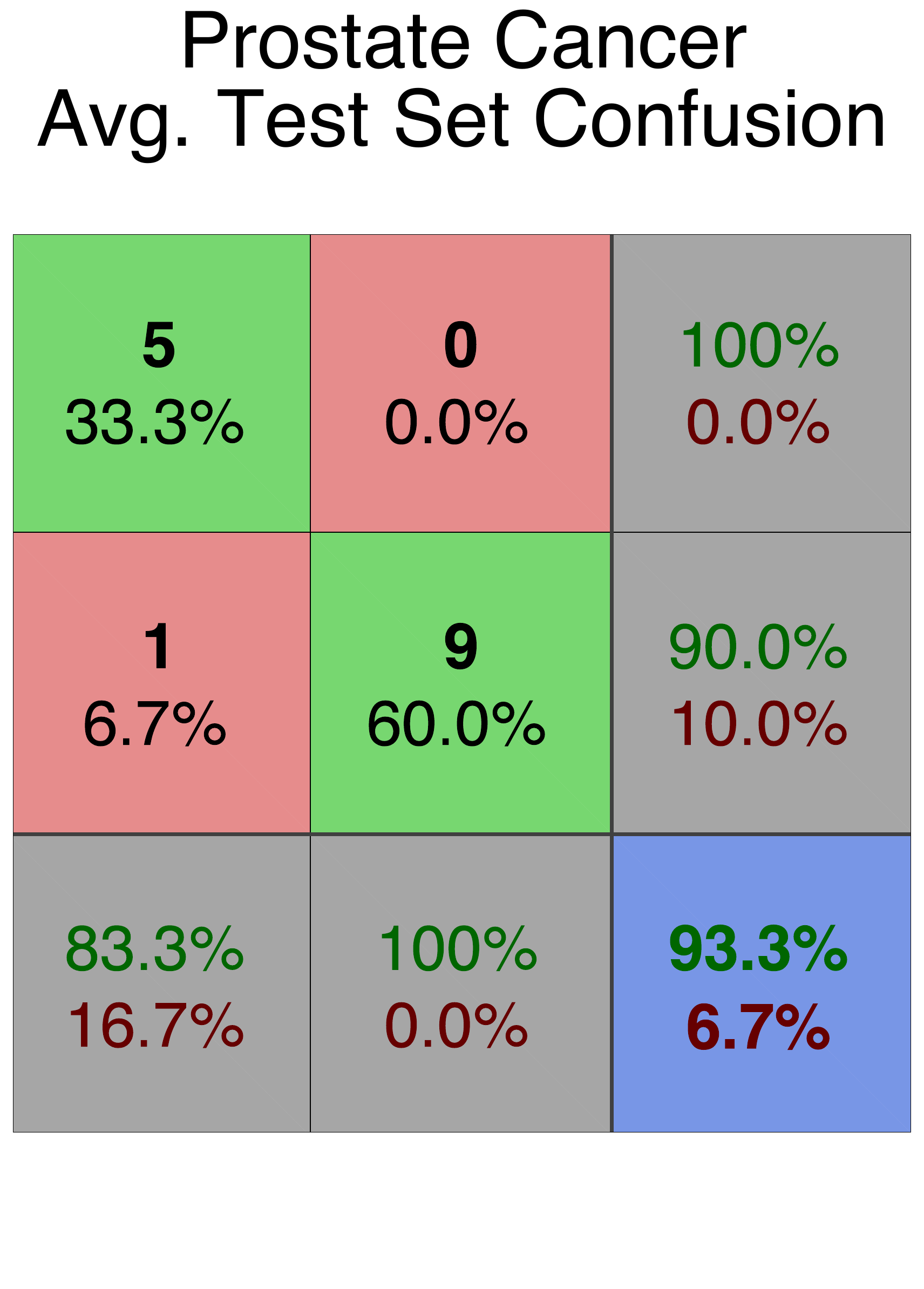} \hspace{0.075cm} \includegraphics[height=1.0in]{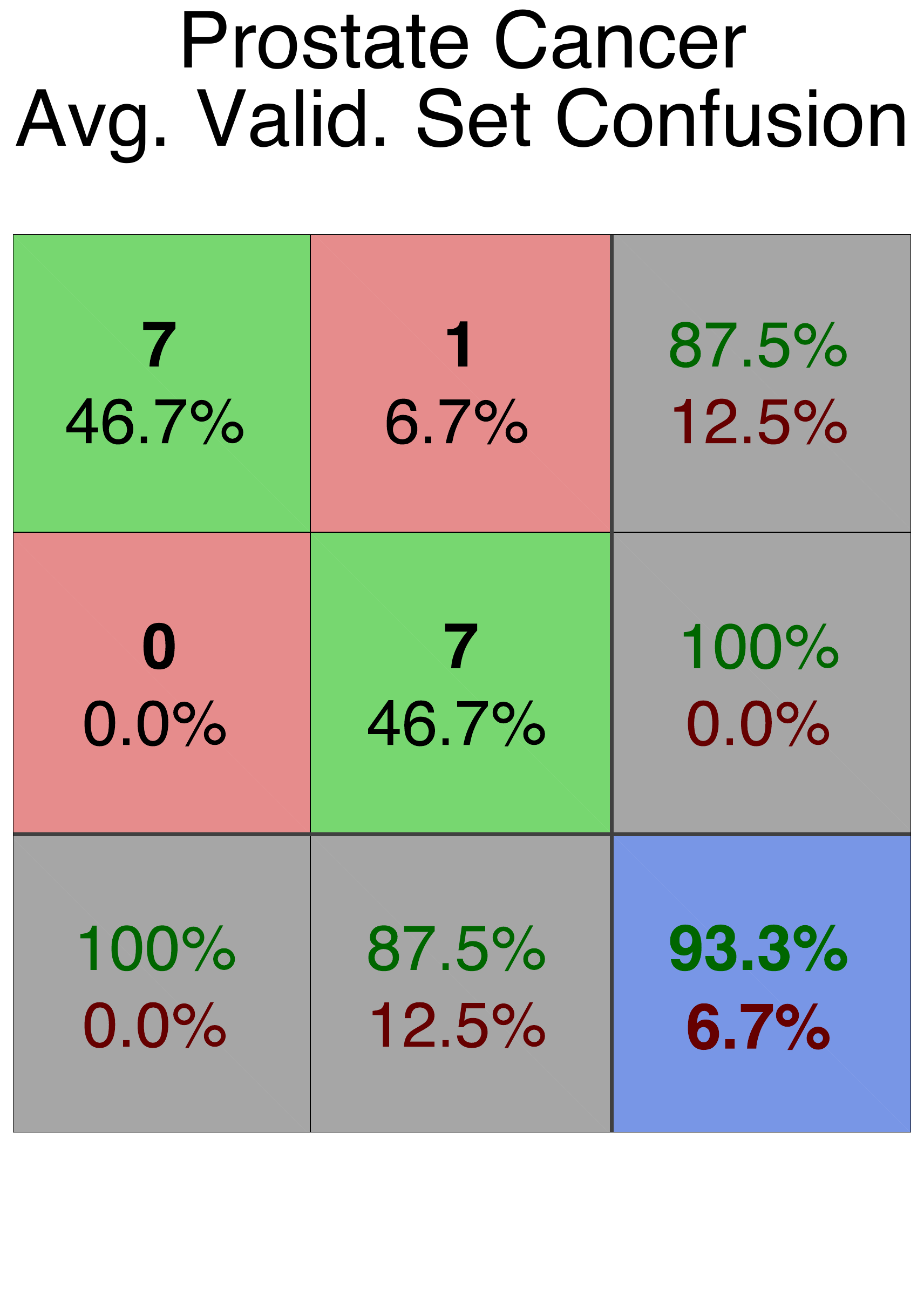}\vspace{-0.15cm}\\
   {\footnotesize (a)}\vspace{0.15cm}\\
   \includegraphics[height=1.0in]{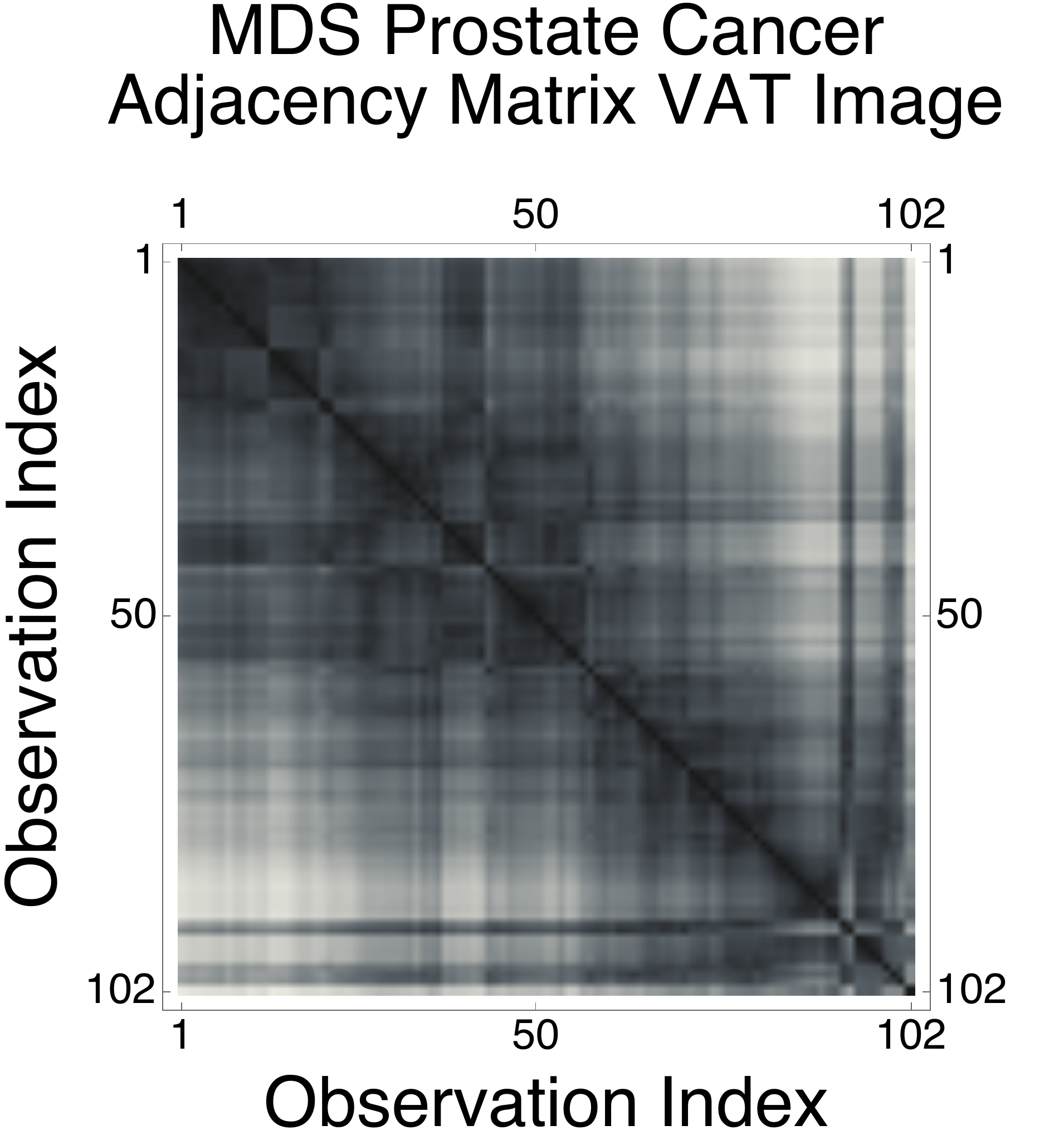} \includegraphics[height=1.0in]{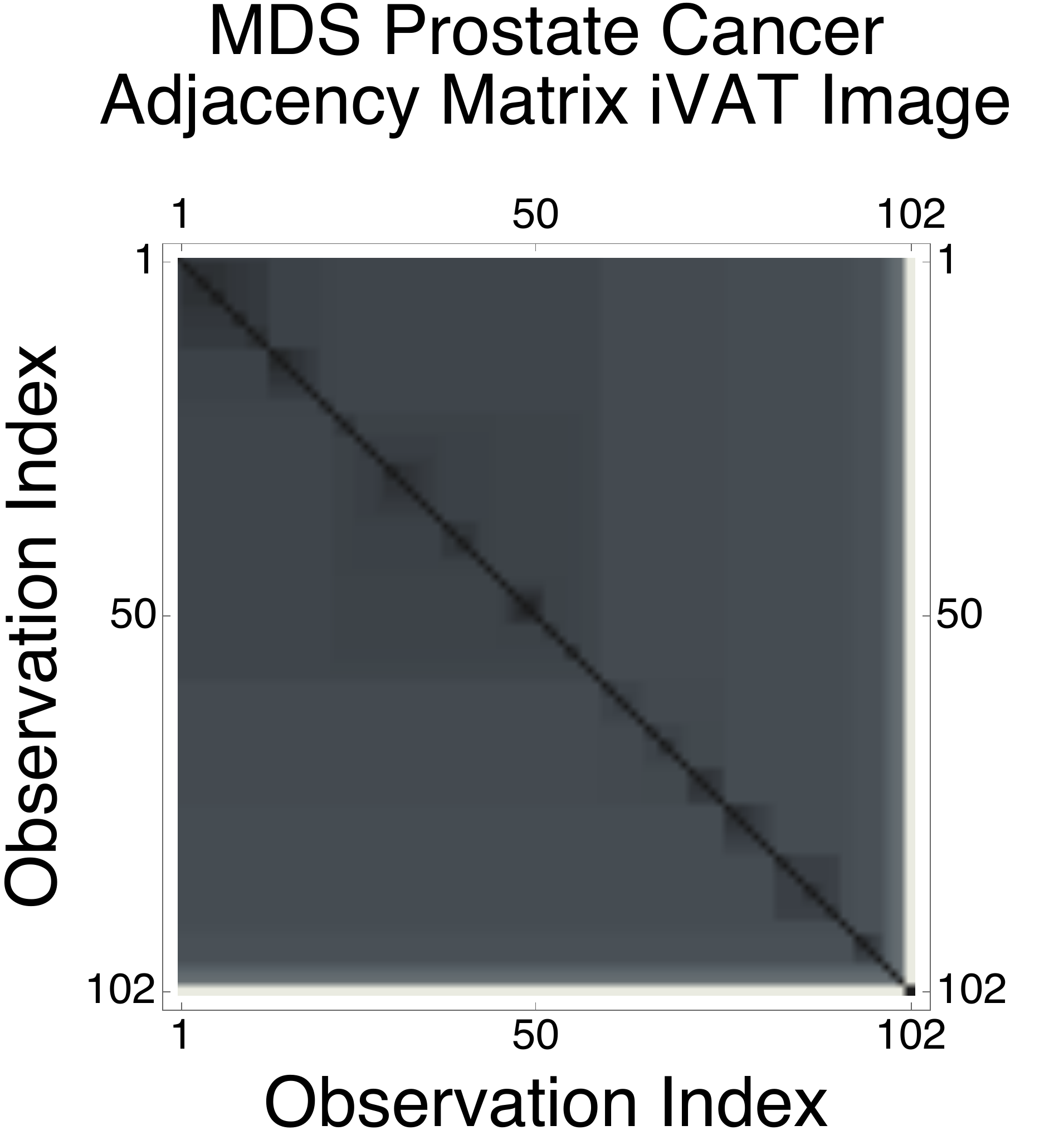} $\;\;$ \includegraphics[height=1.0in]{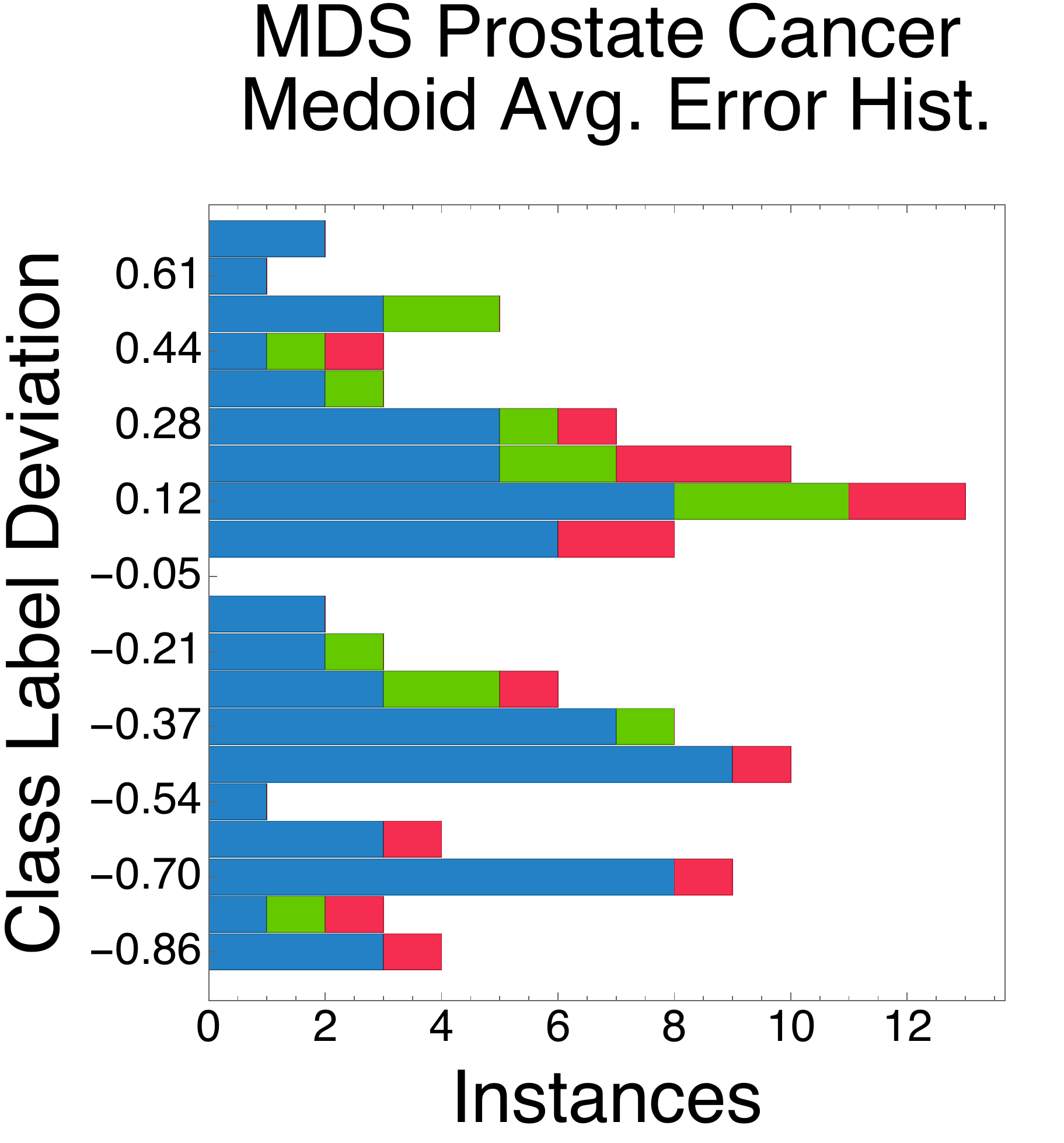} \hspace{-0.15cm} \includegraphics[height=1.0in]{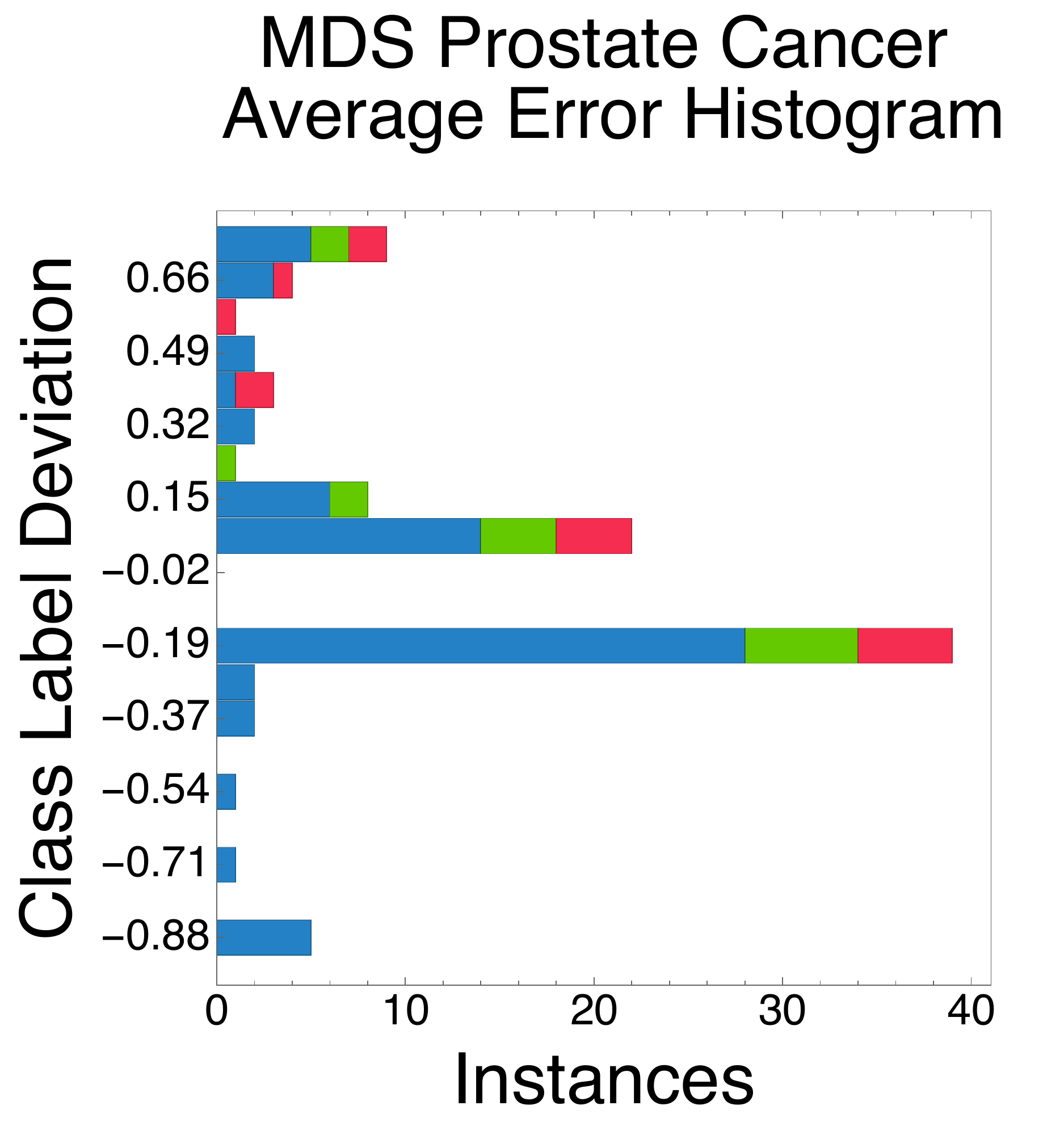} \hspace{-0.0675cm} \includegraphics[height=1.0in]{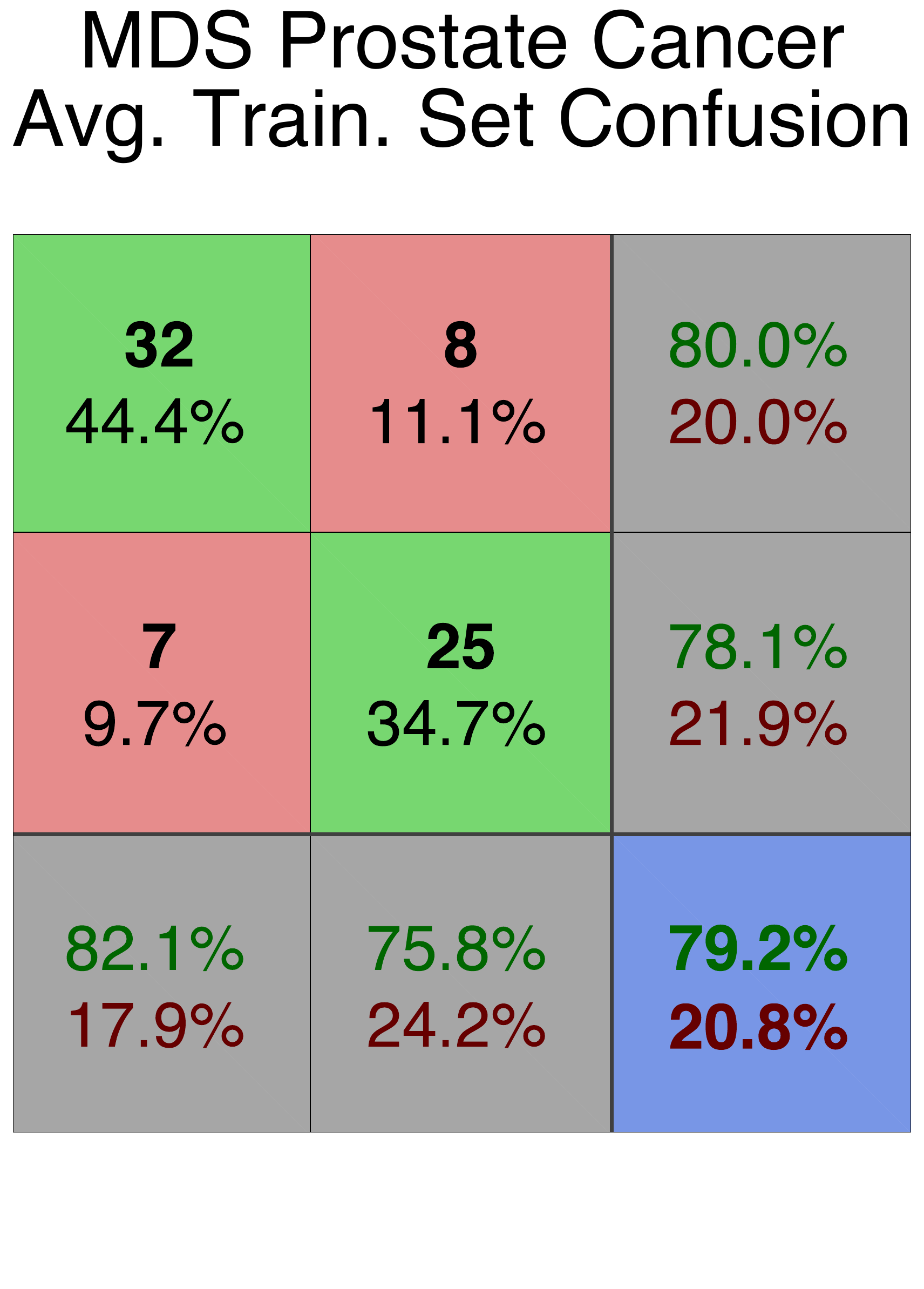} \hspace{0.075cm} \includegraphics[height=1.0in]{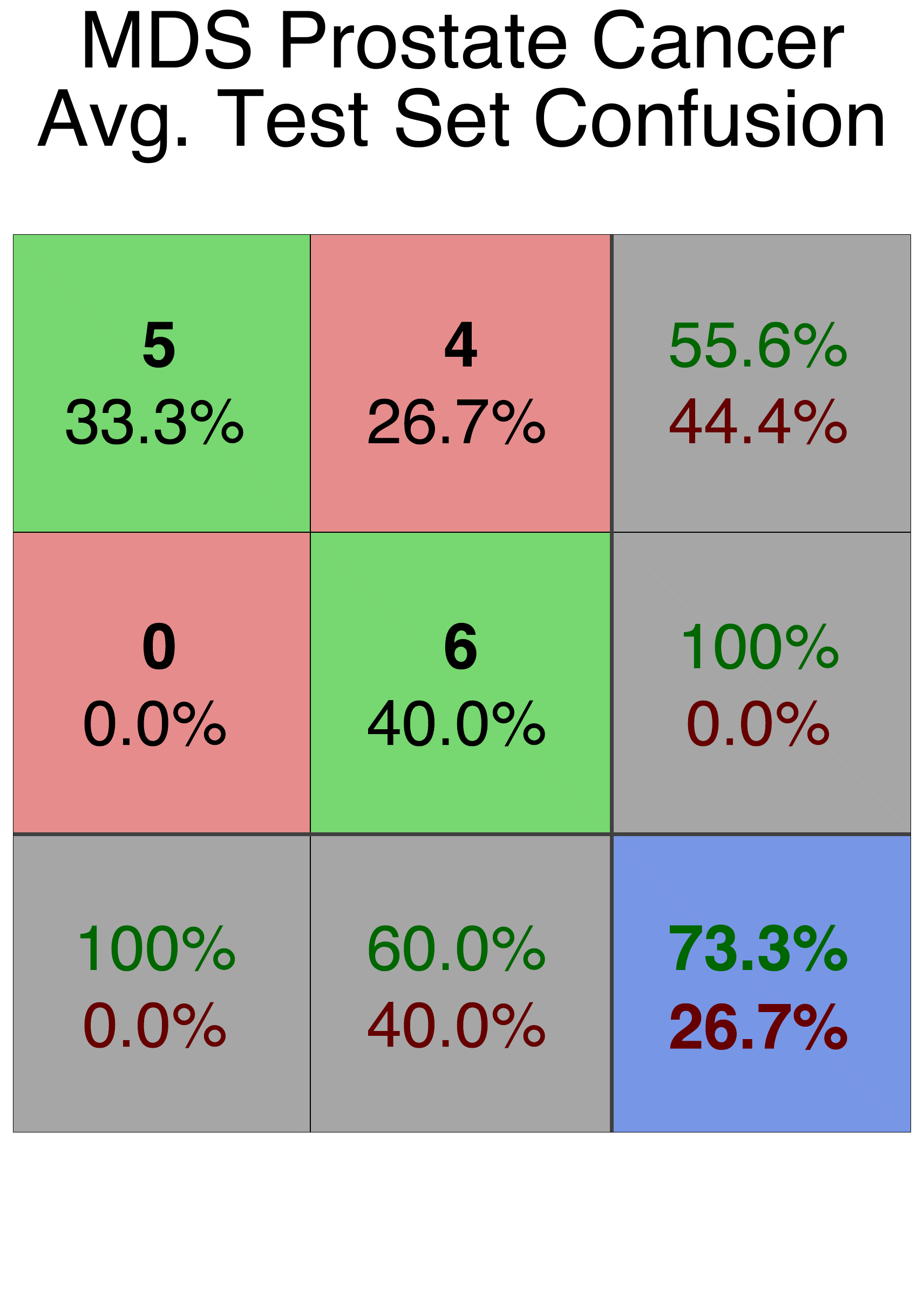} \hspace{0.075cm} \includegraphics[height=1.0in]{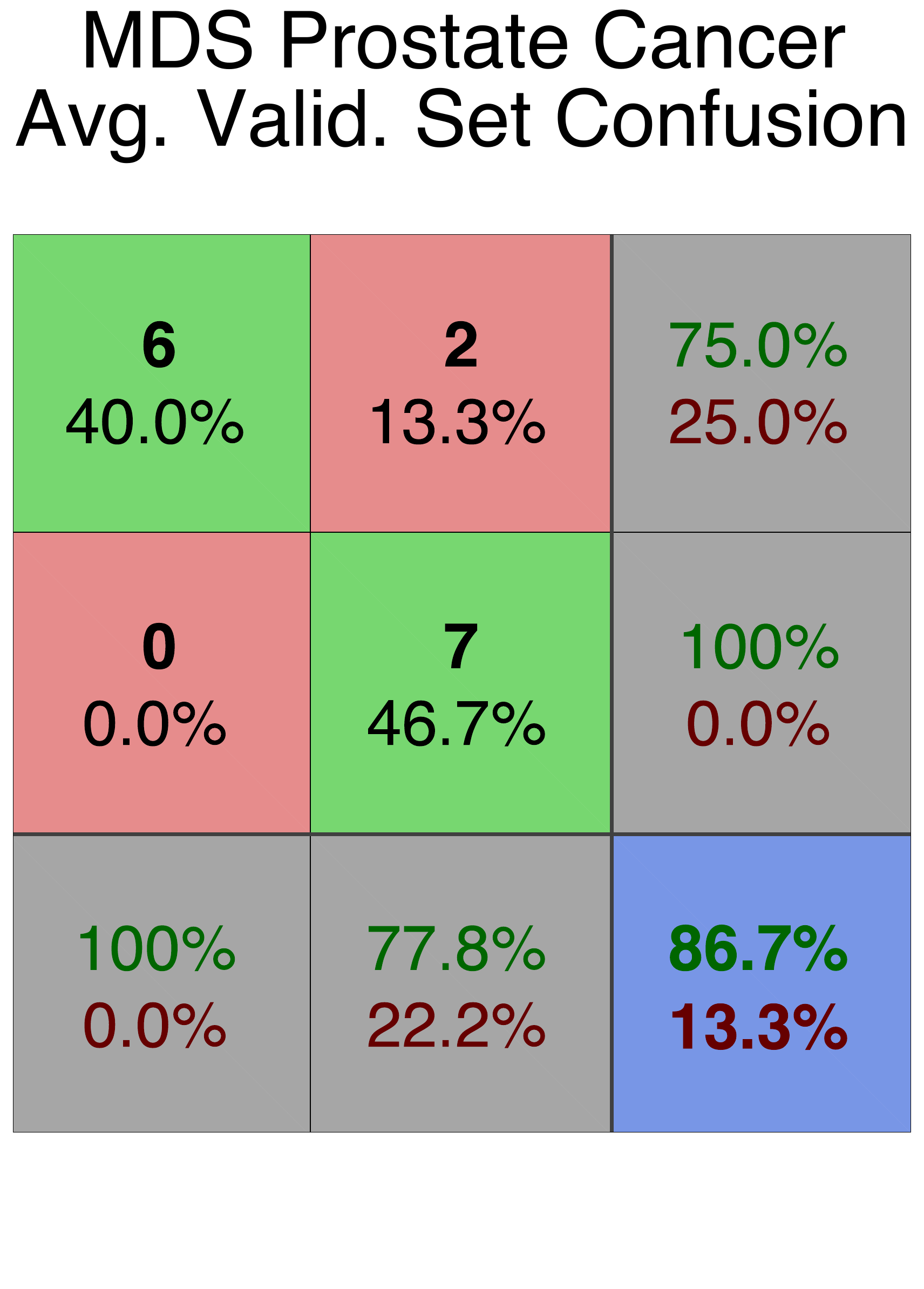}\vspace{-0.15cm}\\
   {\footnotesize (b)}\vspace{0.15cm}\\
   \includegraphics[height=1.0in]{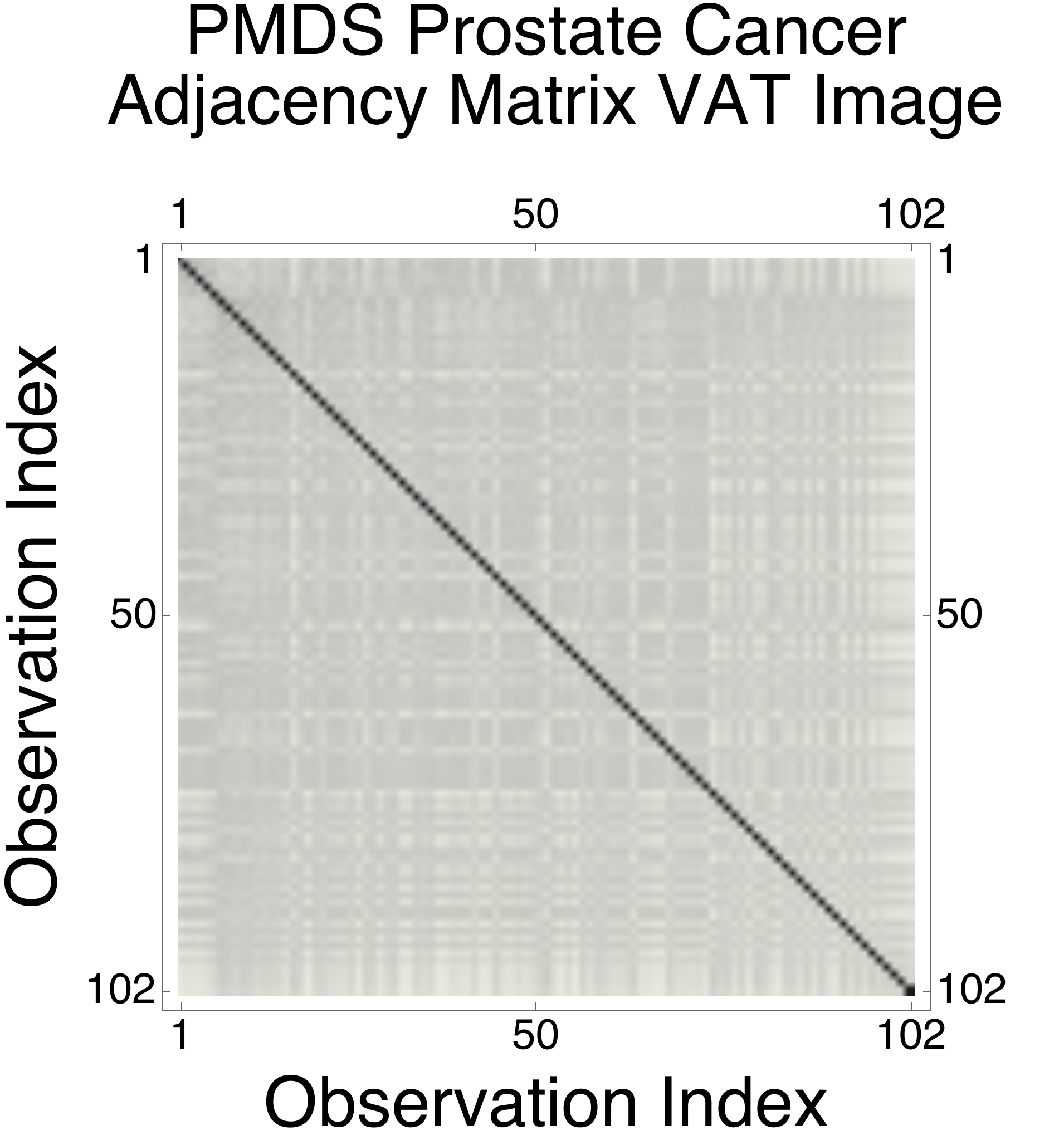} \includegraphics[height=1.0in]{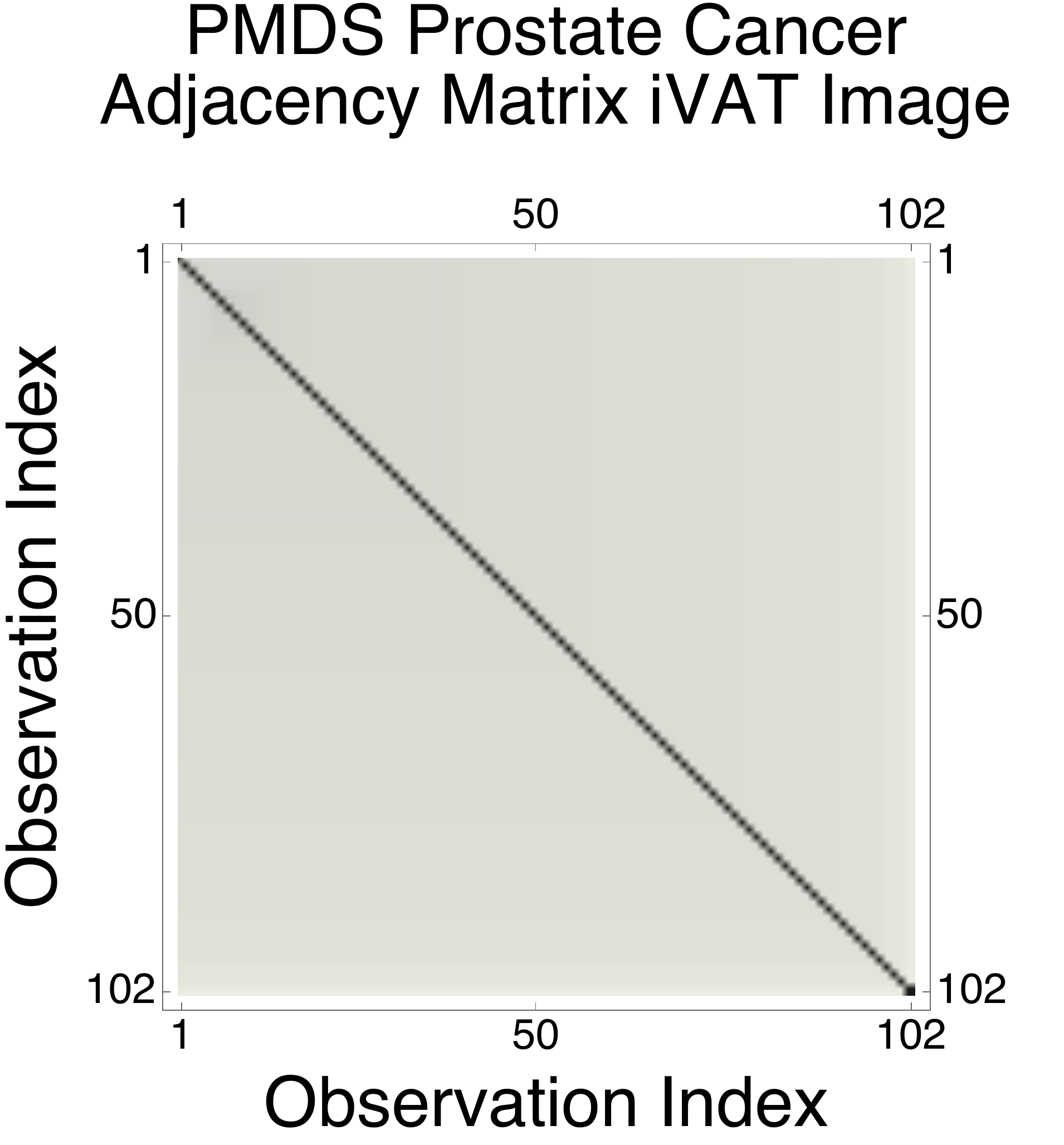} $\;\;$ \includegraphics[height=1.0in]{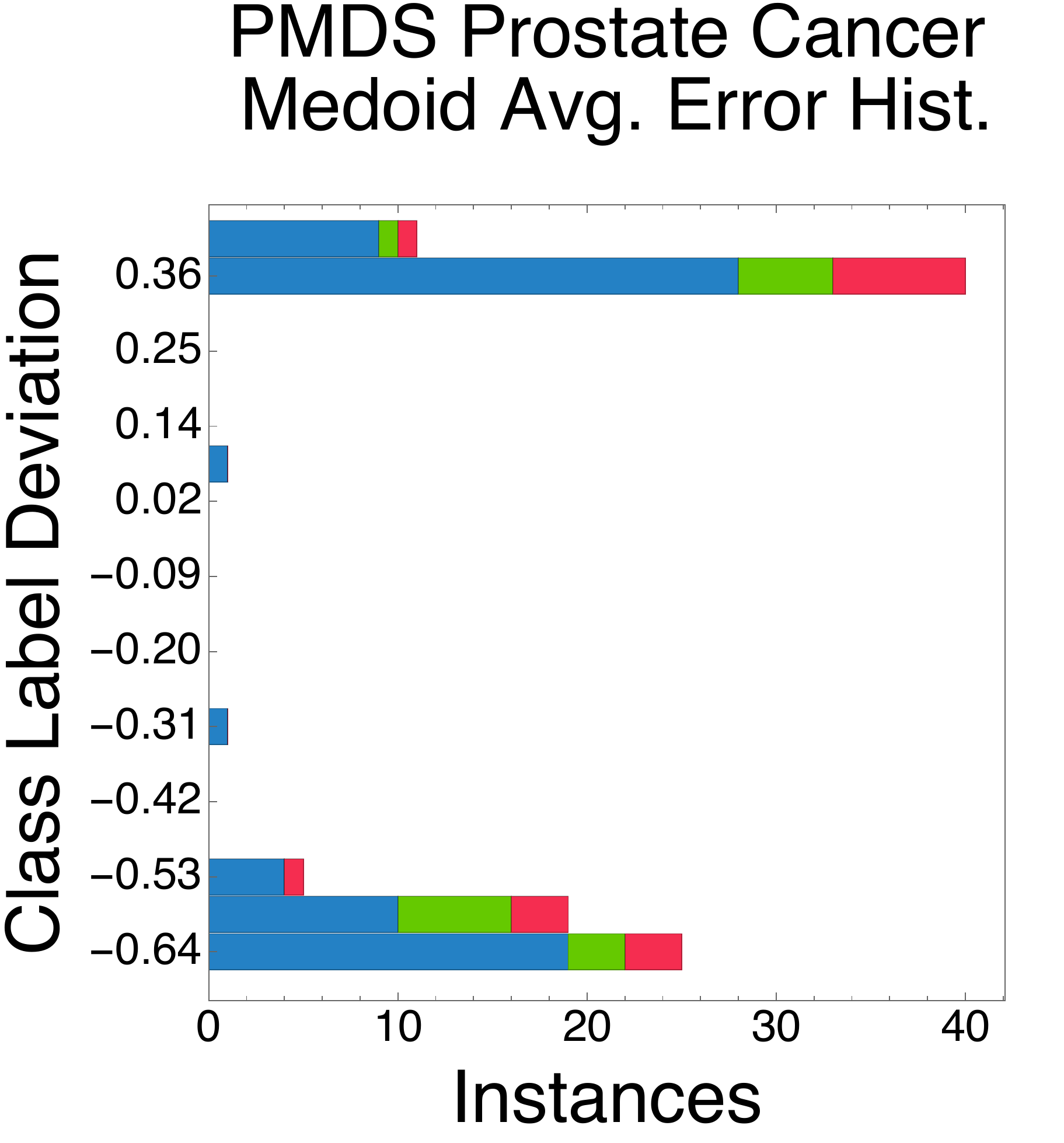} \hspace{-0.15cm} \includegraphics[height=1.0in]{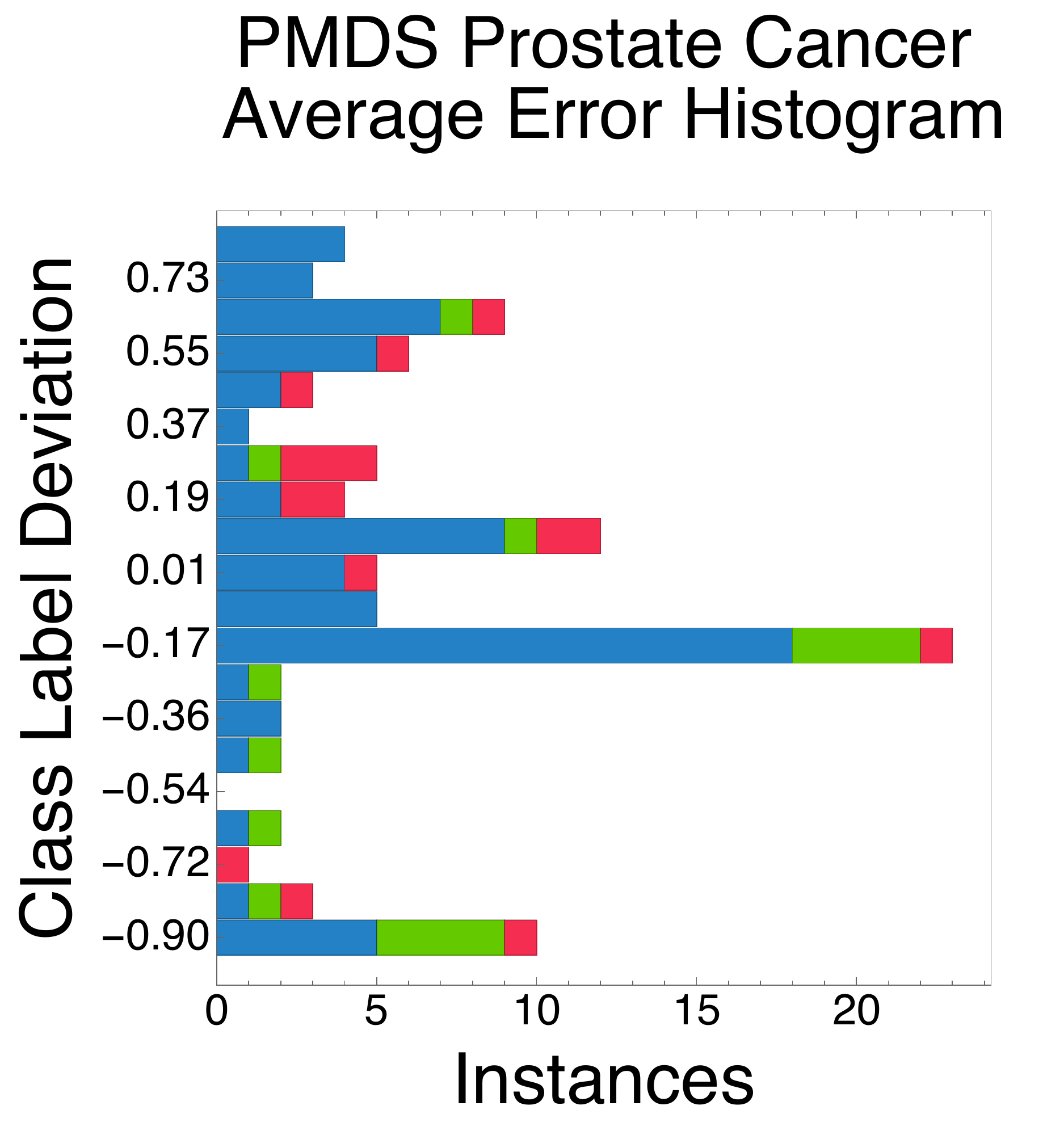} \hspace{-0.0675cm} \includegraphics[height=1.0in]{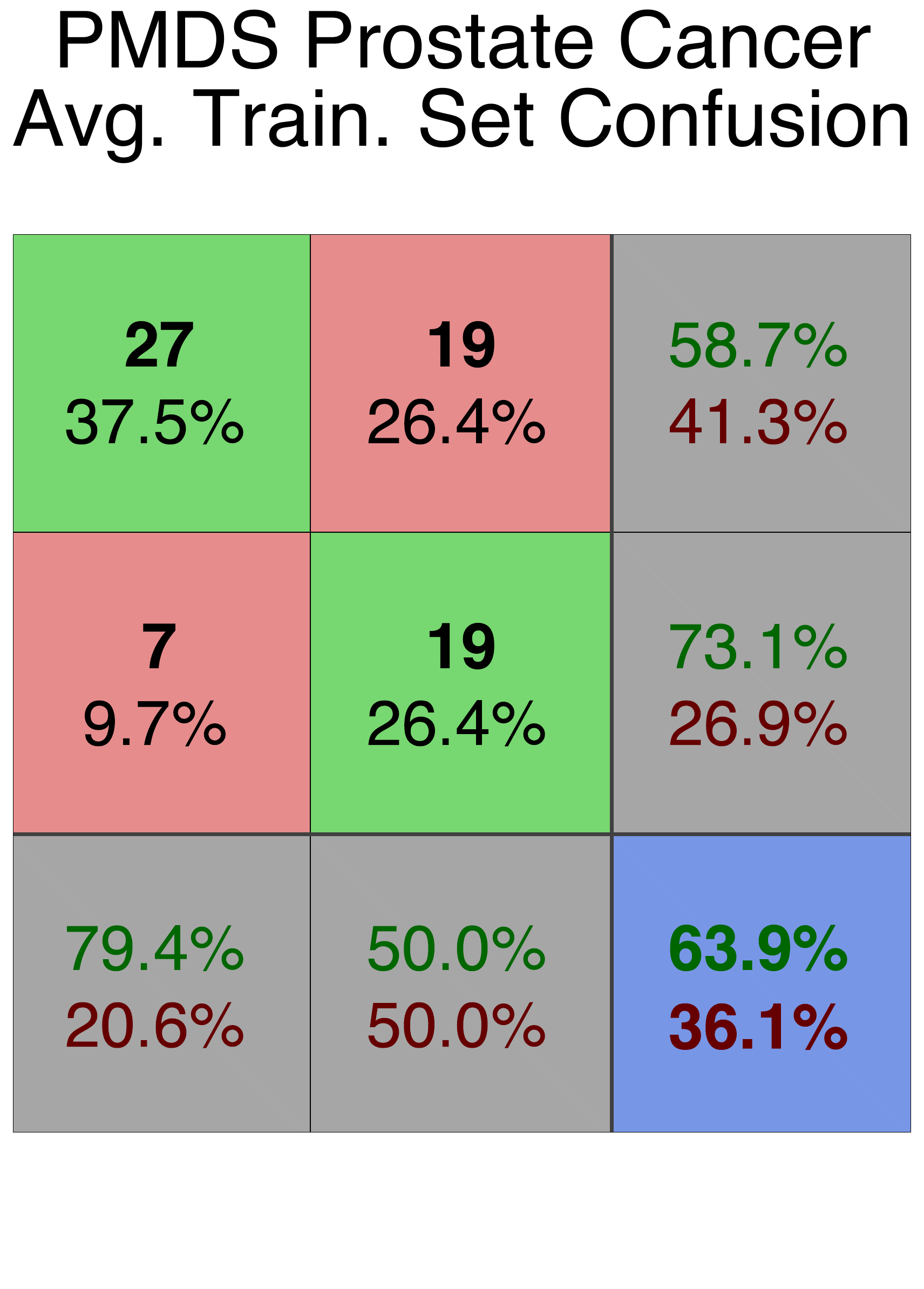} \hspace{0.075cm} \includegraphics[height=1.0in]{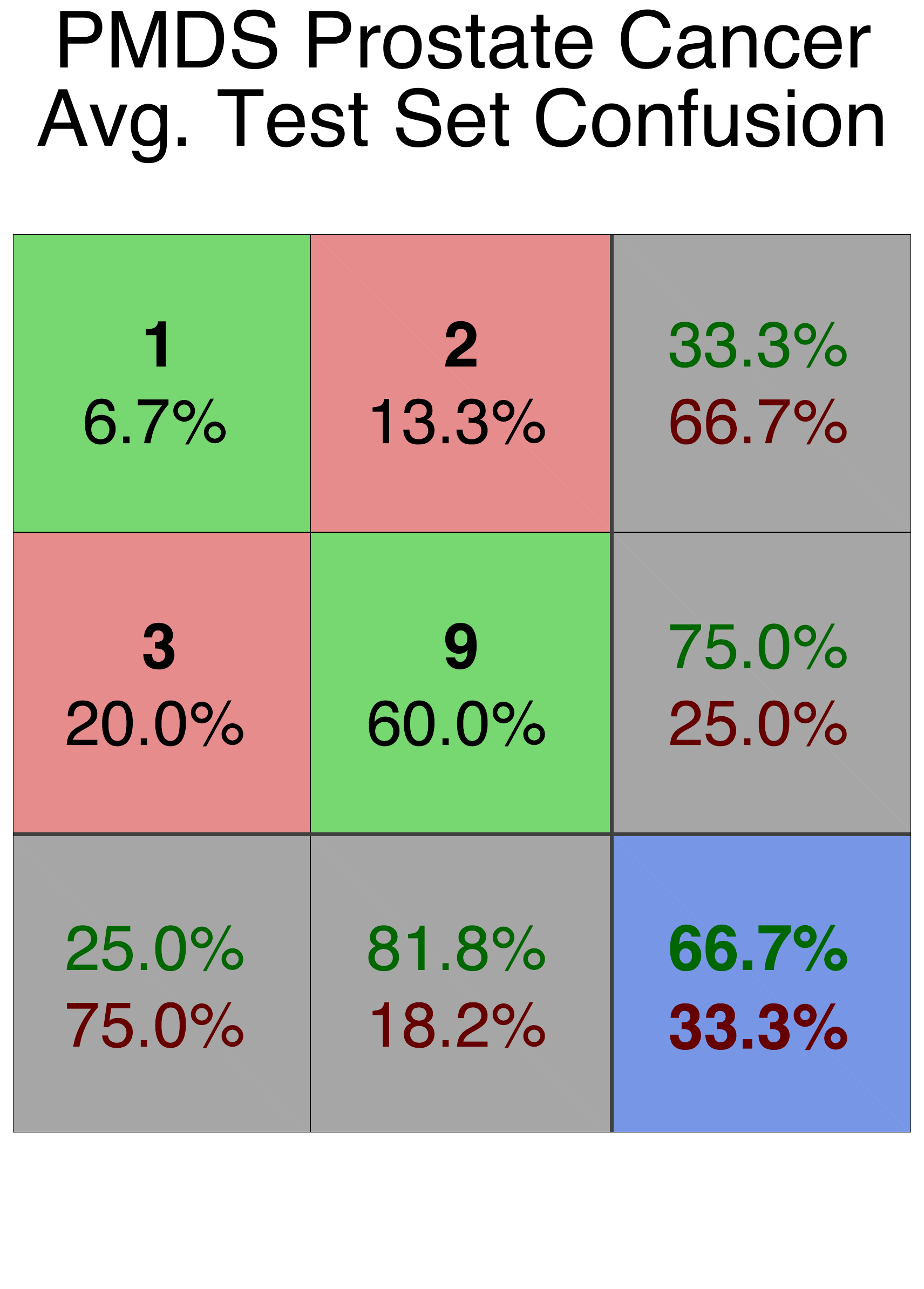} \hspace{0.075cm} \includegraphics[height=1.0in]{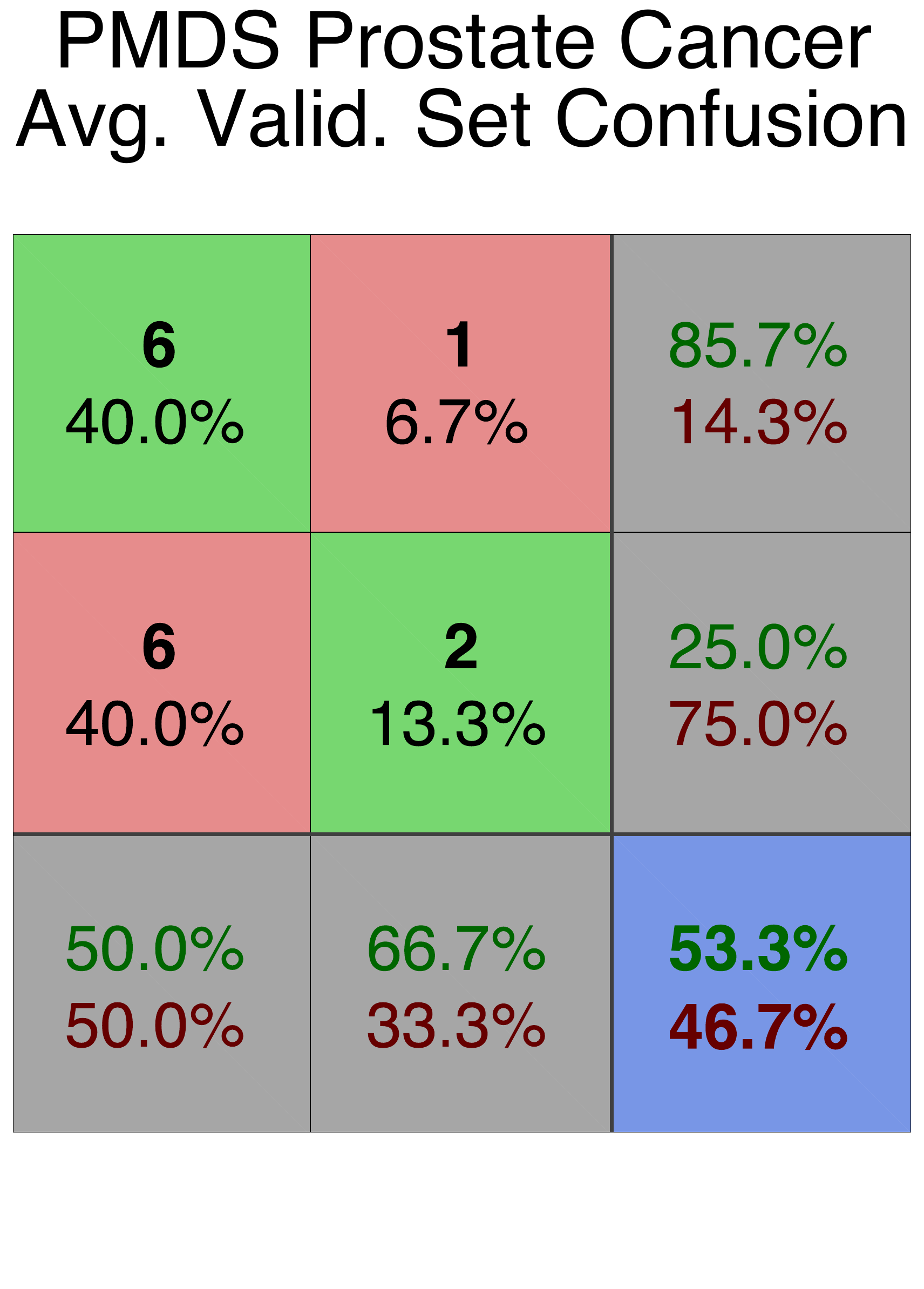}\vspace{-0.15cm}\\
   {\footnotesize (c)}\\
   \end{tabular}
\caption*{Figure 3.2: Classification results for a purely graph-based dataset compared to transformed versions of it.  We consider a dataset, $R_\textnormal{PC-102}$, based on applying a Bayesian similarity measure to a series of oligonucleotide microarray expression profiles.  See the caption in figure 3.1 for a description of the plots.  \vspace{-0.4cm}} 
\end{figure*}


The VAT and iVAT images for $R_\textnormal{PC-102}$, in figure 3.2(a), reveal that there is less structure in the relationships  than in $R_\textnormal{HG-194}$, owing to the heterogeneity of prostate tumors.  The iVAT plot is mostly uniform, suggesting that the relationships correspond to a set of feature vectors that are arranged in single distribution.  The VAT plot indicates that this single distribution has some substructure, shown by the dark block in the bottom-right corner, which is connected with some of the tumor prostate samples.  Despite the fact that there is likely significant overlap between the classes, the graph-based RBF networks were able to separate between tumor and non-tumor samples well.  The error histogram plot shows that predicted network responses diverged little from the supplied labels across the Monte Carlo simulations.  The confusion matrices demonstrate that no mistakes were made, on average, in classifying the training set while only one sample was often misclassified in both the testing and validation sets.  In approximately a third of the simulations, there were no classification errors in any of the $R_\textnormal{PC-102}$ datasets.

Applying both CMDS and PMDS to $R_\textnormal{PC-102}$ again had the effect of disrupting the underlying relationship structure.  When using CMDS, some of the intra-class relationships have been increased while the inter-class relationships were decreased in $R_\textnormal{PC-102}^\textnormal{CMDS}$ compared to $R_\textnormal{PC-102}$.  More complicated decision surfaces were needed to segment the two classes, which were not easy to construct given the small amount of RBF prototypes.  The average recognition rates hence dropped by anywhere from about 6\% to 20\% on the training, test, and validation sets compared to $R_\textnormal{PC-102}$, as shown in figure 3.2(b).  PMDS had a worse impact than CMDS on performance: recognition rates fell by approximately 36\% to 40\% with respect to $R_\textnormal{PC-102}$, as indicated by the confusion matrices in figure 3.2(c).  This was a byproduct of the underlying vector realization for $R_\textnormal{PC-102}^\textnormal{PMDS}$ was non-linearly spread out and the classes mixed, which can be inferred from the associated VAT and iVAT images.  A great number of RBF prototypes would hence be needed to construct a reasonable decision boundary for $R_\textnormal{PC-102}^\textnormal{PMDS}$.

Since there are vector realizations for these dataset transformations, the recognition results for $R_\textnormal{HG-194}^\textnormal{CMDS}$ and $R_\textnormal{HG-194}^\textnormal{PMDS}$ along with $R_\textnormal{PC-102}^\textnormal{CMDS}$ and $R_\textnormal{PC-102}^\textnormal{PMDS}$ can be interpreted as being equivalent to those for conventional, vector-based RBF networks.  The above results indicate that the performance of the graph-based RBF networks is superior, for these graphs, to transforming weighted graphs into feature vectors and then training vector-based RBF networks.

CMDS and PMDS are not the only multi-dimensional scaling techniques that can create feature vectors from $R_\textnormal{HG-194}$ and $R_\textnormal{PC-102}$.  We also considered landmark multi-dimensional scaling (LMDS) \cite{deSilva-coll2002}, which creates an initial feature-space embedding using a subset of the observations then approximates the remainder of the embedding.  LMDS is useful for large datasets, however, $R_\textnormal{HG-194}$ and $R_\textnormal{PC-102}$, are relatively small, so they did not benefit.  On average the results were highly sensitive to the initial embedding and its size, causing the classification performance to vary greatly: the performance was about 5\% to 30\% worse compared to CMDS.  Additionally, we applied our approach from \cite{SledgeIJ-conf2010a} to optimally project the adjacency matrices onto the set of Euclidean distance matrices before using CMDS \cite{BoyleJP-coll1986}.  The recognition rates improved by anywhere from about 3\% to 7\% over CMDS and 27\% to 40\% over PMDS, due to fewer edits being made to the relationships than either scheme.  

All of the aforementioned transformation schemes project the adjacency matrices onto the set of the Euclidean distance matrices before finding a realization.  Arbitrary graphs, though, may have edge weights that are metric distances yet not Euclidean, so converting them to Euclidean distances can perturb the class statistics and make recognition more challenging.  It would therefore be prudent to discern a metric from the edge weights, versus imposing an arbitrary one.  Toward this end, we considered the metric learning approach of Xing et al. \cite{XingEP-coll2002a}, which attempts to construct a positive semi-definite matrix such that the Mahalanobis distance between pairwise vectors matches, as well as possible, the original edge weights.  An embedding can be found via traditional multi-dimensional optimization schemes when using this learned metric.  We found that such a scheme led to improvements of approximately 6\% to 9\% over CMDS and 30\% to 42\% over PMDS.

Our reformulation theory relies on converting and iteratively updating the observation-prototype distance that is used when propagating the relationships through the network.  There are other ways, though, that we could have reformulated the observation-prototype distance.  One example is by constraining the RBF prototypes to be data medoids: the observation-prototype distance could hence be found by indexing the adjacency matrix.  Only minor alterations would be needed for the network parameter update equations.  This approach, however, tended to perform poorly.  In figures 3.1(a)--3.1(c) and 3.2(a)--3.2(c), we show how the average response deviation would change when medoids are used.  For both these datasets, constraining the prototypes to belong to the observation space led to an increased divergence between the predicted network response and the true label.  For $R_\textnormal{HG-194}$, the recognition rates declined by 2\% to 6\%, while for $R_\textnormal{PC-102}$ they dropped by 3\% to almost 10\% compared to when we used the exact reformulation theory developed in the previous section.  When applying multi-dimensional scaling, the issue was exacerbated: an 18\% to 61\% decrease in performance was observed.

To provide some context for these results, we compare against alternate classification schemes.  The first is the weighted $k$-nearest-neighbor classifier, which is suitable for both vector- and graph-based data.  When applied to $R_\textnormal{HG-194}$, the best-performing models were 4\% to 12\% worse, on average, than our RBF networks, depending on the number $k$ of neighbors used in the voting process.  Using three to five neighbors tended to yield the best classification accuracy when relying on the random-data-splitting strategy described at the beginning of this section; we tested up to fifteen neighbors.  When applied to $R_\textnormal{HG-194}^\textnormal{CMDS}$ and $R_\textnormal{HG-194}^\textnormal{PMDS}$, the average classification performance of the best-performing models decreased by 5\% and 15\%, respectively, compared to when graph-based RBF networks were applied to $R_\textnormal{HG-194}$.  Similar results were encountered for $R_\textnormal{PC-102}$.  The recognition rates for the entire dataset dropped by 6\% to 14\% when using weighted $k$-nearest-neighbor models compared to when graph-based RBF networks were employed.  They also dropped, on average, for $R_\textnormal{PC-102}^\textnormal{CMDS}$ and $R_\textnormal{PC-102}^\textnormal{PMDS}$ by 24\% and 47\%, respectively, versus when RBF networks were applied to $R_\textnormal{PC-102}$.  Three to six neighbors appeared to give the best results for this dataset.

We consider two other classifiers, support vector machines and random forests, neither of which is natively defined for graphs.  In the support-vector-machine case, we relied on Gaussian kernels with automatically tuned kernel variances.  Sequential minimum optimization was employed for solving the quadratic optimization problem.  For this classifier, the average recognition rates were 7\% and 13\% lower, respectively, when using the vector realization associated with $R_\textnormal{HG-194}^\textnormal{CMDS}$ and $R_\textnormal{HG-194}^\textnormal{PMDS}$ than when graph RBF networks were applied to $R_\textnormal{HG-194}$.  Likewise, the average recognition rates dropped by 22\% to 42\%, respectively, when applied to the vector realizations associated with $R_\textnormal{PC-102}^\textnormal{CMDS}$ and $R_\textnormal{PC-102}^\textnormal{PMDS}$ compared to when graph RBF networks operated on $R_\textnormal{PC-102}$.  The performance of random forests mirrored that of support vector machines; it deviated only by three percent at most.



\vspace{0.05cm}{\small{\sf{\textbf{Methodological Discussions and Insights.}}}} These results showcase the advantage of our graph-based RBF network.  That is, no modification of the weighted graphs is often necessary before classification can occur, which preserves the original class statistics.  Vector-based RBF networks, in contrast, require the construction of feature vectors for data that is inherently expressed as weighted graphs.  This graph-to-vector conversion can alter the latent relationship structure, thereby aversely changing the class structure and impeding recognition in some situations.  Such an outcome stems from the information loss during the transformation process.

More specifically, CMDS non-optimally projects non-Euclidean adjacency matrices onto the set of Euclidean distance matrices \cite{SledgeIJ-conf2010a}.  This is carried out by an affine projection onto the cone of positive-semi-definite matrices, which has a closed form solution for the $L_2$-based Hilbert-Schmidt norm.  The use of an $L_2$-type norm in CMDS means that many of the edge weights can be modified by a small to large amount to ensure that the matrix is Euclidean, though, even if only a few of the pairwise relationships are slightly non-Euclidean.  This occurred for $R_\textnormal{HG-194}^\textnormal{CMDS}$, as gene sequences for the two largest families became increasingly related even thought they were not in $R_\textnormal{HG-194}$.  A larger number of RBF centers were hence needed for $R_\textnormal{HG-194}^\textnormal{CMDS}$, in contrast to $R_\textnormal{HG-194}$, to counteract the changes; for many simulations, the limit on RBF prototypes was reached.  A similar situation happened for $R_\textnormal{PC-102}^\textnormal{CMDS}$.  However, for $R_\textnormal{PC-102}^\textnormal{CMDS}$, there was a greater performance drop than for $R_\textnormal{HG-194}^\textnormal{CMDS}$, despite that the absolute deviation between $R_\textnormal{PC-102}$ and $R_\textnormal{PC-102}^\textnormal{CMDS}$, after normalizing by the number of weights, was lower than for $R_\textnormal{HG-194}$ and $R_\textnormal{HG-194}^\textnormal{CMDS}$.  This indicates that while the class statistics changed only slightly, a highly complicated decision boundary was necessary; the low maximal number of RBF prototypes for the $R_\textnormal{PC-102}$ datasets precluded its formation.

Other types of multi-dimensional scaling also have issues that hamper RBF-network-based classification.  PMDS also non-optimally transforms adjacency matrices into Euclidean distance matrices, albeit facilitated via a non-$L_p$-optimal projection onto the positive-semi-definite cone.  That is, PMDS linearly scales by the absolute smallest eigenvalue of a double-centered adjacency matrix versus zeroing negative eigenvalues as in CMDS.  This has the effect of non-linearly increasing both the inter- and intra-class relationships, spreading out the underlying vector realization.  The impact of applying PMDS are often worse than when CMDS is employed.  To counteract this issue, the RBFs should have a wide response, which can be effectuated by choosing a high kernel bandwidth value.  It may not always be possible, though, to arrive at a data-appropriate bandwidth value during training: either the learning rate may be too low or the consecutive gradients too small or both to implement this functionality before training concludes.  This happened in many of our simulations for both $R_\textnormal{PC-102}^\textnormal{PMDS}$ and $R_\textnormal{HG-194}^\textnormal{PMDS}$.  Alternatively, many RBF prototypes might be essential, and, in some instances, the amount of prototypes may have to approach the total number of training-set observations to achieve viable performance.  As with CMDS, the RBF prototype limits prevented the manifestation of a good decision surface.  Only local regions of feature vectors could be reliably classified as a consequence.

Our work in \cite{SledgeIJ-conf2010a} relies on optimally projecting, in an $L_2$-norm sense, adjacency matrices onto the set of Euclidean distance matrices.  This is accomplished by alternately projecting onto the convex cone of positive-semi-definite doubly-centered adjacency matrices and onto the convex subspace of symmetric, anti-reflexive matrices.  The iterates are hence driven towards the translated normal cone to both convex sets at the global-best solution, which implies the overall change in the adjacency matrix entries will be less than or equal to that of CMDS and PMDS.  The class statistics are thus better preserved, which explains the improved classification rates that we encountered for $R_\textnormal{HG-194}^\textnormal{AP}$ and $R_\textnormal{PC-102}^\textnormal{AP}$.  However, by virtue of using $L_2$-norm-based affine projections, many edge weights were modified to ensure that the few non-Euclidean entries correspond to Euclidean distances, which can prove detrimental in some situations.  A better approach would be to consider $L_1$-norm projections onto the convex sets, which can be approximated via Bregman projections.  Only those relationships that do not satisfy the Euclidean assumption would have the strongest chance of being changed in this case.  For weighted graphs that are entirely non-Euclidean yet satisfy the reformulation-theory properties, though, this process can still disrupt the class statistics, as information is lost in the transformation.  

Another option would be to simultaneously learn a suitable metric and a feature-space embedding of the relationships.  Such a scheme tended to yield the best performance for $R_\textnormal{HG-194}$ and $R_\textnormal{PC-102}$, since the relationships did not naturally correspond to isotropic metric distances.  Nonetheless, errors were still encountered.  While the metric-learning problem can be convex, the optimization problem for forming an embedding is generally not, so only locally-optimal solutions were uncovered.  Poor embedding initializations and poor starting parameter choices can also undermine both the vector realization quality and the search for a suitable metric.  In view of these issues and those for the other methodologies, it would hence be prudent to simply forgo constructing a vector realization and instead rely on graph-based RBF networks, as no data transformation would be needed for many datasets.  That is, a vector realization does not need to be found in order for the duality between vector- and graph-based RBF networks to be satisfied; the realization just must provably exist. 

The reformulation that we considered here is exact, in the sense that the graph- and vector-based networks are guaranteed to yield the same responses under general circumstances.  Non-exact reformulations are also possible, which trade off the response guarantee for a simpler RBF-prototype-datum distance expression; this reduces the complexity of the network update equations, as it can replace many terms with an adjacency matrix look-up.  However, as our simulations indicated, relying on non-exact reformulations can reduce the effectiveness of the graph-based RBF networks.  In the case when medoids are used, the prototypes are constrained to be nodes in the graphs, which may not be helpful in all situations, such as when dealing with relationships that correspond to manifold-distributed vectors.  The prototypes may need to exist outside the set of observations so that they can cover each class well and in a parsimonious fashion.  This is precisely the functionality offered by our exact reformulation.

\subsection*{\small{\sf{\textbf{4$\;\;\;$Conclusions}}}}\addtocounter{section}{1}

Traditional RBF networks are mappings from a vector-based feature space to some output space.  In this paper, we have provided a reformulation of vectorial RBF networks for graphs.  The two-layer graph-based RBF networks that we developed implement mappings from a relationship-based feature space, versus a vector space, to an output space.  Such networks are well suited for dealing with the purely graph-based datasets generated by a variety of processes and considered in a wide range of disciplines.

Our relational reformulation strategy relies upon the generalization of the prototype-observation distance.  We have shown that this distance can be re-written in terms of the chosen weights and entries from the relational matrix whenever the RBF prototypes are represented as weighted averages of the feature-space observations.

Provided that a vectorial realization of the relationships exists, both vector RBF networks and relational RBF networks are guaranteed to yield the same responses.  This is an attractive quality, as it implies that investigators do not need to explicitly construct a vector-based realization to train this class of neural networks.  For many types of graph-based observations, the underlying measure used to generate the weights for the graph may either not be known or not have a simple algebraic expression, which makes finding an exact realization quite challenging.  Even if the distance measure is known, the process for creating a realization may not be entirely accurate, which can perturb the natural structure of the observations.  This can have a major impact on the recognition rates, as we showed in our simulations.  It may also be computationally expensive to produce a realization, especially if the embeddings must be re-derived for each new batch of observations.

There are many ways in which RBF networks can be trained.  Here, we developed a hybrid, gradient-descent approach for tuning both the network weights and the RBF prototype positions.  Various heuristics, such as regularization, can be applied to this cost function with little modification of what we presented.  Higher-order approaches, which account for curvature information of the solution space, can also be considered.  However, regardless of whether we use first- or multi-order training approaches, we do not assume direct access to a realization, which makes altering the position of the RBF prototypes a tricky affair.  One viable option is to simulate the change in the vectorial RBF prototypes by iteratively adjusting the generalized prototype-observation distance measure.  This updated measure is then utilized for future forward passes of the relational matrix through the network. 

Our emphasis has been on the use of two-layer RBF networks.  They are the simplest topologies for illustrating the appropriateness of our reformulation.  We are not limited to only two layers, though.  In our future endeavors, we will demonstrate that a relational RBF input layer permits the construction of deep neural networks for non-image-based graph data.  We will also develop recurrent and time-delayed neural networks, with a relational RBF input layer, for addressing temporal, relationship-based observations.  In either of these cases, we will need to extend our gradient-descent-based strategy to handle the back-propagation of errors through multiple network layers.



\renewcommand*{\bibfont}{\raggedright}
\renewcommand\bibsection{\subsection*{\small{\sf{\textbf{References}}}}}
{\singlespacing\fontsize{9.75}{10}\selectfont \bibliography{sledgebib} \bibliographystyle{IEEEtran}}

\clearpage\newpage

\subsection*{\small{\sf{\textbf{A$\;\;\;$Vector- and Graph-based RBF Network Equivalency}}}}

\subsection*{\small{\sf{\textbf{A.1$\;\;\;$Dataset Descriptions}}}}

We utilize three datasets for empirically establishing the equivalency of the vector-based and graph-based RBF networks.  The datasets come from the University of California-Irvine Machine Learning Repository and correspond to binary classification problems:\vspace{0.05cm}
\begin{itemize}
\item[] \-\hspace{0.0cm}$R_\textnormal{BC-569}${\small{\sf{\textbf{: Breast Cancer Dataset.}}}} The first is the University of Wisconsin diagnostic breast cancer dataset, which is composed of features extracted from 569 fine-needle-aspirate images of either malignant or benign breast masses.  The two classes are unbalanced, as there are more benign mass observations than malignant.  From these images, 10 real-valued features were extracted to describe traits of the cell nuclei present in the image, which included geometric properties such as smoothness and compactness along with image-based properties that included the fractal dimension and texture information.  An additional 20 real-valued features correspond to supplemental statistics about the cell-nuclei that describe the standard error and largest-magnitude value observed.\vspace{0.025cm}
\item[] \-\hspace{0.0cm}$R_\textnormal{USC-435}${\small{\sf{\textbf{: Congressional Vote Dataset.}}}} The second is the US Congressional Voting Records dataset.  It contains voting entries for 267 Democratic and 168 Republican Congressmen across 16 different issues proposed during the second session of the 98th Congress in 1984.  These issues ranged from providing aid to El Salvador to education spending to funding the Nicaraguan Contras.  The Congressional Quarterly Almanac lists nine different types of votes on the issues, which we simplified to yea, nay, and either did not vote or otherwise make a position known so as to create a trinary-valued set of features.  The aim is to determine the political party affiliation, either Democrat or Republican, of the Congressmen based on their votes across these issues.\vspace{0.025cm}
\item[] \-\hspace{0.0cm}$R_\textnormal{HD-303}${\small{\sf{\textbf{: Heart Disease Dataset.}}}} The third is the Cleveland Clinic Foundation heart disease dataset.  The entire dataset is composed of entries from 920 patients in the United States, Hungary, and Switzerland, out of which we use only 303 patients who visited the Cleveland Clinic Foundation.  For each patient, 75 features were extracted; this feature set was reduced to 13 attributes, such as the age, chest pain type, resting blood pressure, and serum cholesterol of the patient.  The classification aim is to distinguish between either the absence or the presence of heart disease, regardless of the severity.  As with the other datasets, the classes are unbalanced: there are 164 patients who did not present with heart disease and 139 patients who did.\vspace{0.05cm}
\end{itemize}

\noindent We converted each vector-based dataset into graph-based ones by applying the standardized Euclidean distance to all pairs of feature vectors.  The visual assessment of cluster tendency (VAT) \cite{BezdekJC-conf2002a,BezdekJC-jour2005a,BezdekJC-jour2006a} and improved VAT (iVAT) \cite{HavensTC-jour2012a} algorithms were applied to the adjacency matrices of the graphs to reorganize the rows and columns of edge weights and highlight the underlying data structure.  The resulting plots are presented in the first two columns of figures A.1(a)--A.1(c).  Principal component analysis was used to visualize the bi-class data structure in two dimensions.  The resulting plots are shown in the third column of figures A.1(a)--A.1(c).

\begin{figure*}
\hspace{-0.225cm}\begin{tabular}{c}
   \includegraphics[height=1.0in]{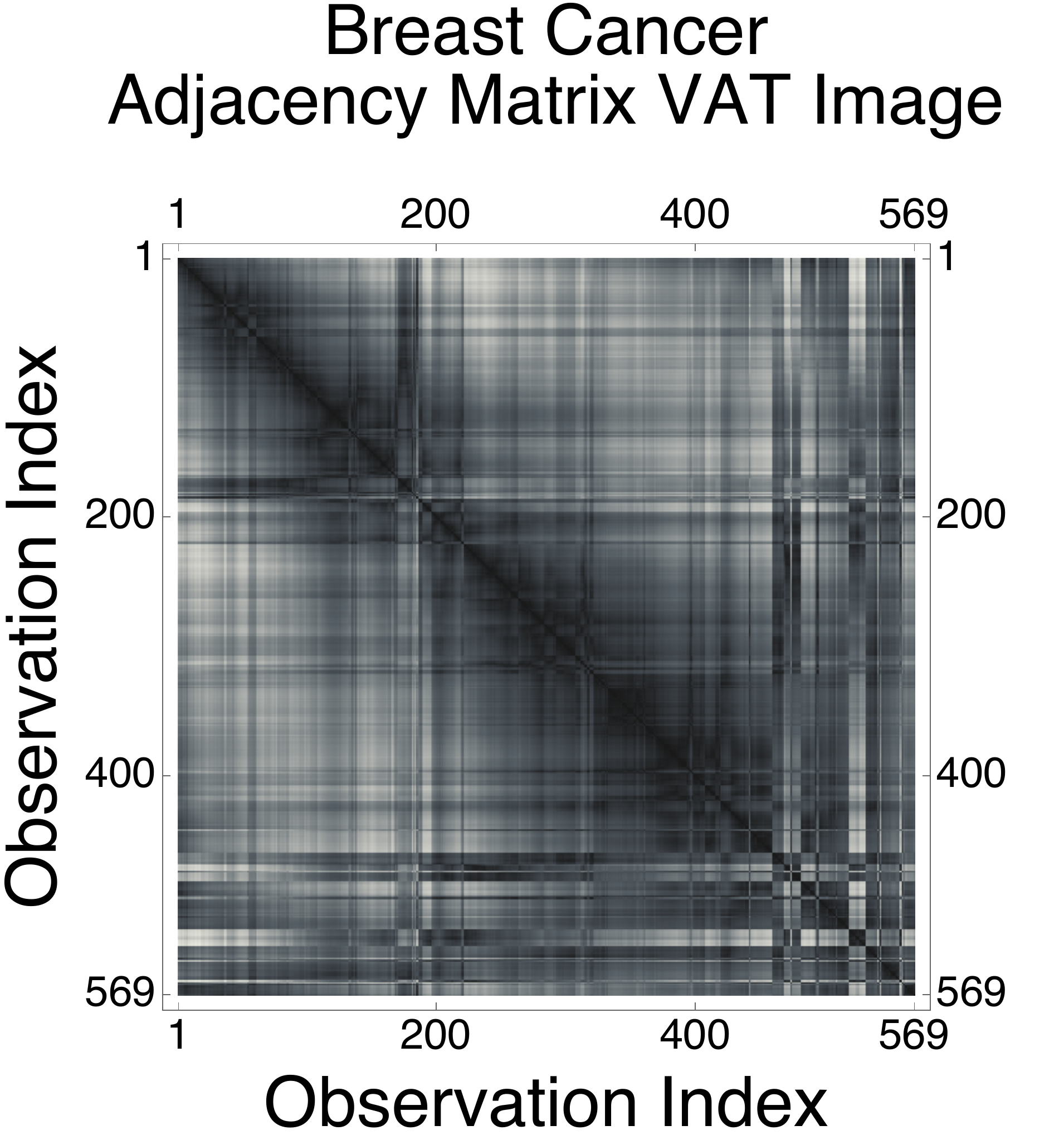} \includegraphics[height=1.0in]{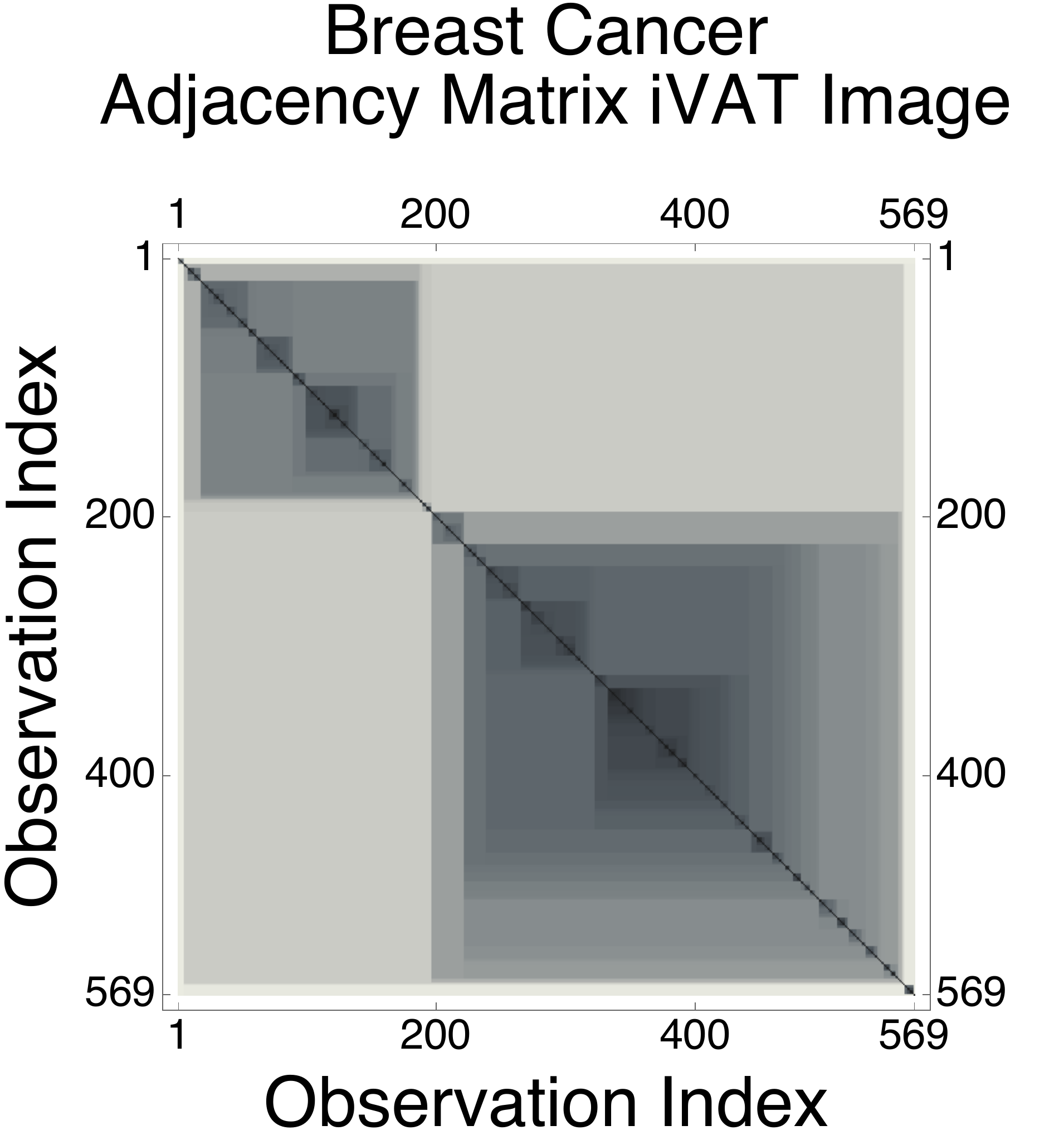} \includegraphics[height=1.0in]{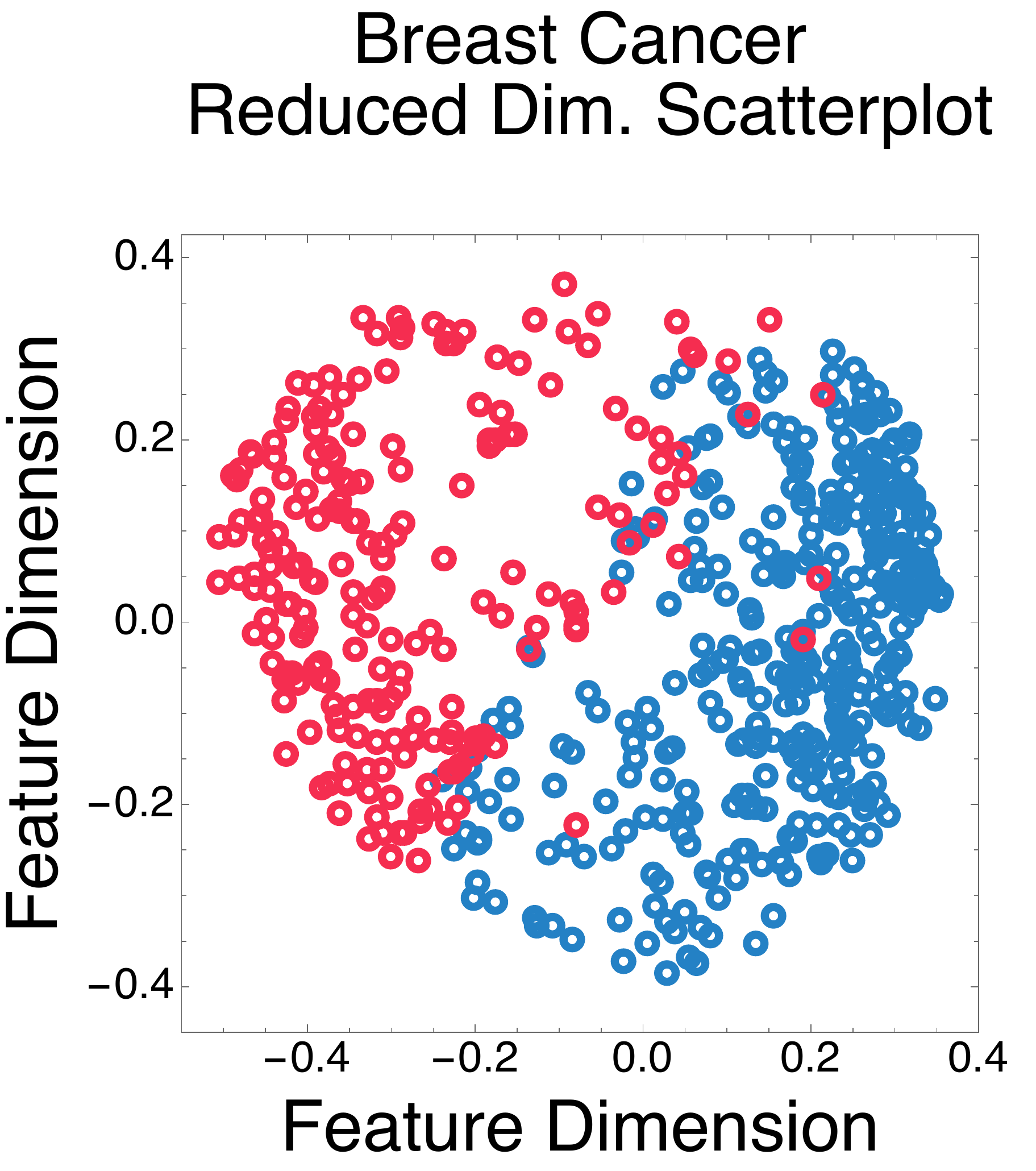} $\;\;\;\;\;\;$ \includegraphics[height=1.0in]{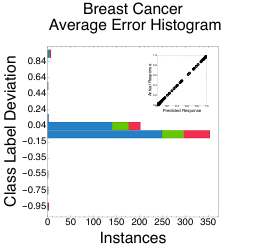} \hspace{-0.25cm} \includegraphics[height=1.0in]{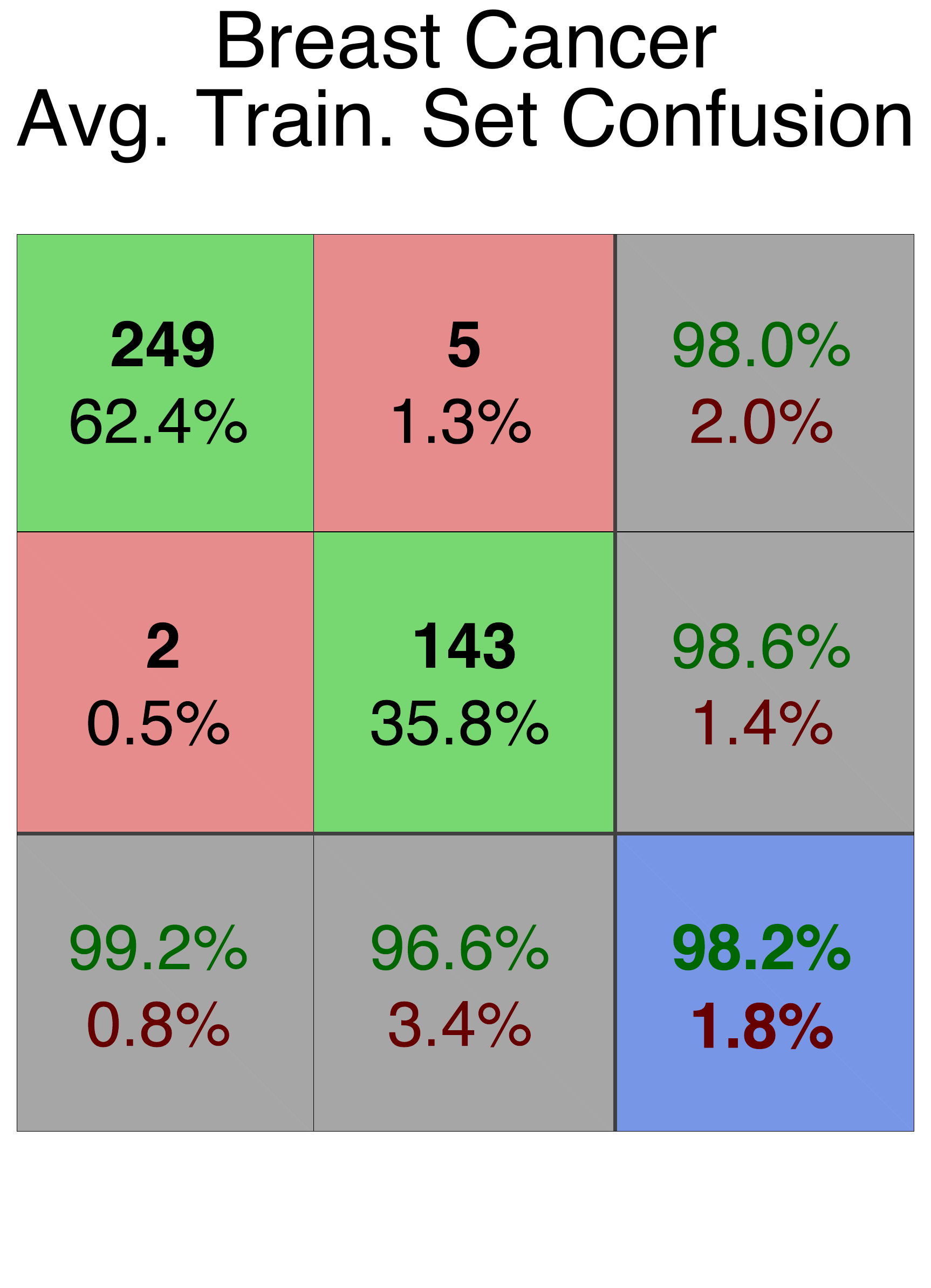} \hspace{0.075cm} \includegraphics[height=1.0in]{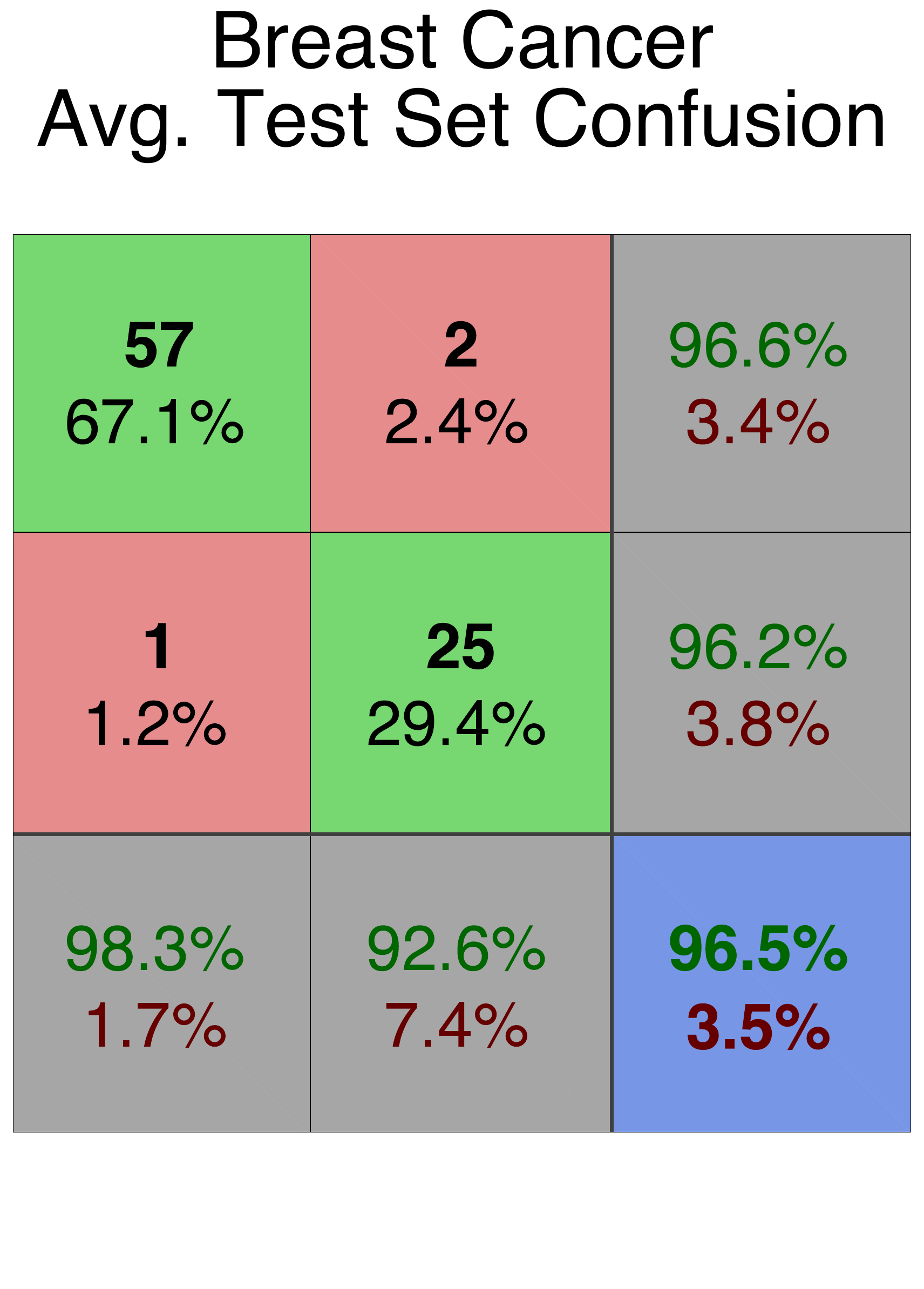} \hspace{0.075cm} \includegraphics[height=1.0in]{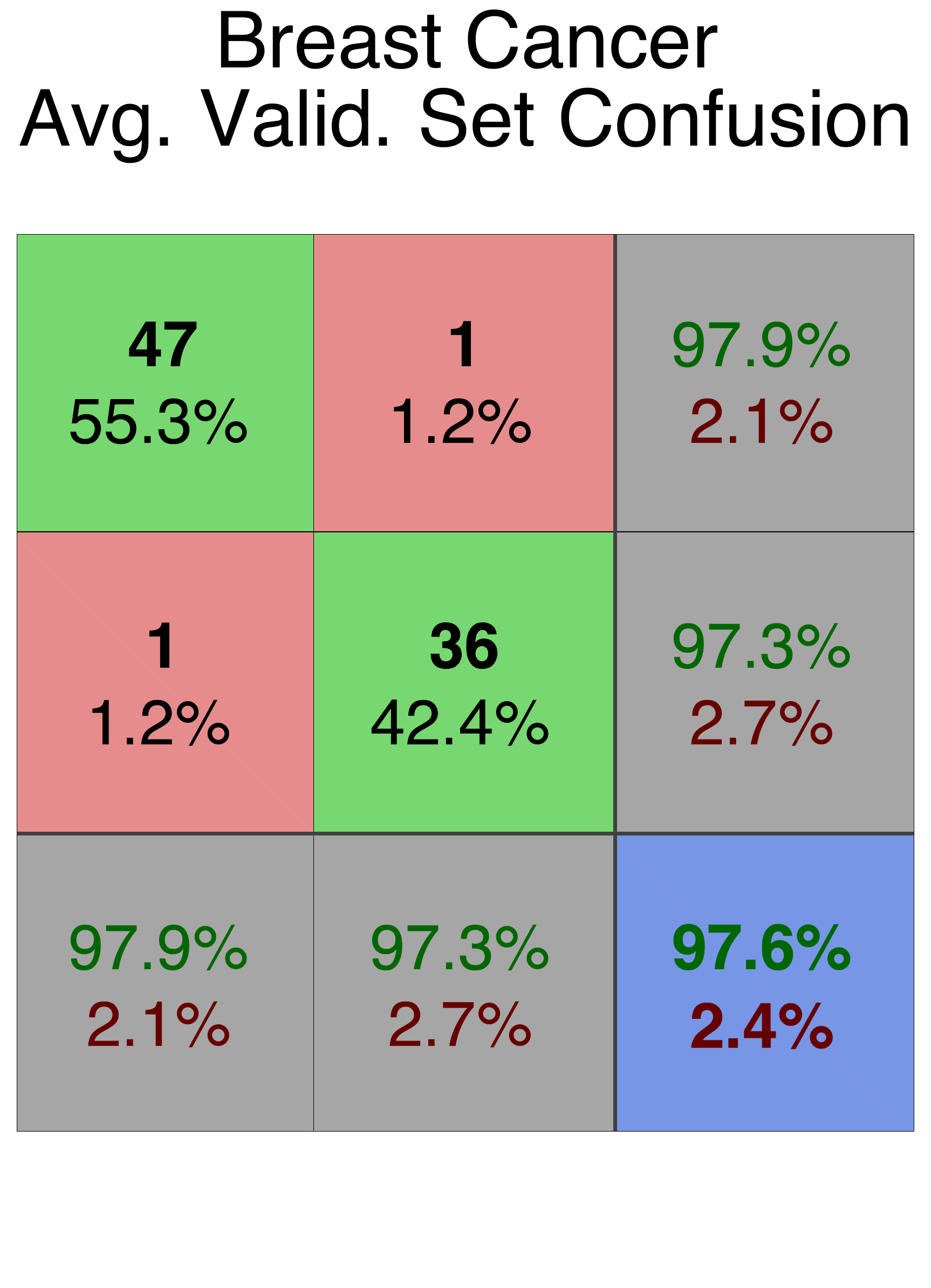}\vspace{-0.15cm}\\
   {\footnotesize (a)}\vspace{0.15cm}\\
   \includegraphics[height=1.0in]{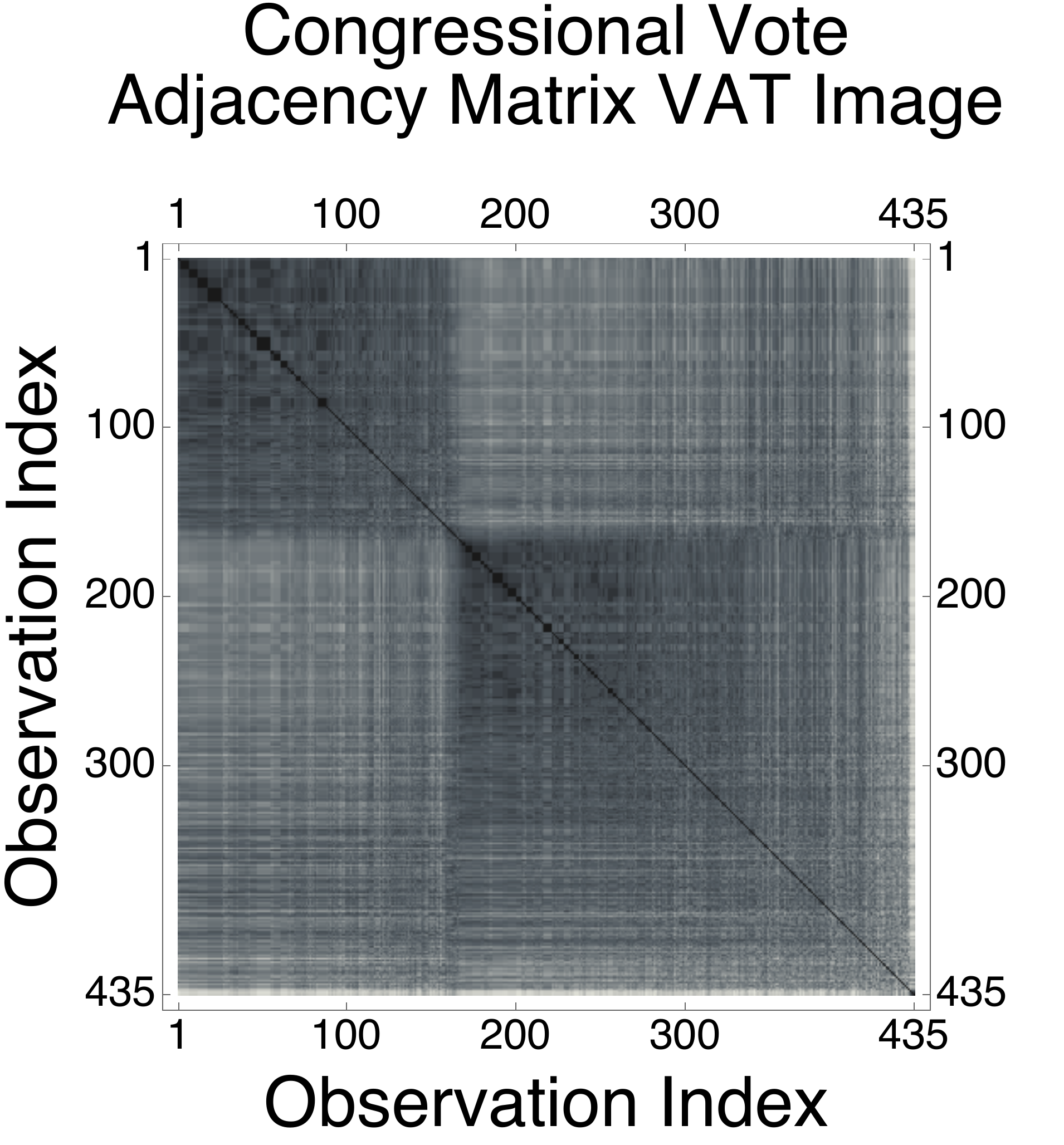} \includegraphics[height=1.0in]{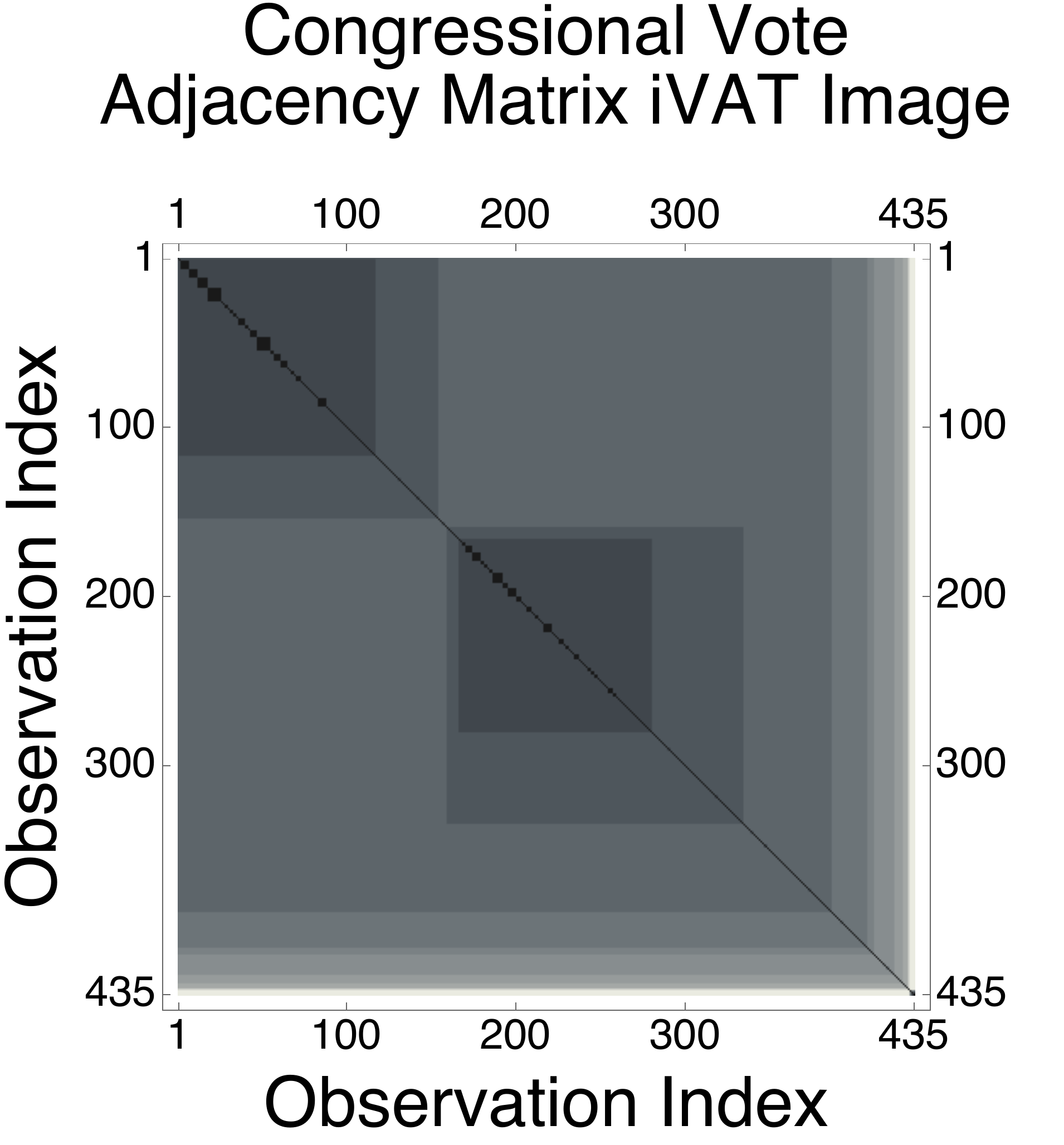} \includegraphics[height=1.0in]{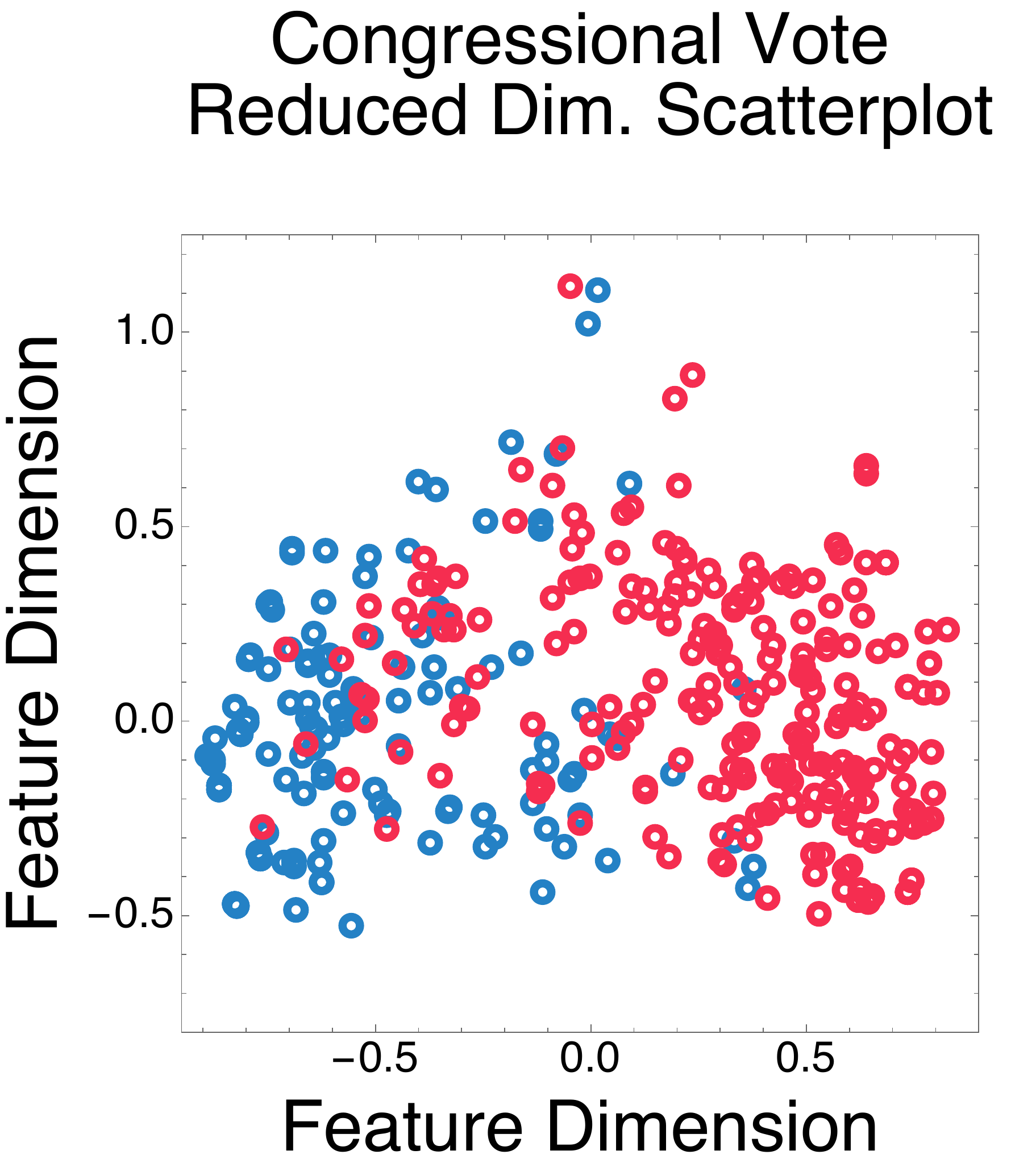} $\;\;\;\;\;\;$ \includegraphics[height=1.0in]{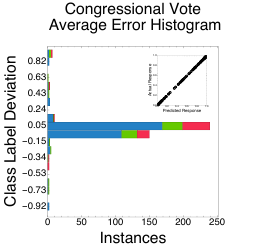} \hspace{-0.25cm} \includegraphics[height=1.0in]{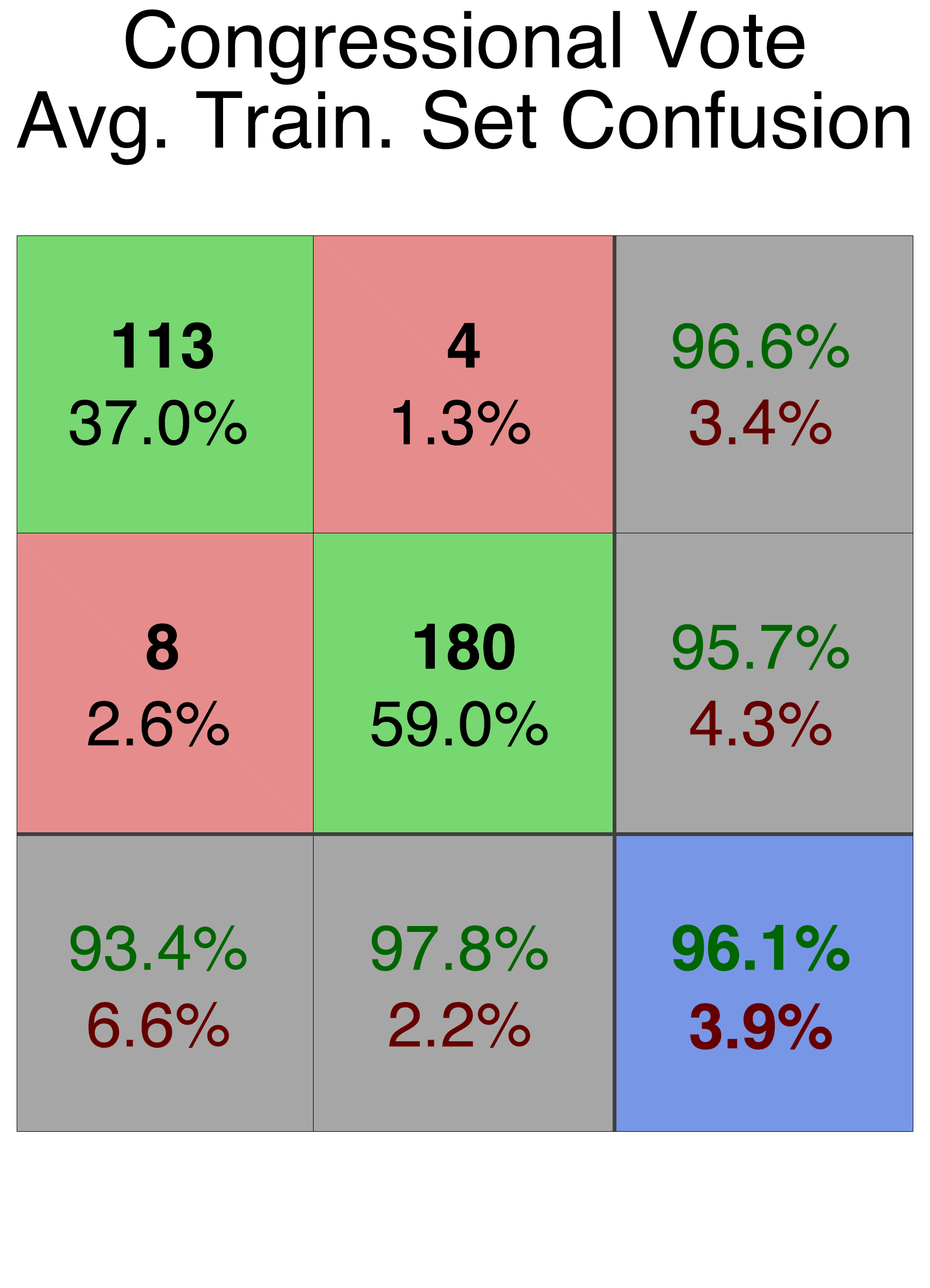} \hspace{0.075cm} \includegraphics[height=1.0in]{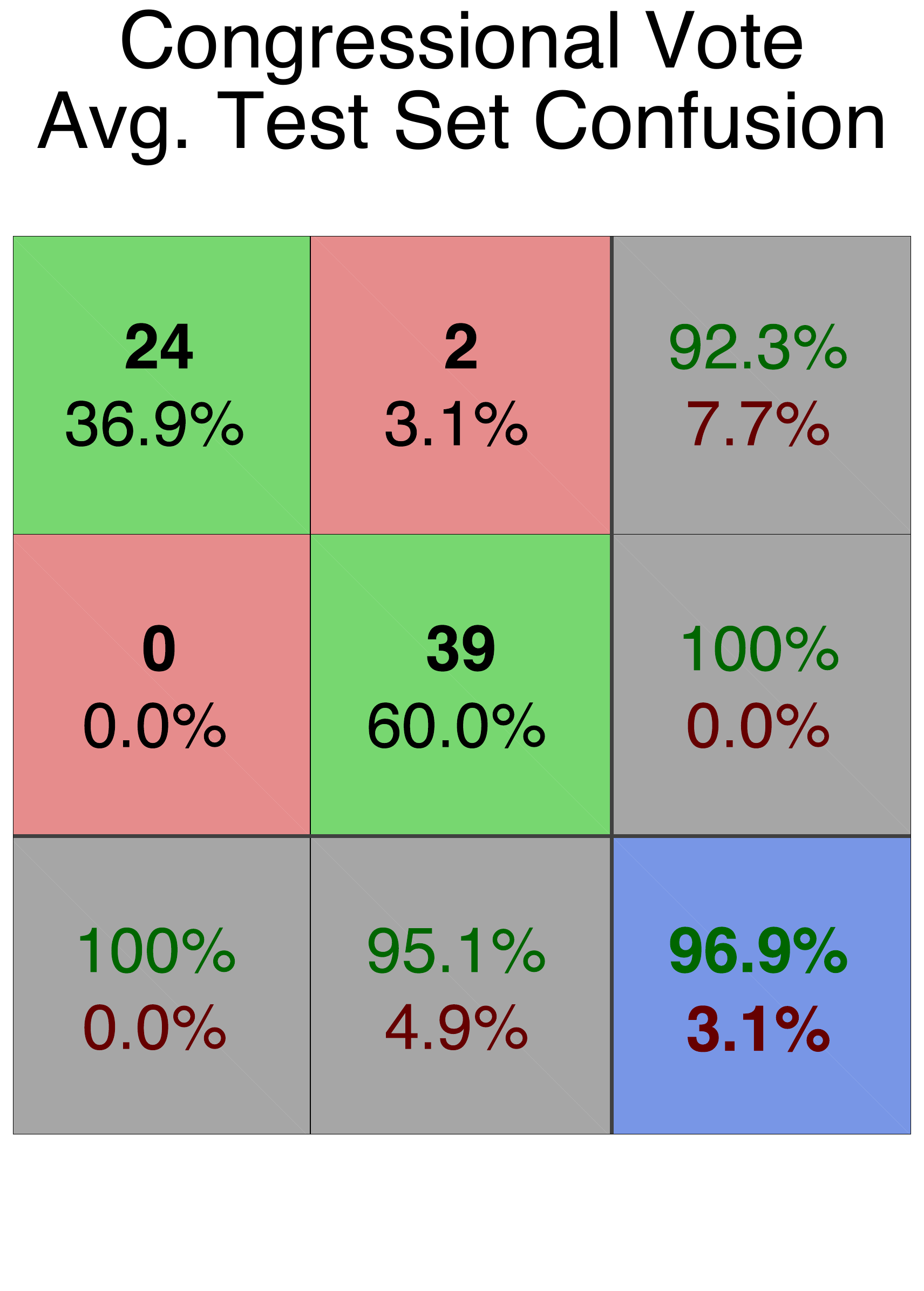} \hspace{0.075cm} \includegraphics[height=1.0in]{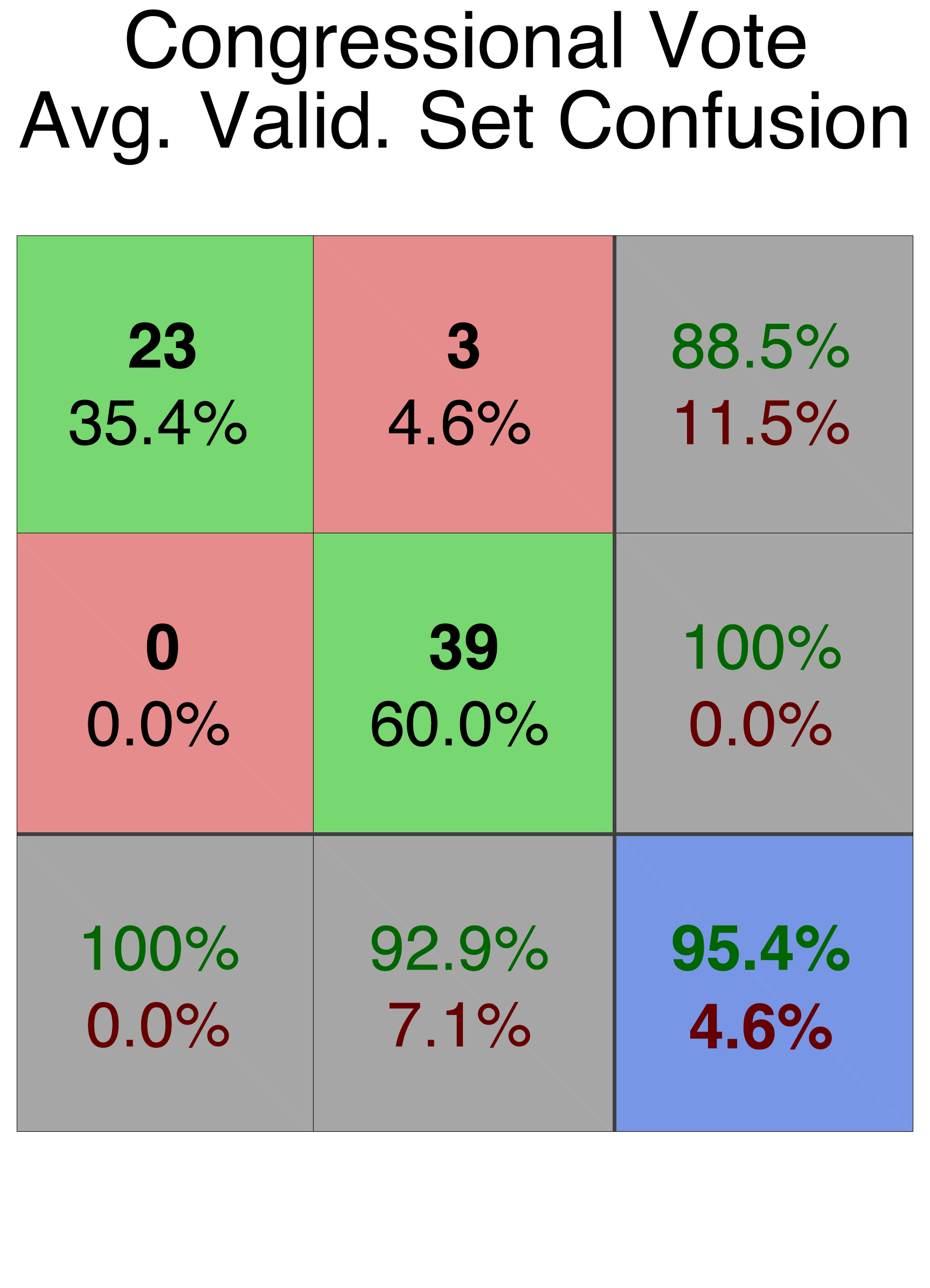}\vspace{-0.15cm}\\
   {\footnotesize (b)}\vspace{0.15cm}\\
   \includegraphics[height=1.0in]{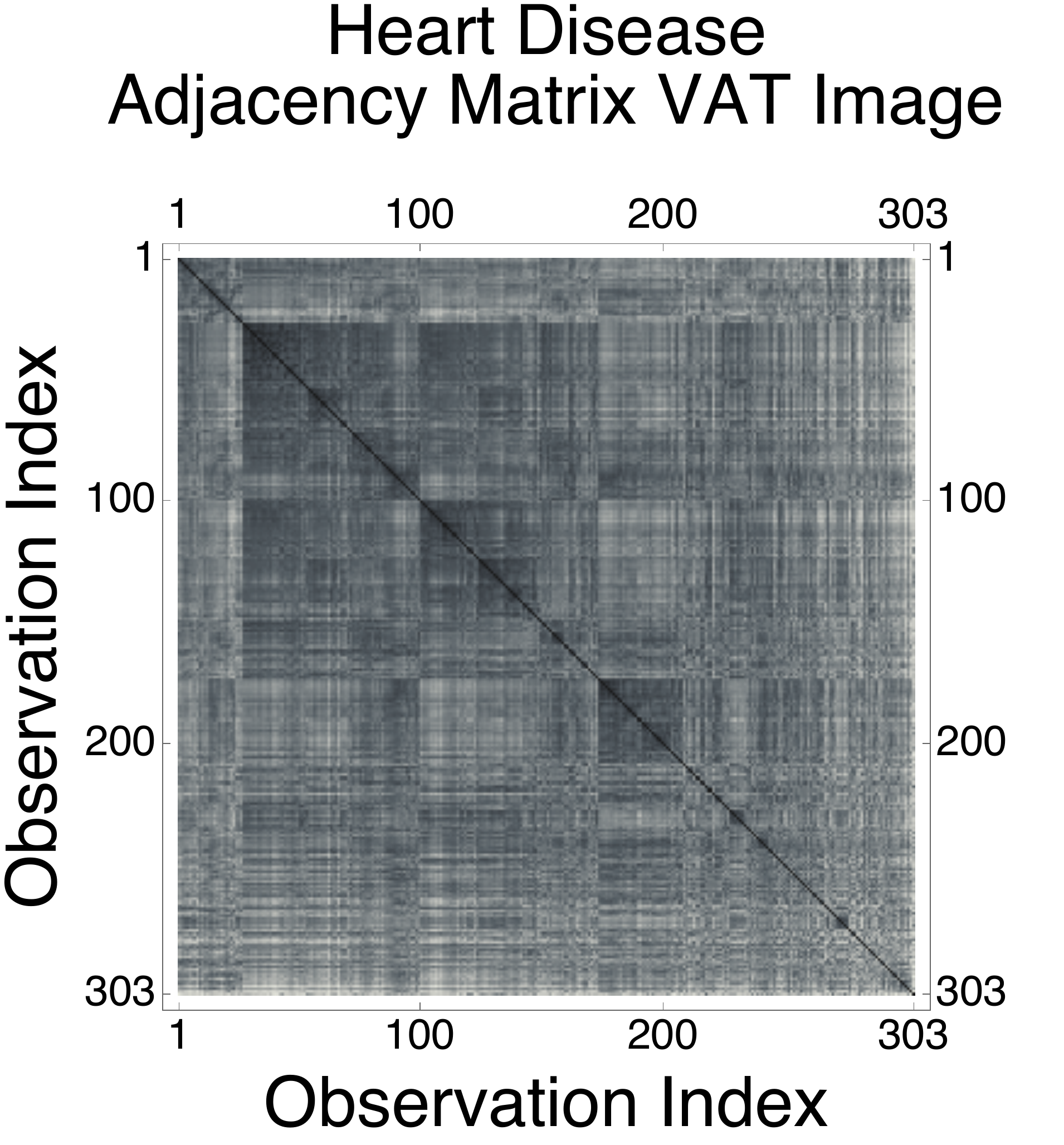} \includegraphics[height=1.0in]{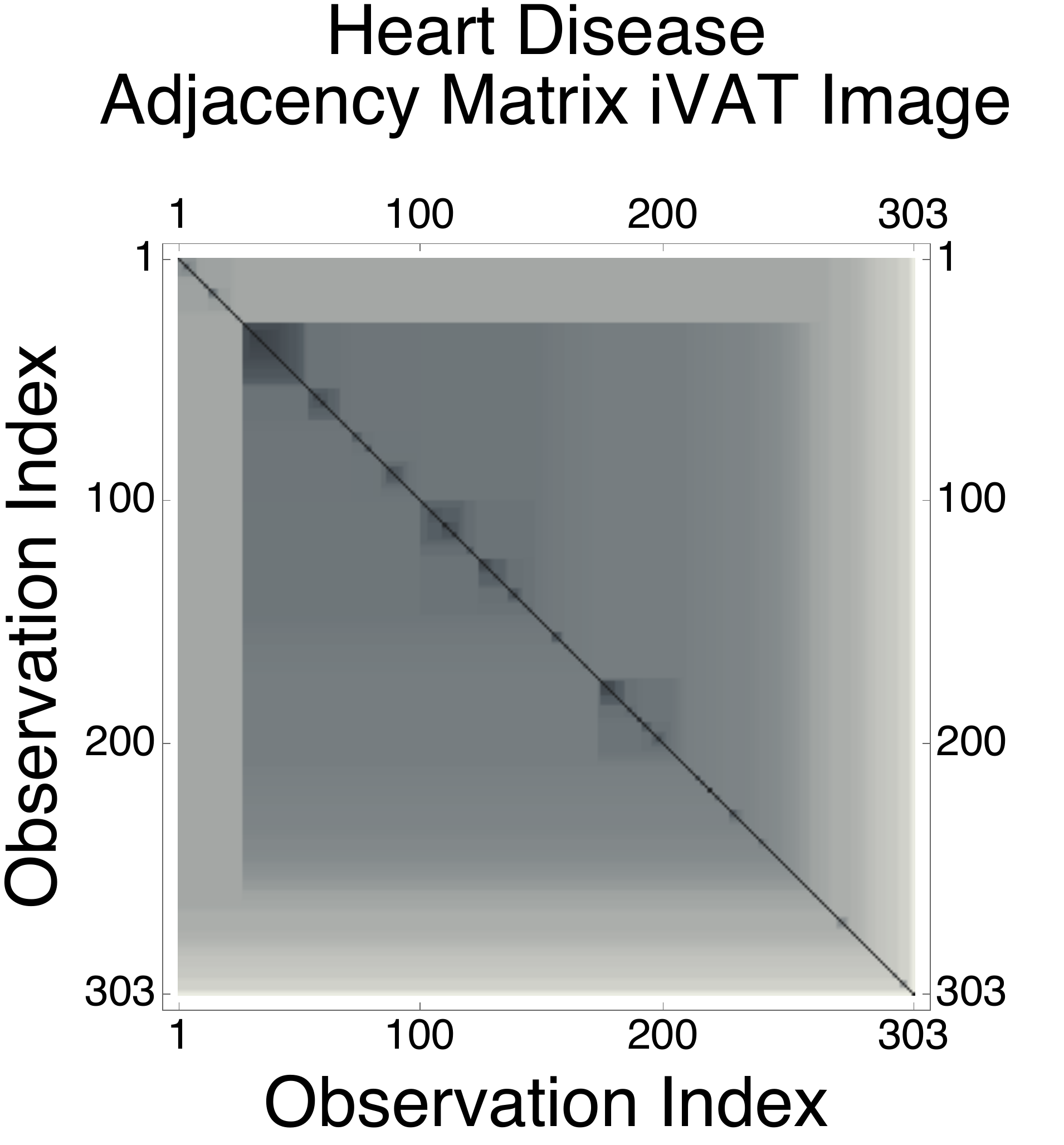} \includegraphics[height=1.0in]{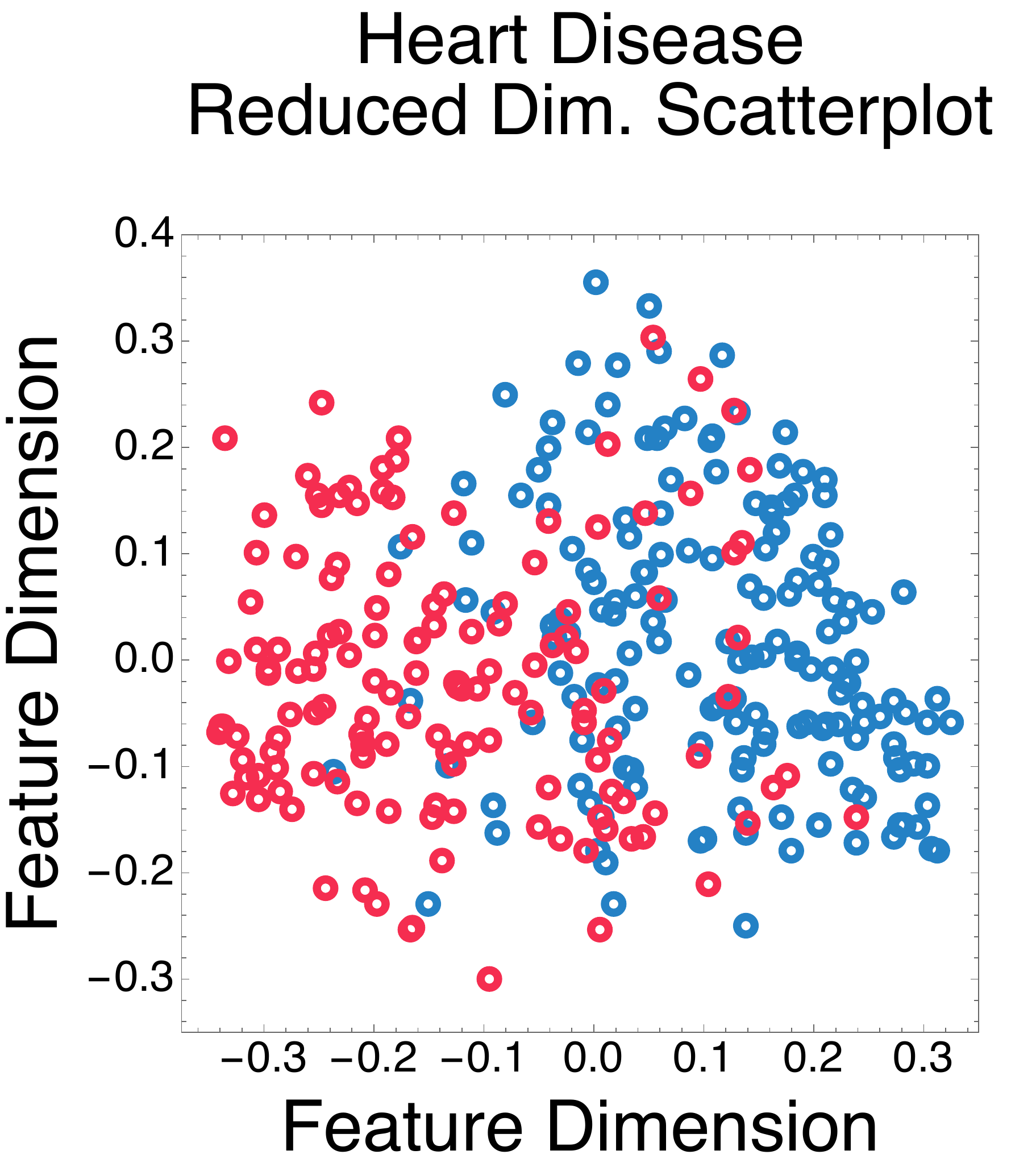} $\;\;\;\;\;\;$ \includegraphics[height=1.0in]{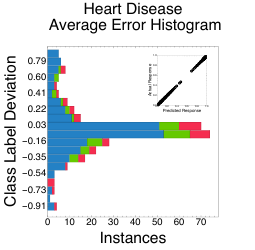} \hspace{-0.25cm} \includegraphics[height=1.0in]{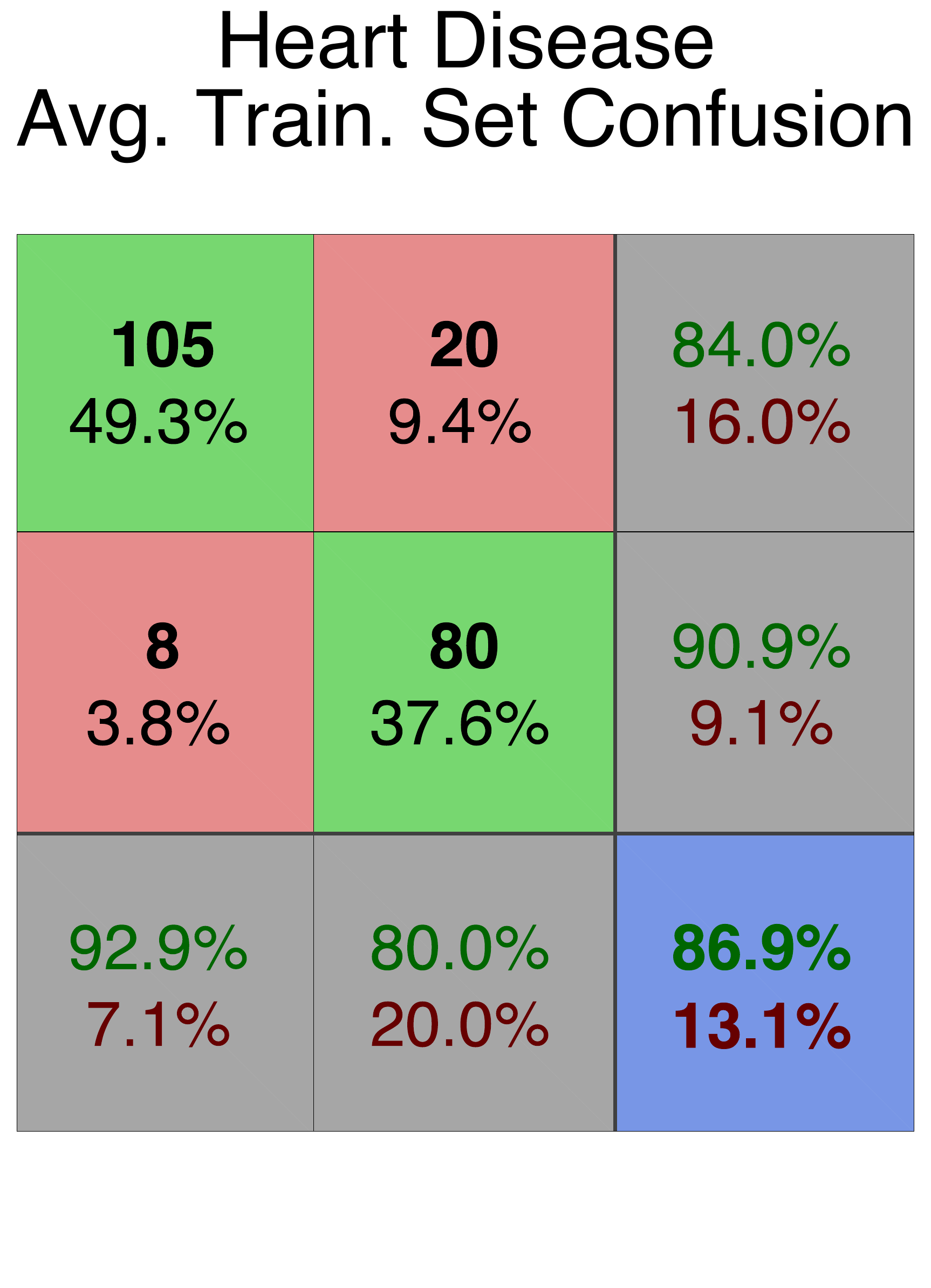} \hspace{0.075cm} \includegraphics[height=1.0in]{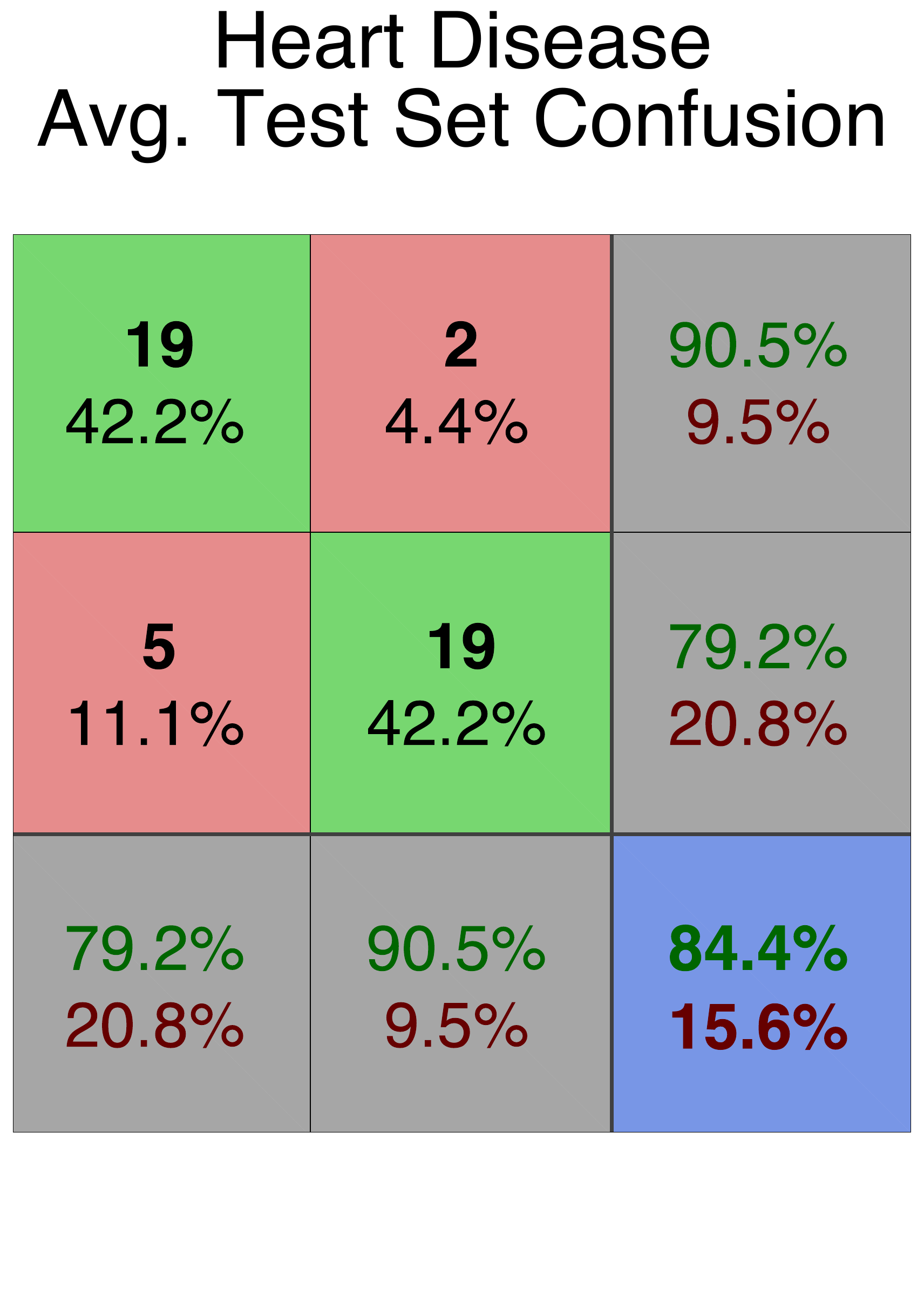} \hspace{0.075cm} \includegraphics[height=1.0in]{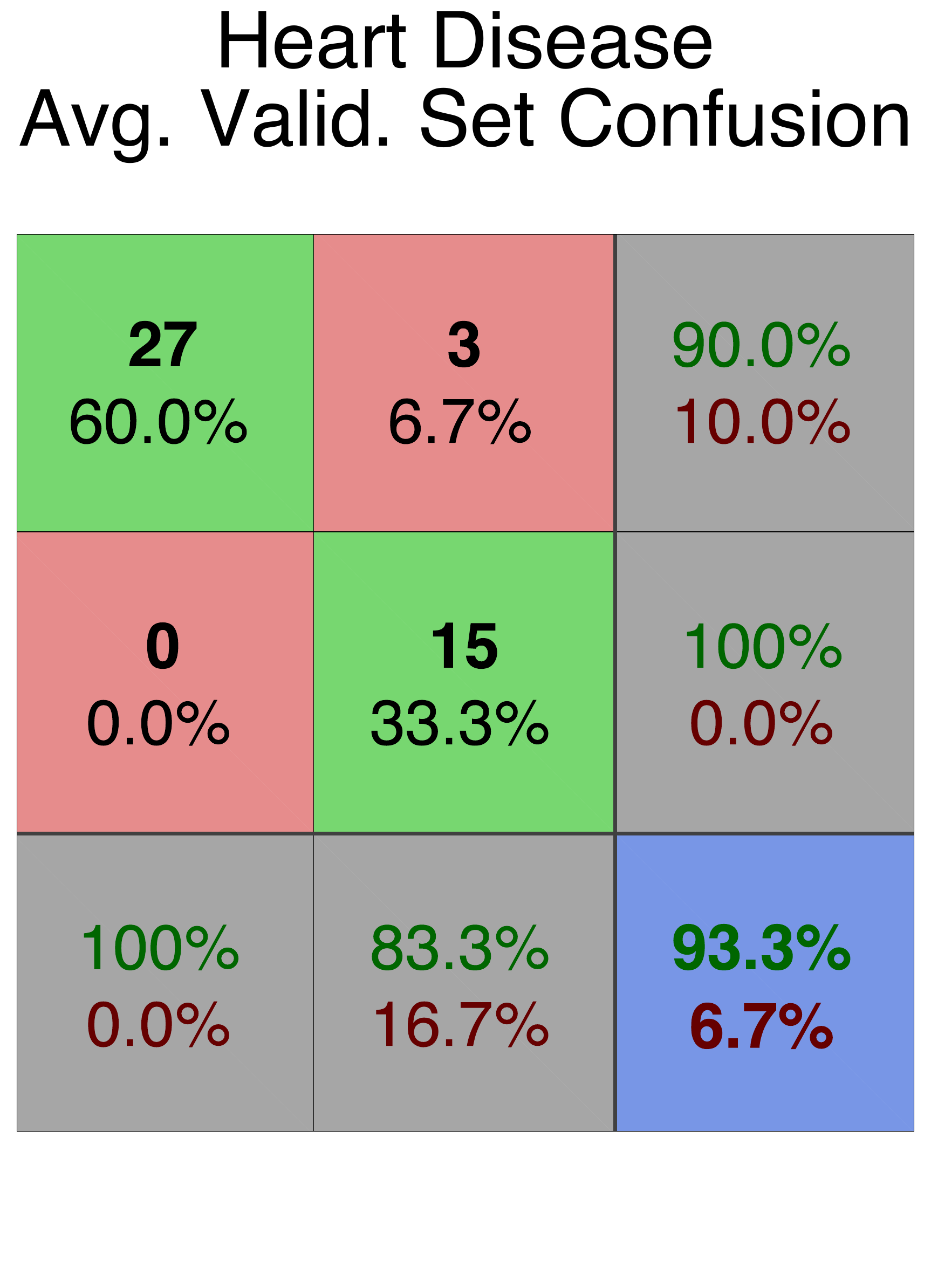}\vspace{-0.15cm}\\
   {\footnotesize (c)}\\
\end{tabular}
\caption*{Figure A.1: Classification results for graph-based datasets derived from feature vectors from the University of California-Irvine Machine Learning Repository.  We consider three datasets: (a) the University of Wisconsin diagnostic breast cancer dataset, (b) the US Congress voting dataset, and (c) the Cleveland Clinic Foundation heart disease dataset.  Each dataset was converted to a graph by applying a similarity measure to all pairs of observations.  For each row, the first two columns show the visual assessment of cluster tendency (VAT) and improved VAT (iVAT) images of the graph adjacency matrices.  Low observation-observation dissimilarities are denoted using dark colors, while high dissimilarities are represented with light colors.  Dark, blocky structures along the main diagonal in these images indicate structure in the data.  The third column shows a projected version of the dataset in a two-dimensional space, where the different colors correspond to the class labels.  The fourth through seventh columns highlight the classification results.  The fourth column shows the deviation in the predicted response from the desired response.  The subplot in this figure shows the predicted response from our graph-based RBF network and the actual response from the feature-vector-based RBF network.  The remaining columns contain confusion matrices for the training, test, and validation sets.  Both the vectorial and graph-based RBF networks returned the same responses, so we only show one set of plots for both network types.\vspace{-0.4cm}}
\end{figure*}

\subsection*{\small{\sf{\textbf{A.2$\;\;\;$Simulation Findings}}}}

Classification results for $R_\textnormal{BC-569}$, $R_\textnormal{USC-435}$, and $R_\textnormal{HD-303}$ are provided in the fourth through sixth columns of figures A.1(a)--A.1(c).  We obtained these results from 100 Monte Carlo simulations with different parameter initializations along with different training, testing, and validation datasets.  In these figures, the fourth column captures the deviation of the produced response from the predicted label, which were averaged across the Monte Carlo simulations.  The remaining columns highlight the average training, test, and validation set confusion matrices; the classification responses in these matrices were rounded and normalized.  We considered the same experimental setup as in section 3.

For $R_\textnormal{BC-569}$, $R_\textnormal{USC-435}$, and $R_\textnormal{HD-303}$, we considered the following minimum and maximum number of prototypes, respectively: 10 and 45, 10 and 35, and 10 and 30.  The same limits were considered for both the graph- and vector-based RBF networks. 

\vspace{0.05cm}{\small{\sf{\textbf{Results.}}}} The iVAT image for $R_\textnormal{BC-569}$ illustrates that the data naturally decompose into two dominant, mostly separated groups, each of which roughly corresponds to the class labels.  These two groups contain some substructure.  The VAT image, in contrast, reveals that the data are distributed along curved surfaces.  These characteristics are corroborated by the reduced-dimensionality scatterplot.  Due to the separatedness of the two classes, the RBF networks achieved near-perfect performance.  The average deviation between the actual and predicted class label responses for each sample was low and on the order of $\textnormal{0.001} \pm \textnormal{0.12}$, as shown by the error histogram plot.  The average training, test, and validation rates were $\textnormal{98.27} \pm \textnormal{0.54}$\%, $\textnormal{96.51} \pm \textnormal{0.97}$\%, and $\textnormal{97.65} \pm \textnormal{0.84}$\%, respectively, across the Monte Carlo\\ \noindent simulations.  The subplot in the error histogram graph highlights that the responses from the graph- and vector-based RBF networks were equivalent for all of the samples, since the responses all lie along the main diagonal.  These classification performance results are hence equivalent for both types of networks, given that they were trained in the same manner and on the same randomly chosen sets.

With regards to the two other datasets, the VAT and iVAT images for $R_\textnormal{USC-435}$ indicate that the relationships have a naturally bi-group structure for the chosen feature metric.  The classes are not, however, relatively compact and separated: both tendency plots indicate that there is a fair amount of mixing between the two classes.  This moderate class overlap is also present in the reduced-dimensionality scatterplot.  For $R_\textnormal{HD-303}$, the VAT and iVAT images do not show any clear multi-modal structure.  They instead, suggest that the classes are heavily mixed.  Despite these issues, the RBF networks achieved reasonably good performance: average recognition rates of $\textnormal{96.14} \pm \textnormal{1.09}$\%, $\textnormal{96.92} \pm \textnormal{1.11}$\%, and $\textnormal{95.40} \pm \textnormal{0.99}$\% were obtained for the $R_\textnormal{USC-435}$ training, testing, and validation sets, respectively.  For $R_\textnormal{HD-303}$, they were $\textnormal{86.94} \pm \textnormal{3.23}$\%, $\textnormal{84.47} \pm \textnormal{3.51}$\%, and $\textnormal{93.35} \pm \textnormal{3.66}$\%, respectively.  The error histograms show that the deviation was about $\textnormal{0.001} \pm \textnormal{0.19}$ for $R_\textnormal{USC-435}$ and $-\textnormal{0.009} \pm \textnormal{0.39}$ for $R_\textnormal{HD-303}$.  Many of the errors were caused by samples\\ \noindent located near the borders of the two classes, especially for $R_\textnormal{HD-303}$.  For both $R_\textnormal{USC-435}$ and $R_\textnormal{HD-303}$, the error histogram subplots again indicate that the vector- and graph-based networks supplied the same responses.  


\vspace{0.05cm}{\small{\sf{\textbf{Methodological Discussions and Insights.}}}} The simulation results indicate that both the graph- and vector-based RBF networks output the same responses for these three datasets, as predicted by the reformulation theory in the previous section.  The reformulation theory guarantees this equivalency in the event that the graph's edge weights correspond to distances from some, potentially unknown, metric pairwise applied to a set of, potentially unknown, vectors.  Note that a vector realization does not need to be found in order for this duality to theoretically and empirically hold.  It merely must provably exist, which occurs whenever the edge weights are positive, symmetric, and anti-reflexive along with satisfying the triangle inequality.  

Due to the equivalency property, both networks share the same advantages and disadvantages.  Investigators must hence be judicious in their choice of similarity or dissimilarity measure if they wish for the graph-based networks to yield the best recognition rates and have good generalization capabilities.  This is analogous to selecting good features to describe the objects in the vector-data case.

Some dissimilarity measures may not innately adhere to the properties needed for network duality.  In such circumstances, it would be necessary to transform the graph weights before network training occurs.  This can be done in a variety of ways, such as using multi-dimensional scaling techniques.  However, as we show in the following experiments, relying on multi-dimensional scaling is not always prudent, as an explicit assumption about the underlying metric is made and may not coincide with the latent metric of the relationships.  It is often better, for data analysis purposes, to let the data describe what is an appropriate metric.  This would entail finding the least number of edits to the relationships such that they are positive and adhere to the triangle inequality.  Converting arbitrary relationships to metric distances can be facilitated by iteratively projecting the relationships onto a series of half-spaces, which is known to have a closed-form solution \cite{RockafellarRT-book1997a}.  That is, for each relationship, the violation error would be computed and used to define the affine projection to the closest valid point on the half-space.  This iterative process yields a Boyle-Dykstra alternating-optimization procedure \cite{BoyleJP-coll1986}.  Convergence is guaranteed, in the limit, to an adjacency matrix with the least number of edits according to the chosen error metric.





\clearpage\newpage
\subsection*{\small{\sf{\textbf{B$\;\;\;$Graph-based RBF Network Properties}}}}

\begin{itemize}
\item[] \-\hspace{0.0cm}{\small{\sf{\textbf{Proposition 2.1.}}}} Let $R \!\in\! \mathbb{R}^{n \times n}_{0,+}$ be a symmetric, non-negative, reflexive adjacency matrix of a weighted, di-\\ \noindent rected graph such that entries $r_{p,q} \!\in\! \mathbb{R}_{0,+}$ correspond to distances from a metric applied to pairs of latent vectors\\ \noindent $x_p,x_q \!\in\! \mathbb{R}^d$.  For $v_j \!=\! \sum_{p=1}^n u_{j,p}x_p/\sum_{r=1}^n u_{j,r}$ and $v_{j,i}^r \!=\! u_{j,i}/\sum_{q=1}^n u_{j,q}$, $u_{j,i}$, $u_{j,q} \!\in\! \mathbb{R}$, the objective\\ \noindent function given in
\begin{equation}
\sum_{i=1}^n\sum_{k=1}^g \Bigg(y_{i,k} \!-\! f_2\Bigg(w_{0,k} \!+\! \sum_{j=1}^c w_{j,k} f_1\Bigg(\!\!\left\langle\! x_i \!-\! \left.\sum_{p=1}^n u_{j,p}x_p\!\right/\!\sum_{r=1}^n u_{j,r},\; x_i \!-\! \left.\sum_{q=1}^n u_{j,q}x_q\!\right/\!\sum_{r=1}^n u_{j,r}\! \right\rangle\!\!\Bigg)\!\Bigg)\!\Bigg)^{\!2} \tag{2.1}
\end{equation}
is equivalent to
\begin{equation}
\sum_{i=1}^n\sum_{k=1}^g \Bigg(y_{i,k} \!-\! f_2\Bigg(w_{0,k} \!+\! \sum_{j=1}^c w_{j,k} f_1\Bigg((R(v_j^r))_i \!-\! (v_j^r)^\top\! R (v_j^r)/2\Bigg)\!\Bigg)\!\Bigg)^{\!2}. \tag{2.5} 
\end{equation}

Alternatively, let $R \!\in\! \mathbb{R}^{n \times n}_{0,+}$ be a matrix such that it is non-negative, $r_{i,j} \!\geq\! 0$, symmetric, $r_{i,j} \!=\! r_{j,i}$, reflexive, $r_{i,i} \!=\! 0$, and where the entries obey the triangle inequality $r_{i,k} \!\leq\! r_{i,j} \!+\! r_{j,k}$, $\forall i,j,k$.  A latent vector realization exists and the objective function given in (2.1) is equivalent to that in (2.5).
 \vspace{0.05cm}
\end{itemize}
\begin{itemize}
\item[] \-\hspace{0.0cm}{\small{\sf{\textbf{Proof:}}}} For the first part of the proposition, equation (2.5) can be obtained as follows for any inner-product-induced norm\vspace{0.25cm}

\noindent $\displaystyle \left\langle\! x_i \!-\! \left.\sum_{p=1}^n u_{j,p}x_p\!\right/\!\sum_{r=1}^n u_{j,r},\; x_i \!-\! \left.\sum_{q=1}^n u_{j,q}x_q\!\right/\!\sum_{r=1}^n u_{j,r}\! \right\rangle$\vspace{0.025cm}\\

\begin{align*}
\phantom{\;\;\;\;\;\;\;\;\;\;} &= \Bigg(\sum_{r=1}^n u_{j,r}\Bigg)^{\!-2}\Bigg\langle x_i \!-\! \sum_{p=1}^n u_{j,p}x_p,\; x_i \!-\! \sum_{q=1}^n u_{j,q}x_q \Bigg\rangle\\
&= -\Bigg(2\sum_{r=1}^n u_{j,r}\Bigg)^{\!-2} \sum_{p=1}^n u_{j,p} \sum_{q=1}^n u_{j,q} \Bigg(\|x_p \!-\! x_q\|^2 - \|x_i \!-\! x_p\|^2 + \|x_i \!-\! x_q\|^2 \Bigg)\\
&= -\Bigg(2\sum_{r=1}^n u_{j,r}\Bigg)^{\!-2} \sum_{p=1}^n u_{j,p} \sum_{q=1}^n u_{j,q} \Bigg(r_{p,q} - r_{i,p} - r_{i,q}\Bigg)\\
&= -\Bigg(2\sum_{r=1}^n u_{j,r}\Bigg)^{\!-2} \Bigg(\sum_{p=1}^n u_{j,p}(e_i - e_p)^\top\Bigg) R \Bigg(\sum_{q=1}^n u_{j,q}(e_i \!-\! e_q) \Bigg)\\
&= -\Bigg(e_i^\top R e_i/2 - e_i^\top R (v_j^r)\Bigg) + (v_j^r)^\top R (v_j^r)/2\\
&= (R(v_j^r))_i \!-\! (v_j^r)^\top\! R (v_j^r)/2,
\end{align*}
where $e_k$, $e_p$, and $e_q$ are the $k$th, $p$th, and $q$th unit vectors, respectively.  Substituting $(R(v_j^r))_i \!-\! (v_j^r)^\top\! R (v_j^r)/2$ in place of the inner product in (2.1) establishes the equivalency.

The second part of the proposition is immediate from the first part and some simple arguments involving the properties of inner-product-induced norms.  $\;\;\Box$
\end{itemize}

\vspace{0.2cm}

\begin{itemize}
\item[] \-\hspace{0.0cm}{\small{\sf{\textbf{Proposition 2.3.}}}} Let $R \!\in\! \mathbb{R}^{n \times n}_{0,+}$ be a symmetric, non-negative, anti-reflexive adjacency matrix of a weighted, di-\\ \noindent rected graph such that entries $r_{p,q} \!\in\! \mathbb{R}_{0,+}$ correspond to distances from a metric applied to pairs $x_p,x_q \!\in\! \mathbb{R}^d$.

Suppose that the output-layer activation function of a two-layer, vector RBF network is linear.  For this network, the gradient-descent update for the $q$th radial basis prototype $v_q \!\in\! \mathbb{R}^d$ is $v_q \!\leftarrow\! v_q \!-\! \eta \sum_{k=1}^n \epsilon_{q,k}^h(x_i \!-\! v_q)$,\\ \noindent with the hidden error $\epsilon_{q,k}^h \!=\! w_{i,q} \nabla_{v_q} h_{q,k}$ for a given $x_i \!\in\! \mathbb{R}^d$.  This is prototype update is equivalent to replacing the distance $h_{q,k}$ between the $q$th prototype and the $k$th feature-space vector by 
\begin{equation}
\left\langle x_i \!-\! \left.\sum_{p=1}^n u_{j,p}x_p\!\right/\!\sum_{r=1}^n u_{j,r} \!-\! \eta\Bigg(\sum_{r=1}^n \epsilon_{r,j}^h x_r \!-\! \epsilon_{r,j}^h \left.\sum_{p=1}^n u_{j,p}x_p\!\right/\!\sum_{r=1}^n u_{j,r}\Bigg),\,\cdot\, \right\rangle. \tag{2.8}
\end{equation}
Here, the term $\epsilon_{k,j}^h \!=\! w_{i,j} \nabla_{v_j} h_{k,j}$ is obtained from the response gradient $\nabla_{v_q}\! \sum_{j=1}^c w_{i,j}h_{j,k} \!=\! w_{i,q}\nabla_{v_q} h_{q,k}$,\\ \noindent with $\nabla_{v_q} h_{q,k} \!=\! 0$ for $j \!\neq\! q$.  Equation (2.8) is the same as requiring that the observation-prototype distance in (2.5), (2.7), and (2.11) be replaced by
\begin{equation}
d_{j,i} + \left(4\sum_{r=1}^n u_{j,r}\right)^{\!\!-2}\!\!\sum_{p,q=1}^n\! u_{j,p} u_{j,q} \eta\sum_{y=1}^n \epsilon_{y,j}^h\Bigg(\! r_{i,y} \!-\! r_{i,q} \!-\! \eta \epsilon_{y,j}^h r_{p,q}\!\Bigg), \tag{2.9}
\end{equation}
for $v_j \!=\! \sum_{p=1}^n u_{j,p}x_p/\sum_{r=1}^n u_{j,r}$ and $v_{j,i}^r \!=\! u_{j,i}/\sum_{q=1}^n u_{j,q}$, for $u_{j,i}$, $u_{j,q} \!\in\! \mathbb{R}$.  This expression is solely in\\ \noindent terms of the adjacency weights.  
\vspace{0.05cm}
\end{itemize}

\begin{itemize}
\item[] \-\hspace{0.0cm}{\small{\sf{\textbf{Proof:}}}} Equation (2.8) follows from differentiating the vector-data sum-of-squared-errors expression with respect to the RBF prototype and substituting $\sum_{p=1}^n u_{j,p} x_p/\sum_{r=1}^n u_{j,r}$ for the prototype.  Equation (2.9) can be obtained as follows\vspace{0.25cm}

\noindent $\displaystyle\left\langle x_i \!-\! \left.\sum_{p=1}^n u_{j,p}x_p\!\right/\!\sum_{r=1}^n u_{j,r} \!-\! \eta\Bigg(\sum_{r=1}^n \epsilon_{r,j}^h x_r \!-\! \epsilon_{r,j}^h \left.\sum_{p=1}^n u_{j,p}x_p\!\right/\!\sum_{r=1}^n u_{j,r}\Bigg),\cdot \right\rangle$\vspace{0.025cm}\\

\begin{align*}
\phantom{\;\;\;\;\;\;\;\;\;\;} &= \left(\sum_{r=1}^n u_{j,r}\right)^{\!-2}\!\left\langle \sum_{p=1}^n u_{j,p} \Bigg(x_i \!-\! x_p \!-\! \eta\sum_{r=1}^n \epsilon_{r,j}^h(x_r \!-\! x_p)\Bigg),\sum_{q=1}^n u_{j,q} \Bigg(x_i \!-\! x_q \!-\! \eta\sum_{r=1}^n \epsilon^h_{r,j}(x_r \!-\! x_q)\Bigg) \right\rangle\\
&= \left(\sum_{r=1}^n u_{j,r}\right)^{\!-2}\!\! \Bigg(\sum_{p=1}^n\sum_{q=1}^n u_{j,p} u_{j,q} \langle x_i,x_i \rangle \!-\! u_{j,p} u_{j,q} \langle x_i,x_q \rangle \!-\! u_{j,p} u_{j,q} \left\langle x_i,\eta\sum_{r=1}^n \epsilon^h_{r,j}(x_r \!-\! x_q) \right\rangle\\
&\;\;\;\;\;\;\;\;\;\;\;\;\;\;\;\;\;\;\;\;\;\;\;\;\;\;\;\;\;\;\;\;\; - u_{j,p} u_{j,q} \langle x_p,x_i \rangle \!+\! u_{j,p} u_{j,q} \langle x_p,x_q \rangle \!+\! u_{j,p} u_{j,q} \left\langle x_p,\eta\sum_{r=1}^n \epsilon^h_{r,j}(x_r \!-\! x_q) \right\rangle\\
&\;\;\;\;\;\;\;\;\;\;\;\;\;\;\;\;\;\;\;\;\;\;\;\;\;\;\;\;\;\;\;\;\; - u_{j,p} u_{j,q} \left\langle x_i,\eta\sum_{r=1}^n \epsilon^h_{r,j}(x_r \!-\! x_p) \right\rangle \!-\! u_{j,p} u_{j,q} \left\langle x_q,\eta\sum_{r=1}^n \epsilon^h_{r,j}(x_r \!-\! x_p) \right\rangle\\
&\;\;\;\;\;\;\;\;\;\;\;\;\;\;\;\;\;\;\;\;\;\;\;\;\;\;\;\;\;\;\;\;\; + u_{j,p} u_{j,q} \left\langle \eta\sum_{r=1}^n \epsilon^h_{r,j}(x_r \!-\! x_p),\eta\sum_{r=1}^n \epsilon^h_{r,j}(x_r \!-\! x_q) \right\rangle\!\Bigg)\\
&= \left(4\sum_{r=1}^n u_{j,r}\right)^{\!-2}\!\! \Bigg(\sum_{p=1}^n\sum_{q=1}^n u_{j,p} u_{j,q} \Bigg(\!-\! r_{i,q} \!-\! r_{p,i} \!+\! r_{p,q} \!-\! \left\langle x_i,\eta\sum_{r=1}^n \epsilon^h_{r,j}(x_r \!-\! x_q) \right\rangle \!+\! \left\langle x_p,\eta\sum_{r=1}^n \epsilon^h_{r,j}(x_r \!-\! x_q) \right\rangle\\
&\;\;\;\;\;\;\;\;\;\;\;\;\;\;\;\;\;\;\;\;\;\;\;\;\;\;\;\;\;\;\;\;\; - \left\langle x_q,\eta\sum_{r=1}^n \epsilon^h_{r,j}(x_r \!-\! x_p) \right\rangle \!+\! \left\langle \eta\sum_{r=1}^n \epsilon^h_{r,j}(x_r \!-\! x_p),\eta\sum_{r=1}^n \epsilon^h_{r,j}(x_r \!-\! x_q) \right\rangle \!\Bigg)\!\Bigg)\\
&= \left(4\sum_{r=1}^n u_{j,r}\right)^{\!-2}\!\! \Bigg(\sum_{p=1}^n\sum_{q=1}^n u_{j,p} u_{j,q} \Bigg(\!\!-\! r_{i,q} \!-\! r_{p,i} \!+\! r_{p,q} \!-\! \eta\sum_{r=1}^n \epsilon^h_{r,j}\langle x_i, x_r \rangle \!+\! \eta\sum_{r=1}^n \epsilon^h_{r,j} \langle x_i, x_q \rangle\\
&\;\;\;\;\;\;\;\;\;\;\;\;\;\;\;\;\;\;\;\;\;\;\;\;\;\;\;\;\;\;\;\;\; + \eta\sum_{r=1}^n \epsilon^h_{r,j} \langle x_p, x_r \rangle \!-\! \eta\sum_{r=1}^n \epsilon^h_{r,j} \langle x_p, x_q \rangle \!-\! \eta\sum_{r=1}^n \epsilon^h_{r,j}\langle x_q, x_r \rangle \!+\! \eta\sum_{r=1}^n \epsilon^h_{r,j} \langle x_q, x_p \rangle\\
&\;\;\;\;\;\;\;\;\;\;\;\;\;\;\;\;\;\;\;\;\;\;\;\;\;\;\;\;\;\;\;\;\; + \eta^2\sum_{r=1}^n (\epsilon^h_{r,j})^2 \langle x_r, x_r\rangle \!+\! \eta^2\sum_{r=1}^n (\epsilon_{r,j}^h)^2\langle x_r, x_q \rangle \!-\! \eta^2\sum_{r=1}^n (\epsilon_{r,j}^h)^2\langle x_r, x_p \rangle \!+\! \eta^2\sum_{r=1}^n (\epsilon^h_{r,j})^2\langle x_p, x_q \rangle\!\Bigg)\!\Bigg)\\
&= \left(4\sum_{r=1}^n u_{j,r}\right)^{\!-2}\! \Bigg(\sum_{p=1}^n\sum_{q=1}^n u_{j,p} u_{j,q} \Bigg(\!-\! r_{i,q} \!-\! r_{p,i} \!+\! r_{p,q} \!-\! \eta\sum_{y=1}^n \epsilon_{y,j}\langle x_i, x_y \rangle \!+\! \eta\sum_{y=1}^n \epsilon^h_{y,j} \langle x_i, x_q \rangle \!+\! \eta\sum_{y=1}^n \epsilon_{y,j} \langle x_p, x_y \rangle\\
&\;\;\;\;\;\;\;\;\;\;\;\;\;\;\;\;\;\;\;\;\;\;\;\;\;\;\;\;\;\;\;\;\; - \eta\sum_{y=1}^n \epsilon^h_{y,j} \langle x_p, x_q \rangle \!-\! \eta\sum_{y=1}^n \epsilon^h_{y,j}\langle x_q, x_y \rangle \!+\! \eta\sum_{y=1}^n \epsilon^h_{y,j} \langle x_q, x_p \rangle \!+\! \eta^2 \sum_{y=1}^n (\epsilon^h_{y,j})^2 (r_{y,q} \!+\! r_{p,y} \!-\! r_{p,q}) \Bigg)\!\Bigg)\\
&= \left(4\sum_{r=1}^n u_{j,r}\right)^{\!-2}\!\! \Bigg(\sum_{p=1}^n\sum_{q=1}^n u_{j,p} u_{j,q} \Bigg(\!\!-\! r_{i,q} \!-\! r_{p,i} \!+\! r_{p,q} \!+\! \eta\sum_{y=1}^n \epsilon_{y,j}^h\Bigg(\!\!-\! \langle x_i, x_y \rangle \!+\! \langle x_i, x_q \rangle \!+\! \langle x_p, x_y \rangle\\
&\;\;\;\;\;\;\;\;\;\;\;\;\;\;\;\;\;\;\;\;\;\;\;\;\;\;\;\;\;\;\;\;\; - \langle x_p, x_q \rangle \!-\! \langle x_q, x_y \rangle \!+\! \langle x_q, x_p \rangle \!\Bigg) \!+\! \eta^2 \sum_{y=1}^n (\epsilon_{y,j}^h)^2 \Bigg(r_{y,q} \!+\! r_{p,y} \!-\! r_{p,q}\Bigg)\! \Bigg)\!\Bigg)\\
&= \left(4\sum_{r=1}^n u_{j,r}\right)^{\!-2}\!\! \Bigg(\sum_{p=1}^n\sum_{q=1}^n u_{j,p} u_{j,q} \Bigg(\!\!-\! r_{i,q} \!-\! r_{p,i} \!+\! r_{p,q} \!+\! \eta\sum_{y=1}^n \epsilon_{y,j}^h\Bigg(\! r_{i,y} \!-\! r_{i,q} \!-\! r_{p,y} \!+\! r_{q,y}\!\Bigg)\\
&\;\;\;\;\;\;\;\;\;\;\;\;\;\;\;\;\;\;\;\;\;\;\;\;\;\;\;\;\;\;\;\;\; +\! \eta^2 \sum_{y=1}^n (\epsilon_{y,j}^h)^2 \Bigg(r_{y,q} \!+\! r_{p,y} \!-\! r_{p,q}\Bigg)\! \Bigg)\!\Bigg)\vspace{0.2cm}\\
&= (R(v_j^r))_i \!-\! (v_j^r)^\top\! R (v_j^r)/2 \!+\! \left(4\sum_{r=1}^n u_{j,r}\right)^{\!-2}\!\sum_{p=1}^n\sum_{q=1}^n u_{j,p} u_{j,q} \eta\sum_{y=1}^n \epsilon_{y,j}^h\Bigg(\! r_{i,y} \!-\! r_{i,q} \!-\! r_{p,y} \!+\! r_{q,y}\!\Bigg)\vspace{0.2cm}\\
&\;\;\;\;\;\;\;\;\;\;\;\;\;\;\;\;\;\;\;\;\;\;\;\;\;\;\;\;\;\;\;\;\; +\! \left(4\sum_{r=1}^n u_{j,r}\right)^{\!-2}\!\sum_{p=1}^n\sum_{q=1}^n u_{j,p} u_{j,q}\eta^2 \sum_{y=1}^n (\epsilon_{y,j}^h)^2 \Bigg(r_{y,q} \!+\! r_{p,y} \!-\! r_{p,q}\Bigg)\\
&= (R(v_j^r))_i \!-\! (v_j^r)^\top\! R (v_j^r)/2 \!+\! \left(4\sum_{r=1}^n u_{j,r}\right)^{\!-2}\!\sum_{p=1}^n\sum_{q=1}^n u_{j,p} u_{j,q} \eta\sum_{y=1}^n \epsilon_{y,j}^h\Bigg(\! r_{i,y} \!-\! r_{i,q} \!-\! \eta \epsilon_{y,j}^h r_{p,q}\!\Bigg),
\end{align*}
which is solely in terms of entries from the adjacency matrix and the adjacency weights.  $\;\;\Box$
\end{itemize}

\end{document}